\documentclass[runningheads]{llncs}

\usepackage[english]{babel}

\usepackage{amsmath}
\usepackage{graphicx}
\usepackage[colorlinks=true, allcolors=blue]{hyperref}
\usepackage{macro}

\usepackage[T1]{fontenc}
\usepackage{graphicx}
\begin{document}
\title{Robustness Verification of Graph Neural Networks Via Lightweight Satisfiability Testing}
\titlerunning{Robustness of GNNs Via Lightweight SAT}
%

\author{Chia-Hsuan Lu\inst{1}
\and
Tony Tan\inst{2}
\and
Michael Benedikt\inst{1}
}
\authorrunning{Lu, Tan, and Benedikt}
%
\institute{University of Oxford, UK  \and
University of Liverpool, UK}

\date{}

\maketitle


\begin{abstract}
Graph neural networks (GNNs) are the predominant architecture for learning over graphs.
As with any machine learning model, an important issue is the detection of attacks, where an adversary can change the output with a small perturbation of the input. 
Techniques for solving the \emph{adversarial robustness problem} --- determining whether an attack exists --- were originally developed for image classification.
In the case of graph learning, the attack model usually considers changes to the graph structure in addition to or instead of the numerical features of the input, and the state of the art techniques proceed via reduction to constraint solving, working on top of powerful solvers, e.g. for mixed integer programming.
We show that it is possible to improve on the state of the art in structural robustness by replacing the use of powerful solvers by calls to efficient \emph{partial solvers}, which run in polynomial time but may be incomplete.
We evaluate our tool $\toolname$ on a diverse set of GNN variants and datasets.
\end{abstract}


\section{Introduction}

Graph neural networks (GNNs) have become the dominant model for graph learning, used in
many critical applications, such as physical and life sciences~\cite{ShlomiBV21,DuvenaudMIBHGA15,KearnesCBPR16}.
An important issue in safety of machine learning is \emph{robustness verification}:
verifying that small changes to an input do not change the systems' classification.
Robustness verification of machine learning systems emerged in
image recognition~\cite{ml-robust-premier-braiek-24,assessing-louloudakis-22,robust-study-benchmarking-25},
and has since been explored for a variety of neural architectures.
In particular, robustness has been extended to GNNs, where new issues arise as one should consider attacks consisting of small perturbations to the graph structure. Indeed, it has been shown that small structural perturbations can represent effective attacks on common graph learning architectures \cite{benf}. The past literature has considered a variety of models for graph structure perturbation, including the addition and deletion of edges \cite{BojchevskiG19,ZugnerG20,JinSPZ20,LadnerEA25}, or even node injection \cite{LaiZPZ24}.
Robustness verification makes sense in the context of every graph learning problem, including node classification and graph classification.

State of the art robustness verification techniques for GNNs are based on two ideas. First one iteratively computes upper and lower bounds to arrive at constraints capturing the existence of a suitable attack. For example the constraint might have variables indexed by edges, representing whether the edge is present or not. Secondly, one solves these constraints, using a general-purpose constraint solver. A canonical example is \cite{HojnyZCM24}, which reduces robustness verification to mixed integer linear programming (MIP). The MIP problems are formed by approximating the behaviour of the GNN, and their satisfiability is tested using the MIP solver Gurobi.

The robustness problem is easily seen to be NP-complete,
and thus reducing to a generic solver for another NP-hard problem is natural. But it is not clear that these reductions allow the solvers to exploit the structure of the robustness problem. And indeed, despite advances, robustness verification is currently limited to very small GNNs: e.g. 3 layers.
Here we take an alternative approach, and make use of \emph{lightweight solvers}. As in prior approaches we iteratively refine approximations to the evaluation of a GNN on a set of graphs obtained from small perturbations, expressing our approximation as a set of constraints. Unlike prior approaches, we test satisfiability of the constraints only approximately, developing our own \emph{partial oracles} that can efficiently determine whether constraints are satisfiable or not, but may return unknown.
Although the lightweight solver approach may seem simpler than one based on existing general-purpose solvers, we show that its performance is significantly better than state-of-the-art.

This paper is organised as follows.
In Section~\ref{sec:prelim} we introduce the necessary notations
and review GNNs and the robustness problem.
In Section~\ref{sec:oracle} we present our notion of partial oracles and how they can be used for
robustness analysis.
We introduce some optimization strategies in Section~\ref{sec:opt}.
We present our experimental results in Section~\ref{sec:exper}.
Related work is discussed in Section~\ref{sec:related}.
Finally, we conclude in Section~\ref{sec:conc}.
Missing proofs, additional variations of the problems that our tool can handle, and experimental results can be found in the appendix.


\section{Preliminaries} \label{sec:prelim}

Throughout the paper we deal with directed graphs in which each node is associated with a vector of reals.
\begin{definition}[Featured graphs]
    A \emph{featured graph} $\agraph$,
    or simply graph for short, is a tuple $\tuple{V, E, X}$,
    where $V$ is the set of vertices,
    $E\subseteq V \times V$ is the set of edges,
    and $\ifeatmap: V \to \reals^{m}$ is the input feature mapping.

    For a vertex $v \in V$,
    we denote the set of incoming neighbors of $v$,
    or simply neighbors for short,
    in a graph $\agraph$ by $\nbr_\agraph(v)$, i.e.,
    $\nbr_\agraph(v) := \setc{u \in V}{(u, v) \in E}$.
\end{definition}

We need to analyze the behavior of a network on a collection of graphs obtained through edge insertions and deletions.
We will abstract the inputs using the following notion of incomplete graphs,
a variant of the general notion of incomplete dataset in database theory \cite{AHV}:

\begin{definition}[Incomplete graphs]
    An \emph{incomplete graph} $\incgraph$ is a tuple \\$\tuple{V, E, E^\unknown, E^\non, X}$,
    where $V$ is the set of vertices,
    $E$ is the set of \emph{normal edges},
    $E^\unknown$ is the set of \emph{unknown edges},
    $E^\non$ is the set of \emph{non-edges},
    and $\ifeatmap: V \to \bbR^{m}$ is the input feature mapping,
    with $E$, $E^\unknown$, and $E^\non$ forming a partition of $V \times V$.

    For a vertex $v \in V$,
    we denote the set of incoming normal and incoming unknown neighbors of $v$ in $\incgraph$
    by $\nbr^\norm_\incgraph(v)$ and $\nbr^\unknown_\incgraph(v)$.
    We will sometimes abuse notation by treating a graph as a special kind of incomplete graph
    where $E^\unknown = \emptyset$.
\end{definition}
We denote the set of all incomplete graphs with vertex set $V$ and input feature mapping
$\ifeatmap$ by $\frH_{V, \ifeatmap}$.
We denote by $\frG_{V,\ifeatmap}$ the subset of ``normal graphs'' --- those with no unknown edges.
We will use $\incgraph$ to denote an incomplete graph, and often assume a vertex set $V$ and feature map $\ifeatmap$.

Incomplete graphs on the same vertex set have a natural ``information order'',
a partial order relation on the set $\frH_{V, \ifeatmap}$.
\begin{definition}[Refinement]
    For incomplete graphs $\incgraph_1, \incgraph_2 \in \frH_{V, \ifeatmap}$,
    we say that $\incgraph_2$ \emph{refines} $\incgraph_1$,
    denoted by $\incgraph_2 \subseteq \incgraph_1$,
    if:
    \begin{equation*}
        E_2 \subseteq E_1 \cup E_1^\unknown,
        \quad
        E_2^\unknown \subseteq E_1^\unknown,
        \ \text{and}\ \quad
        E_2^\non \subseteq E_1^\non \cup E_1^\unknown.
    \end{equation*}
\end{definition}
Intuitively, $\incgraph_2 \subseteq \incgraph_1$
means $\incgraph_2$ can be obtained from $\incgraph_1$ through a sequence converting an unknown edge into a normal edge and converting an unknown edge into a non-edge.
Featured graphs are the minimal elements of this order.

\begin{definition}[Completion]
    For an incomplete graph $\incgraph \in \frH_{V, \ifeatmap}$,
    the completions of $\incgraph$, denoted by $\comp{\incgraph}$,
    is the set of graphs that refine $\incgraph$.
\end{definition}
It is easy to see that $\incgraph_2 \subseteq \incgraph_1$ if and only if
$\comp{\incgraph_2} \subseteq \comp{\incgraph_1}$.

\begin{definition}[Grounding]
    For an incomplete graph $\incgraph \in \frH_{V, \ifeatmap}$,
    and a graph $\agraph \in \frG_{V, \ifeatmap}$,
    the \emph{grounding} $\agraph'$ of $\incgraph$ to $\agraph$ is the completion of $\incgraph$
    obtained by the procedure:
    for every edge $e \in E^\unknown$, if $e$ is an edge in $\agraph$,
    then $e$ is converted to a normal edge in $\agraph'$;
    otherwise, if $e$ is not an edge in $\agraph$,
    then $e$ is converted to a non-edge in $\agraph'$.
\end{definition}
Intuitively, grounding of $\incgraph$ to $\agraph$ is
the completion of $\incgraph$ that is closest to $\agraph$.
We will formalize this once we define a notion of distance.

We will also need a way of introducing incompleteness:
\begin{definition}[Relaxation]\label{def:relaxation-graph}
    For a graph $\agraph \in \frG_{V, \ifeatmap}$ and
    a set $E_p \subseteq V \times V$
    the relaxation of $\agraph$ by $E_p$,
    denoted by $\incgraph_{\agraph, E_p}$,
    is the incomplete graph obtained from $\agraph$
    by making all the edges in $E_p$ unknown, and subtracting off any edges in $E_p$ from both the edge set $E$ and
    the set of non-edges.
\end{definition}

Next, we define the notion of distance on the set of incomplete graphs.
\begin{definition}[Distance]
    For incomplete graphs $\incgraph_1, \incgraph_2 \in \frH_{V, \ifeatmap}$,
    for vertices $v, u \in V$,
    we say that $(v, u)$ is inconsistent between $\incgraph_1$ and $\incgraph_2$
    if either:
    $(v, u) \in E_1$ and $(v, u) \in E_2^\non$; or $(v, u) \in E_1^\non$ and $(v, u) \in E_2$.
    The distance between $\incgraph_1$ and $\incgraph_2$, denoted by $\dist{\incgraph_1, \incgraph_2}$ is defined as
    \begin{equation*}
        \dist{\incgraph_1, \incgraph_2}
        \ :=\ 
        \abs{\setc{(v, u) \in V \times V}{\text{$(v, u)$ is inconsistent between $\incgraph_1$ and $\incgraph_2$}}}.
    \end{equation*}
\end{definition}

Note that the distance between $\incgraph_1$ and $\incgraph_2$ can equivalently be defined as the minimum of the distances between completions of $\incgraph_1$ and of $\incgraph_2$.
One can check that
for all incomplete graphs $\incgraph, \incgraph_1, \incgraph_2 \in \frH_{n, \ifeatmap}$:
\begin{inparaenum}[(a)]
    \item $\dist{\incgraph, \incgraph} = 0$.
    \item $\dist{\incgraph_1, \incgraph_2} = \dist{\incgraph_2, \incgraph_1}$.
    \item $\incgraph_2 \subseteq \incgraph_1$ implies $\dist{\incgraph_2, \incgraph} \ge \dist{\incgraph_1, \incgraph}$.
\end{inparaenum}

A way to achieve the minimal distance from a graph to an incomplete graph is to use the grounding:
\begin{lemma}\label{lemma:ground}
    For every incomplete graph $\incgraph \in \frH_{V, \ifeatmap}$
    and graph $\agraph \in \frG_{V, \ifeatmap}$,
    letting $\agraph'$ be the grounding of $\incgraph$ to $\agraph$,
    $\dist{\agraph', \agraph} = \dist{\incgraph, \agraph}$.
\end{lemma}

The incomplete graph generated by relaxing a graph $\agraph$ has distance zero to the graph, since $\agraph$ is one of its completions:
\begin{lemma} \label{lemma:relax}
    For every graph $\agraph \in \frG_{V, \ifeatmap}$ and
    every set $E_p$ of pairs of nodes, $\agraph$ is a completion of its relaxation with respect to $E_p$, and hence
    $\dist{\incgraph_{\agraph, E_p}, \agraph} = 0$.
\end{lemma}

\subsubsection{Graph Neural Networks}
Graph neural networks are neural network architectures that can be used for a variety of machine learning tasks, including node-level classification: in this context, they take a featured graph as input and output predictions for each node.
We focus on \emph{message-passing neural networks}~\cite{GilmerSRVD17, HamiltonYL17},
which are architectures with a fixed number of layers.
Following the literature on robustness of GNNs~\cite{HojnyZCM24}, we use a variation that involves directed graphs, with aggregation over incoming nodes. We discuss modifications
for graph-level classification and for undirected graphs in Appendix~\ref{app:graphvariant} and \ref{app:undirected}.

Intuitively, at each layer, a GNN produces a new feature vector for each node, aggregating the previous-layer features from its incoming neighbors and applying a feedforward neural networks to the node's own previous-layer features.

\begin{definition}[Graph neural network]\label{def:gnn}
    An $L$-layer \emph{graph neural network} $\cA$
    consists of a sequence of dimensions $\gnndim{0}, \gnndim{1}, \ldots, \gnndim{L} \in \bbN^+$,
    an aggregation function $\aggr \in \set{\summ, \maxx, \mean}$,
    and, for each $1 \le \ell \le L$,
    learnable coefficient matrices $\coefC{\ell}, \coefA{\ell} \in \bbR^{\gnndim{\ell} \times \gnndim{\ell-1}}$
    together with bias vectors $\coefb{\ell} \in \bbR^{\gnndim{\ell}}$.
\end{definition}

\begin{definition}[Computation of a GNN] \label{def:gnn-computation}
    For an $L$-layer GNN $\cA$ and a featured graph $\agraph\in\frG_{V,\ifeatmap}$,
    the \emph{computation of $\cA$ on $\agraph$} is a sequence of features $\feat{\ell}_\agraph(v)$
    for $0 \le \ell \le L$ and $v \in V$.
    For $\ell = 0$, we set $\feat{0}_\agraph(v) := \ifeatmap(v)$.
    For $1 \le \ell \le L$,
    \begin{equation*}
        \feat{\ell}_\agraph(v)
        := \relup{
            \coefC{\ell} \! \cdot \! \feat{\ell-1}_\agraph(v)
            +
            \coefA{\ell} \! \cdot \! \aggr\left(\msetc{\feat{\ell-1}_\agraph(u)}{u \in \nbr_\agraph(v)}\right)
            +
            \coefb{\ell}
        },
    \end{equation*}
    where $\msettext{\cdot}$ denotes a multiset.
\end{definition}

Once the features are computed, we can apply a threshold to produce a classification
of nodes. The GNN assigns each node to one of the classes $\set{1, \ldots, \gnndim{L}}$ based on its final-layer features.

\begin{definition}[Classifier induced by a GNN]
    \label{def:gnn-semantics}
    Let $\cA$ be an $L$-layer GNN and
    $\agraph\in\frG_{V,\ifeatmap}$.
    For a vertex $v \in V$,
    the \emph{predicted class} of $\cA$ on $v$, denoted by $\pred{\agraph, v}$, is
    \begin{equation*}
        \pred{\agraph, v}
        \ :=\ 
        \argmax_{1 \le i \le \gnndim{L}} \left(\feat{L}_\agraph(v)\right)[i].
    \end{equation*}
\end{definition}

\subsubsection{Adversarial Robustness of GNNs}
Unlike standard feedforward neural networks, whose features and predictions depend only on the input features;
GNN computation depends on both the input features and the structure of the input graph.
Consequently, GNN features and predictions may change under two types of
perturbations~\cite{BojchevskiG19,HojnyZCM24,martaminghaoch}:
(i) \emph{feature perturbations}, where an adversary modifies node features,
and (ii) \emph{structural perturbations}, where an adversary inserts or deletes edges.
In this work we study adversarial robustness under \emph{structural perturbations only}:
given a GNN $\cA$, a featured graph $\cG$, and an admissible perturbation budget,
we ask whether the prediction of $\cA$ remains unchanged for \emph{every} graph within the admissible perturbation margin around $\cG$.

We formalize admissible perturbations via an \emph{admissible perturbation space of graphs},
following prior work~\cite{BojchevskiG19,HojnyZCM24,martaminghaoch}.
We restrict to structural perturbations --- edge insertions and deletions ---
while keeping node features fixed.

\begin{definition}[Admissible perturbation space of a graph]\label{def:admitperturbspace}
    Given a featured graph $\agraph = \tuple{V, E, \ifeatmap}$,
    a set of fragile edges $F \subseteq V \times V$,
    a global perturbation budget $\Delta$,
    and a local perturbation budget $\delta$,
    the \emph{admissible perturbation space} $\frQ(\agraph, F, \Delta, \delta)$ of $\agraph$ with respect to $F$, $\Delta$, and $\delta$
    is the set of graphs $\tuple{V, E', \ifeatmap}$ that satisfy:
    \begin{enumerate}
        \item $E \backslash F \subseteq E' \subseteq E \cup F$.
        \item $\abs{E \backslash E'} + \abs{E' \backslash E} \le \Delta$.
        \item For every $v \in V$,
            $\abs{\nbr_\agraph(v) \backslash \nbr_{\agraph'}(v)} + \abs{\nbr_{\agraph'}(v) \backslash \nbr_\agraph(v)} \le \delta$.
    \end{enumerate}
\end{definition}
The conditions above are equivalent to requiring that $\agraph'$ can be obtained from $\agraph$
by converting at most $\Delta$ edges from $F$.
Moreover, for each $v \in V$, at most $\delta$ of its incident edges are converted.
We also consider the admissible perturbation space without local budget limitations.
In this case, we simply write $\frQ(\agraph, F, \Delta)$.

The choice of the fragile-edge set $F$ captures different scenarios.
For example, in the \emph{deletion-only} case,
setting $F = E$ allows the adversary to delete existing edges but not insert new ones;
if we set $F = (V \times V) \setminus \{(v,v) : v \in V\}$,
the adversary may insert or delete any non-self-loop edge.

We now formalize the notion of adversarial robustness for GNNs.
\begin{definition}[Adversarial robustness of GNNs]
    Let $\cA$ be an $L$-layer GNN,
    $\agraph$ a featured graph,
    $F \subseteq V \times V$ a set of fragile edges,
    $\Delta \in \bbN$ a global budget,
    and $\delta \in \bbN$ a local budget.
    For a class $1 \le c \le \gnndim{L}$,
    we define the \emph{adversarial robustness} of $\cA$ with respect to $c$ as follows:

    For a vertex $v \in V$,
    we say that $\cA$ is \emph{adversarially robust} for $v$ with class $c$,
    (w.r.t. $F$, $\Delta$, and $\delta$)
    if, for every perturbed graph $\agraph' \in \frQ(\agraph, F, \Delta, \delta)$,
    $\pred{\agraph', v} = c$.
    When the limitation on local budget $\delta$ is dropped,
    the perturbed graph $\agraph'$ is taken from $\frQ(\agraph, F, \Delta)$.
\end{definition}
Note that the condition in the above definition is equivalent to requiring that,
for every $1 \le c' \neq c \le \gnndim{L}$,
$\feat{L}_{\agraph'}(v)[c]\ \ge\ \feat{L}_{\agraph'}(v)[c']$.
This formulation requires robustness against \emph{every} alternative class $c' \neq c$.
These problems are easily seen to be NP-complete even for a fixed GNN and a very simple update model: see Appendix~\ref{app:missing-proof} for a precise statement and a proof, which follows along the lines of similar results for feedforward networks \cite{reluplex17,langenpcomplete21}.

\begin{definition}[Adversarial robustness radius of GNNs]
    Let $\cA$ be an $L$-layer GNN,
    $\agraph$ a featured graph, and
    $F \subseteq V \times V$ a set of fragile edges.
    For a class $1 \le c \le \gnndim{L}$,
    for a vertex $v \in V$,
    the \emph{adversarial robustness radius} of $\cA$ for $v$ with class $c$
    is the maximum global budget $\Delta$
    such that
    $\cA$ is adversarial robust for $v$ with class $c$ (w.r.t. $F$ and $\Delta$).
\end{definition}



\section{Lightweight robustness analysis graph neural networks} \label{sec:oracle}

We overview our approach to robustness analysis.
Consider the following generalization of the robustness verification problem:

\begin{definition}[$d$-radius satisfaction]\label{prob:dpenaltysat}
    Given property $\query$,
    a normal graph $\agraph\in \frG_{V, \ifeatmap}$,
    an incomplete graph $\incgraph \in \frH_{V, \ifeatmap}$,
    and a vertex $v \in V$,
    we say that property $\query$ is \emph{satisfied within radius $d$ of
    $\agraph$ at $v$ with respect to $\incgraph$}
    if there is a completion $\agraph'$ of $\incgraph$ with
    $\dist{\agraph', \agraph} \le \dist{\incgraph, \agraph}+d$
    and $\tuple{\agraph', v}$ satisfies $\query$.
\end{definition}

We will also consider the following search problem variant.

\begin{definition}[Radius of satisfaction]
    Given property $\query$,
    a normal graph $\agraph\in \frG_{V, \ifeatmap}$,
    an incomplete graph $\incgraph \in \frH_{V, \ifeatmap}$,
    a vertex $v \in V$,
    and a maximum budget $d_m \in \bbN$,
    the \emph{radius of satisfaction} of $\query$ of $\cG$ at $v$ with respect to $\incgraph$ is
    the largest $d \le d_m$, such that
    for every completion $\agraph'$ of $\incgraph$ with $\dist{\agraph', \agraph} \leq d$,
    $\tuple{\agraph', v}$ does not satisfy $\query$.
\end{definition}

Assuming membership in the property $\query$ is decidable,
we can decide the problems defined above,
since a naive algorithm can enumerate
completions of $\incgraph$,
and check whether they satisfy $\query$ and the distance constraint.
Rather than using a na\"ive algorithm,
we aim to solve these problems
by employing a \emph{partial oracle}.

\begin{definition}[Partial oracle]
    For a node property $\query$,
    a normal graph $\agraph \in \frG_{V, \ifeatmap}$,
    and a vertex $v \in V$,
    a \emph{partial oracle for $\query$, $\agraph$, and $v$}, denoted by $\oracle_{\query, \agraph, v}$,
    is a function that receives an incomplete graph $\incgraph \in \frH_{V, X}$ and a budget $d \in \bbN$ as input,
    and which outputs either $\sat$, $\unsat$, or $\unknowntext$, and satisfy the following conditions:
    \begin{compactenum}
        \item (Correctness for $\sat$)
            If $\oracle_{\query, \agraph, v}(\incgraph, d)$ returns $\sat$,
            then there is a completion $\agraph'$ of $\incgraph$ with $\dist{\agraph', \agraph} \le \dist{\incgraph, \agraph} + d$
            and $\tuple{\agraph', v}$ satisfies $\query$.
        \item (Correctness for $\unsat$)
            If $\oracle_{\query, \agraph, v}(\incgraph, d)$ returns $\unsat$,
            then for each completion $\agraph'$ of $\incgraph$ with $\dist{\agraph', \agraph} \le \dist{\incgraph, \agraph} + d$,
            $\tuple{\agraph', v}$ does not satisfy $\query$.
        \item (Soundness on normal graphs or zero budget)
            If $\incgraph$ is normal or $d = 0$,
            then $\oracle_{\query, \agraph, v}(\incgraph, d)$ never returns $\unknowntext$.
    \end{compactenum}
\end{definition}

We can solve the $d$-radius satisfaction problem
by the following depth-first search (DFS) Algorithm~\ref{algo:dfs},
which searches through completions of $\incgraph$ for boosting a partial oracle into an exact solution.
The search procedure is guided by the partial oracle $\oracle_{\query, \agraph, v}$,
that is, if the oracle returns either $\sat$ or $\unsat$,
then the search procedure returns from the current branch with $\sat$ or $\unsat$ immediately.
If the oracle returns $\unknowntext$, the algorithm converts an unknown edge to either a non-edge or a normal edge and proceeds recursively.
The correctness can be established by induction on the number of unknown edges.

The runtime of Algorithm~\ref{algo:dfs} is exponential in the number of unknown edges in $\incgraph$ in the worst case.
Since the partial oracle is complete for normal graphs,
when the incomplete graph $\incgraph$ is normal,
it will return $\sat$ at line 4 or $\unsat$ at line 6.
The partial oracle will only return $\unknowntext$ when the incomplete graph $\incgraph$ is not normal.
The graphs $\incgraph_1$ and $\incgraph_2$ are obtained by converting an unknown edge in $\incgraph$ into a normal edge or a non-edge, respectively, which implies that the number of unknown edges in $\incgraph_1$ or $\incgraph_2$ is one less than in $\incgraph$. Thus the number of recursive calls is bounded by $2^{|E_{\unknown}|}$.

Designing a higher-quality oracle can significantly reduce the number of oracle calls compared to the na\"ive approach.
However, there is always a trade-off between the quality of the oracle and its cost: an exact oracle can be obtained by exhaustively checking all exponentially many normal graphs, whereas a trivial sloppy oracle may simply return $\unknowntext$ for every non-normal graph.

\begin{algorithm}[ht!]
    \caption{Algorithm for solving the $d$-radius satisfaction problem.}\label{algo:dfs}
    \begin{algorithmic}[1]
        \Procedure{Check}{$\query$, $\agraph$, $v$, $\incgraph$, $d$}
        \State $\oracleres \gets \oracle_{\query, \agraph, v}(\incgraph, d)$
        \If{$\oracleres$ is $\sat$}
        \State \Return $\sat$
        \ElsIf{$\oracleres$ is $\unsat$}
        \State \Return $\unsat$
        \Else
        \State Pick $e \in E_\unknown$
        \If{$e$ is an edge in $\agraph$}
        \State $\incgraph_1 \gets \incgraph$ by converting $e$ into a normal edge.
        \State $\incgraph_2 \gets \incgraph$ by converting $e$ into a non-edge.
        \Else
        \State $\incgraph_1 \gets \incgraph$ by converting $e$ into a non-edge.
        \State $\incgraph_2 \gets \incgraph$ by converting $e$ into a normal edge.
        \EndIf
        \If{(\Call{Check}{$\query$, $\agraph$, $v$, $\incgraph_1$, $d$} \Return $\sat$) or (\Call{Check}{$\query$, $\agraph$, $v$, $\incgraph_2$, $d-1$} \Return $\sat$)}
        \State \Return $\sat$
        \Else
        \State \Return $\unsat$
        \EndIf
        \EndIf
        \EndProcedure
    \end{algorithmic}
\end{algorithm}

We will solve the adversarial robustness problem for GNNs via Algorithm~\ref{algo:dfs},
using a polynomial time partial oracle.
We focus on node classification under an unbounded local budget:
no per-vertex constraint on the number of perturbed edges.
Once we have obtained a partial oracle for the $d$-radius satisfaction problem,
we could apply it na\"ively to compute the distance for satisfaction,
which corresponds to the adversarial robustness radius problem.

We fix an $L$-layer GNN $\cA$, a graph $\agraph$ with set of vertices $V$ and feature mapping $X$,
a vertex $v_0 \in V$,
a set of fragile edges $F \subseteq V \times V$,
a global budget $\Delta$,
and a class $1 \le c \le \gnndim{L}$.

Consider the relaxation of $\agraph$ with respect to $F$,
which is denoted by $\incgraph_{\agraph,F}$.
Recall that $\incgraph_{\agraph,F}$ is the incomplete graph obtained by
converting all the edges and non-edges in $F$ into unknown edges.
By Lemma~\ref{lemma:relax}, we have $\dist{\incgraph_{\agraph,F}, \agraph} = 0$.
We first observe that the admissible perturbation space $\frQ(\agraph, F, \Delta)$
coincides with the set of completions $\agraph'$ of $\incgraph_{\agraph,F}$
satisfying
$\dist{\agraph', \agraph} \le \Delta$.

Let $\cQ_{\agnn, c}$ be the node property that holds on node $v_0$ in $\agraph'$
when there exist $1 \leq c' \neq c \leq \gnndim{L}$ such that
$\feat{L}_{\agraph'}(v_0)[c]\ <\ \feat{L}_{\agraph'}(v_0)[c']$.
If there exists a perturbed graph $\agraph' \in \frQ(\agraph, F, \Delta)$,
such that $\tuple{\agraph', v_0}$ satisfies $\cQ$,
then $\agnn$ is not adversarially robust for $v_0$ with class $c$.
Thus, verifying the \emph{non-adversarial robustness} of $\agnn$ for $v_0$ with class $c$
can be reduced to solving the $d$-radius satisfaction problem as follows:
\begin{quote}
    Does there exist a normal graph $\agraph' \subseteq \incgraph_{\agraph,F}$
    with $\dist{\agraph', \agraph} \le \Delta$
    such that $\tuple{\agraph', v}$ satisfies $\cQ_{\agnn,c}$?
\end{quote}

We next present a polynomial time partial oracle for the node property $\cQ_{\agnn,c}$.
We will describe it for general incomplete graphs $\incgraph$,
not just for $\incgraph_{\agraph,F}$.
The partial oracle consists of two sequential components: a non-robustness tester, and a bound propagator.
It returns $\sat$ (resp. $\unsat$) if any of the components returns $\sat$ (resp. $\unsat$).
Otherwise, it returns $\unknowntext$.

\subsubsection{Non-robustness tester}
The non-robustness tester evaluates the grounding $\agraph'$ of $\incgraph$ with respect to $\agraph$
and returns $\sat$ if the grounding $\agraph'$ satisfies $\cQ_{\agnn,c}$;
otherwise, it returns $\unknowntext$.
Recall that the grounding is the completion of $\incgraph$ that is closest to $\agraph$.
By Lemma~\ref{lemma:ground},
we have $\dist{\agraph', \agraph} = \dist{\incgraph, \agraph} \le \dist{\incgraph, \agraph} + d$ for any $d \in \bbN$.
Therefore, if $\agraph'$ satisfies $\cQ_{\agnn,c}$,
then $\agraph'$ is a non-robust normal graph within the admissible perturbation space.

\subsubsection{Bound propagator}
For the bound propagator, we abstract the computation of a GNN on an incomplete graph by computing
upper and lower bounds for features
at each vertex $v$ and each layer $\ell$: we
denote these by $\featu{\ell}_{\incgraph}(v)$ and $\featl{\ell}_\incgraph(v)$. The bounds are computed
in a bottom-up manner, with the correctness condition being that for every completion $\agraph'$ of $\incgraph$,
\begin{equation*}
    \featl{\ell}_\incgraph(v)
    \ \le\ 
    \feat{\ell}_{\agraph'}(v)
    \ \le\ 
    \featu{\ell}_{\incgraph}(v).
\end{equation*}
Note that vector comparisons are performed entrywise.
After computing the over-approximated bounds,
we check that,
for every $1 \le c' \neq c \le \gnndim{L}$,
\begin{equation*}
    \featu{L}_{\incgraph}(v_0)[c]
    \ <\ 
    \featl{L}_{\incgraph}(v_0)[c'].
\end{equation*}
If this condition holds, then for every completion $\agraph'$ of $\incgraph$,
we have
\begin{equation*}
    \feat{L}_{\agraph'}(v_0)[c]
    \ \le\ 
    \featu{L}_{\incgraph}(v_0)[c]
    \ <\ 
    \featl{L}_{\incgraph}(v_0)[c']
    \ \le\ 
    \feat{L}_{\agraph'}(v_0)[c'],
\end{equation*}
which implies that $\tuple{\agraph', v_0}$ do not satisfy $\cQ_{\agnn,c}$.

We need functions for capturing bound propagation for matrix multiplication and aggregation functions.
First, for $\bfA \in \bbR^{m \times n}$ and $\bfvu, \bfvl \in \bbR^n$, 
we define
\begin{equation*}
    \relaxu(\bfA, \bfvu, \bfvl)
    \ :=\ 
    \bfA^+ \cdot \bfvu + \bfA^- \cdot \bfvl
    \quad \text{and} \quad
    \relaxl(\bfA, \bfvu, \bfvl)
    \ :=\ 
    \bfA^+ \cdot \bfvl + \bfA^- \cdot \bfvu,
\end{equation*}
where $\bfA^+, \bfA^- \in \bbR^{m \times n}$
are defined entrywise by
$\bfA^+ := \max\left(\bfA, 0\right)$
and
$\bfA^- := \min\left(\bfA, 0\right)$.
These are lower and upper approximations for matrix multiplication, as captured in the following lemma:
\begin{lemma}\label{lem:matrix}
    For every $\bfA \in \bbR^{m \times n}$ and $\bfvu, \bfv, \bfvl \in \bbR^n$
    with $\bfvl \le \bfv \le \bfvu$,
    \begin{equation*}
        \relaxl\left(\bfA, \bfvu, \bfvl\right)
        \ \le\ 
        \bfA \cdot \bfv
        \ \le\ 
        \relaxu\left(\bfA, \bfvu, \bfvl\right).
    \end{equation*}
\end{lemma}

Next, we define
approximations for the aggregation functions.

\begin{definition}
    Let $S_1$ and $S_2$ be multisets of reals.
    \begin{compactitem}
    \item For $\summ$ aggregation, we define
        \begin{equation*}
            \begin{aligned}
                \sumu\left(S_1, S_2\right) \ :=\ &
                \sum_{s \in S_1} s + \sum_{s \in S_2} \max(s, 0)
                \\
                \suml\left(S_1, S_2\right)\ :=\ &
                \sum_{s \in S_1} s + \sum_{s \in S_2} \min(s, 0).
            \end{aligned}
        \end{equation*}
    \item For $\maxx$ aggregation, we define
        $\maxu\left(S_1, S_2\right) := \max\left(S_1 \cup S_2\right)$.
        If $S_1 = \emptyset$, then we set
        $\maxl\left(S_1, S_2\right) := \min\left(0, \min\left(S_2\right)\right)$;
        otherwise
        $\maxl\left(S_1, S_2\right):= \max\left(S_1\right)$.
        For convention, we let $\max(\emptyset) = 0$.
    \item For $\mean$ aggregation,
        let $s_1, \ldots, s_{\abs{S_2}}$ be the elements of $S_2$ arranged in descending order.
        We define
        \begin{equation*}
            \begin{aligned}
                \meanu\left(S_1, S_2\right)
                \ :=\ &
                \max_{1 \le i \le \abs{S_2}}\left(
                    \left(\sum_{s \in S_1} s + \sum_{1 \le j \le i}s_j \right) /
                    \left(\abs{S_1} + i\right)
                \right)
                \\
                \meanl\left(S_1, S_2\right)
                \ :=\ &
                \max_{1 \le i \le \abs{S_2}}\left(
                    \left(\sum_{s \in S_1} s + \sum_{\abs{S_2} - i \le j \le \abs{S_2}}s_j \right) /
                    \left(\abs{S_1} + i\right)
                \right).
            \end{aligned}
        \end{equation*}
    \end{compactitem}
\end{definition}
We now define the over-approximated upper and lower bounds for GNN computation.
For $\ell = 0$, $\featu{0}_\incgraph(v) = \featl{0}_\incgraph(v) := X(v)$.
For $1 \le \ell \le L$ and $v \in V$,
\begin{equation*}
    \begin{aligned}
        \featu{\ell}_\incgraph(v)
        \ :=\ &
        \relup{
            \relaxu\left(\coefC{\ell}, \featu{\ell-1}_\incgraph(v), \featl{\ell-1}_\incgraph(v)\right) +
            \relaxu\left(\coefA{\ell}, \bfsu, \bfsl\right) +
            \coefb{\ell}
        }
        \\
        \featl{\ell}_\incgraph(v)
        \ :=\ &
        \relup{
            \relaxl\left(\coefC{\ell}, \featu{\ell-1}_\incgraph(v), \featl{\ell-1}_\incgraph(v)\right) +
            \relaxl\left(\coefA{\ell}, \bfsu, \bfsl\right) +
            \coefb{\ell}
        },
    \end{aligned}
\end{equation*}
where
\begin{equation*}
    \bfsu\ :=\ \aggru\left( \setsu^\norm, \setsu^\unknown \right)
    \quad \text{and} \quad
    \bfsl\ :=\ \aggrl\left( \setsl^\norm, \setsl^\unknown \right),
\end{equation*}
$\setsu^\norm := \msetc{\featu{\ell-1}_\incgraph(u)}{u \in \nbr^\norm_\incgraph(v)}$
and $\setsl^\norm$, $\setsu^\unknown$, $\setsl^\unknown$ defined analogously.

The upper and lower bounds are computed in a bottom-up manner.
For each layer $\ell$ and each vertex $v$:
\begin{compactitem}
    \item With $\summ$ and $\maxx$ aggregations,
        the bounds can be obtained in time linear in the number of neighbors of $v$.
    \item With $\mean$ aggregation,
        the bounds are computed by first sorting the neighbor bounds from the previous layer, with this sorting step dominating the overall cost.
\end{compactitem}
Combining these results, the total time complexity is
$O\left(L\abs{V}^2\right)$ for $\summ$ and $\maxx$, and
$O\left(L\abs{V}^2 \log \abs{V}\right)$ for $\mean$,
where $\abs{V}$ denotes the number of vertices in the graph and $L$ denotes the number of GNN layers.
The correctness of the over-approximated upper and lower bounds can be proven inductively.

\begin{lemma}\label{lem:boundcorrect}
    For every completion $\agraph'$ of $\incgraph$,
    $0 \le \ell \le L$,
    and vertex $v \in V$,
    \begin{equation*}
        \featl{\ell}_\incgraph(v)
        \ \le\ 
        \feat{\ell}_{\agraph'}(v)
        \ \le\ 
        \featu{\ell}_{\incgraph}(v).
    \end{equation*}
\end{lemma}


\section{Optimization}
\label{sec:opt}

We discuss several optimizations of the na\"ive algorithm.
Some are based on the architecture of GNNs, while others leverage the structure of the input graph.

\subsubsection{Incremental Computation for the Partial Oracle}
Recall that the partial oracle for adversarial robustness of GNNs, defined previously
consists of two components: the non-robustness tester and the bound propagator.
Both components compute features or their over-approximated bounds in a bottom-up manner,
where each value depends on the features or bounds of its neighbors from the previous layer.
This process involves $\abs{V} \cdot L$ computations in total.

The main idea for speeding up the computation for the partial oracle is therefore to cache the features and bounds of all vertices at all layers for the current recursive call.
During recursion, we only update values for the vertices affected by the edge operation,
while reusing previously stored results for unaffected vertices.
When returning from a recursive call, the cache of the modified features and bounds is restored to its prior state.

For example, suppose we have computed the features and bounds for the current graph,
and then convert the edge $(v, u)$ in the graph, either from non-edge to edge, or vice versa.
\begin{itemize}
    \item (Layer $0$.)
        The features are given by $X$, which are independent of the graph structure,
        so no updates are required.

    \item (Layer $1$.)
        Only $v$ and $u$ need to update their features and over-approximated bounds.

    \item (Layer $2$.)
        In addition to $v$ and $u$,
        their neighbors $\nbr^\norm_\cH(v) \cup \nbr^\unknown_\cH(v)$ and $\nbr^\norm_\cH(u) \cup \nbr^\unknown_\cH(u)$ must also be updated.

    \item (Layer $\ell$.)
        At layer $\ell$,
        we updates vertices which are updated in layer $\ell-1$ and their neighbors.
        In general, we update all vertices within distance at most $\ell-1$ from either $v$ or $u$.
\end{itemize}

Furthermore
if at any layer we detect that a vertex $w$'s feature or bound does not change,
then its neighbors need not be updated in the next layer.

We can further reduce the number of updates by tracking the distance between a vertex
$w$ and the target vertex $v_0$.
Recall that the outcome of node classification depends on the feature of $v_0$ at the last layer $L$.
For a vertex $w$ with distance $d$ to $v_0$,
for any $\ell > L-d$, the $\ell^{th}$ layer feature of $w$ will \emph{not} affect the final feature of $v_0$.
Therefore, at the beginning of updating the $\ell^{th}$,
we can prune nodes that are farther than $L - \ell$ from $v_0$.
These pruning avoids unnecessary recomputation and further improves efficiency.

\subsubsection{Reordering Operations}

In the computation of vertex features, aggregation over neighbors is followed by a matrix multiplication.
For GNNs with $\summ$ or $\mean$ aggregations, the order of these operations can be exchanged,
that is
\begin{equation*}
    \begin{aligned}
        \coefA{\ell} \cdot \summ\left(\msetc{\feat{\ell-1}_\cG(u)}{u \in \nbr_\cG(v)}\right)
        \ =\ &
        \summ\left(\msetc{\coefA{\ell} \cdot \feat{\ell-1}_\cG(u)}{u \in \nbr_\cG(v)}\right) \\
        \coefA{\ell} \cdot \mean\left(\msetc{\feat{\ell-1}_\cG(u)}{u \in \nbr_\cG(v)}\right)
        \ =\ &
        \mean\left(\msetc{\coefA{\ell} \cdot \feat{\ell-1}_\cG(u)}{u \in \nbr_\cG(v)}\right).
    \end{aligned}
\end{equation*}

This reordering is trivial for non-robustness tester.
However, it improves efficiency of bound propagator
by allowing us to shrink the over-approximated upper and lower bounds.
Intuitively, over-approximated upper bounds for $\aggr$ are computed for each entry by selecting certain neighbors.
If we perform aggregation first, each entry is chosen independently,
and the resulting bounds are later obtained through matrix multiplication.
But we know that once a neighbor is chosen for the $i^{th}$ entry,
the corresponding $j^{th}$ entry is also determined.
By performing the matrix multiplication first,
we preserve this dependency between entries, which leads to tighter bounds.

Formally, let $\bfA \in \mathbb{R}^{m \times n}$, and let
$\setsu_1 = \msettext{\bfsu_{1, 1}, \ldots, \bfsu_{1, k_1}}$
$\setsl_1 = \msettext{\bfsl_{1, 1}, \ldots, \bfsl_{1, k_1}}$,
$\setsu_2 = \msettext{\bfsu_{2, 1}, \ldots, \bfsu_{2, k_1}}$, and
$\setsl_2 = \msettext{\bfsl_{2, 1}, \ldots, \bfsl_{2, k_1}}$
be multisets of real vectors in $\bbR^n$.
Then we have the bound relaxation for the usual order:
\begin{equation*}
    \begin{aligned}
        &
        \relaxu\left(
            \bfA,
            \sumu\left(\setsu_1, \setsu_2\right),
            \suml\left(\setsl_1, \setsl_2\right)
        \right) \\
        \ =\ &
        \relaxu\left(
            \bfA,
            \sum_{1 \le i \le k_1} \bfsu_{1, i} +
            \sum_{1 \le i \le k_2} \max\left(\bfsu_{2, i}, 0\right),
            \sum_{1 \le i \le k_1} \bfsl_{1, i} +
            \sum_{1 \le i \le k_2} \min\left(\bfsl_{2, i}, 0\right)
        \right) \\
        \ =\ &
        \sum_{1 \le i \le k_1} \left(
            \bfA^+ \cdot \bfsu_{1, i} + \bfA^- \cdot \bfsl_{1, i}
        \right) +
        \sum_{1 \le i \le k_2} \left(
            \bfA^+ \cdot \max\left(\bfsu_{2, i}, 0\right) +
            \bfA^- \cdot \min\left(\bfsl_{2, i}, 0\right)
        \right).
    \end{aligned}
\end{equation*}

On the other hand, if we apply matrix multiplication before aggregation, we obtain
\begin{equation*}
    \begin{aligned}
        &
        \sumu\left(
            \setc{\relaxu\left( \bfsu_{1, i}, \bfsl_{1, i}\right)}{1 \le i \le k_1},
            \setc{\relaxu\left( \bfsu_{2, i}, \bfsl_{2, i}\right)}{1 \le i \le k_2}
        \right) \\
        \ =\ &
        \sum_{1 \le i \le k_1} \left(
            \bfA^+ \cdot \bfsu_{1, i} + \bfA^- \cdot \bfsl_{1, i}
        \right) +
        \sum_{1 \le i \le k_2}
        \max\left(
            \left(\bfA^+ \cdot \bfsu_{2, i} + \bfA^- \cdot \bfsl_{2, i}\right), 0
        \right).
    \end{aligned}
\end{equation*}

Since $\bfA^+$ are positive, and $\bfA^-$ are negative entrywise,
we have that for each $1 \le i \le m$:
\begin{equation*}
    \begin{aligned}
        \bfA^+ \cdot \max\left(\bfsu_{2, i}, 0\right) +
        \bfA^- \cdot \min\left(\bfsl_{2, i}, 0\right)
        \ \ge\ &
        \bfA^+ \cdot \bfsu_{2, i} + \bfA^- \cdot \bfsl_{2, i} \\
        \bfA^+ \cdot \max\left(\bfsu_{2, i}, 0\right) +
        \bfA^- \cdot \min\left(\bfsl_{2, i}, 0\right)
        \ \ge\ &0.
    \end{aligned}
\end{equation*}
Therefore, the reordered computation yields a tighter (smaller) upper bound.

The arguments for the lower bound and for $\mean$ aggregation are analogous.

We can further reduce the number of computations by incorporating operator reordering and incremental computation,
caching the feature and bounds at each layer before computing the next layer.
For example, suppose that at layer $\ell$, we update only a vertex $v$.
According to the setup of incremental computation,
we then need to update all $k$ neighbors of $v$ at layer $\ell+1$.
In the normal computation order, we would perform aggregation and matrix multiplication $k$ times.
However, with operator reordering,
we can first compute the matrix multiplication between the coefficients and the updated feature of $v$ at layer $\ell$;
then, in the next step, we only need to apply aggregation $k$ times,thereby saving $k-1$ matrix multiplications.

\subsubsection{Bound Tightening of Aggregation Functions Using Budgets}

The bound propagator described earlier computes over-approximated bounds for \emph{all} completions of the input incomplete graph $\cH$.
However, when the perturbation budget is limited,
these bounds can be further tightened by explicitly incorporating the budget constraint.

For example,
consider a GNN with $\summ$ aggregation, where the global perturbation budget is $2$.
Suppose vertex $v$ has $5$ unknown incoming edges, all of which are normal edges in the target graph $\cG$.
Under the na\"ive propagation rule given earlier, the lower bound is obtained by summing all negative lower bounds from these $5$ neighbors.
If all neighbor features are positive, this procedure yields a trivial lower bound of $0$.
Now note that, because the budget is $2$, at least $3$ of the $5$ unknown edges must remain present.
Therefore, instead of discarding all $5$ contributions,
the lower bound should be given by the sum of the $3$ smallest neighbor features.
Since all features are positive, this bound is strictly greater than $0$.
This example shows how budget constraints can significantly tighten the propagated bounds.

The budget-aware bound tightening for $\summ$ aggregation was established in~\cite{HojnyZCM24}.
The corresponding results for $\maxx$ and $\mean$ aggregations were shown by~\cite{martaminghaoch}.

Note that, unlike reordering operations, bound shrinking can be achieved without additional overhead.
For bound tightening of aggregation functions using budgets,
however, extra computation (e.g. sorting) is required to determine the bounds based on the budget,
compared to the na\"ive algorithms.
Although the tighter bounds may reduce the number of oracle calls,
the computation time per oracle call increases.
As a result, there is no guarantee of a decrease in the overall runtime.

\subsubsection{Graph Structure Heuristics for Edge Picking in the High-level algorithm}
Thus far we have focused on optimizations within the lightweight oracles. In the high-level Algorithm~\ref{algo:dfs} that uses these oracles, graph structure can be exploited to accelerate termination. The high-level algorithm is order-sensitive: the amount of branch pruning depends heavily on which unknown edge is selected next (cf. line~8).
We outline several practical heuristics; the high-level idea is to pick the edge whose resolution is expected to affect most amount of features.

\begin{itemize}
    \item
        For node classification,
        for the target vertex $v_0$, prioritize unknown edges that are \emph{closest} to $v_0$.
        If an edge is at distance $r$ from $v_0$, it can influence the feature of $v_0$ only starting from layer $L-r$, where $L$ is the GNN depth.
        Choosing small-$r$ edges typically maximizes the impact on the final-layer feature of $v_0$
        and tends to minimize the connected region that contains $v_0$.

    \item
        For node classification,
        if the incomplete graph $\cH$ decomposes into several disconnected components,
        restrict edge choices to the component that contains the target vertex $v_0$.
        Edges outside this component cannot affect the feature of $v_0$ and only enlarge the search space.

    \item
        After picking the unknown edge $e$,
        we recursively consider two subproblems with inputs $\cH_1$ and $\cH_2$
        by setting $e$ to match its status in the target graph $\cG$ for $\cH_1$,
        and to the opposite status for $\cH_2$.
        In general, the order in which we perform the recursive calls for $\cH_1$ and $\cH_2$
        does not affect the correctness of the algorithm, since we later consider the disjunction of results of both recursive calls (cf. line~16).
        However, the order does affect the runtime of the algorithm, because we can terminate early if the first call returns
        Our strategy is to explore $\cH_2$ first,
        since this branch immediately consumes one unit of budget, yielding a smaller search space.
        Intuitively, it is also more likely to change the features than $\cH_1$ branch (the one consistent with $\cG$).
\end{itemize}

\subsubsection{Edge inference for local budgets}
The previous optimization exploits the limited \emph{global} budget.
In the case of \emph{local} budgets, we can further infer the status of unknown edges whenever a vertex exhausts its local budget.

Consider the following example.
Suppose vertex $v$ has $5$ unknown incoming edges, all of which are normal edges in the target graph $\cG$, and its local budget is $2$.
If, after some recursive calls, two of these unknown edges have already been converted into non-edges, then the local budget of $v$ is exhausted.
Consequently, all remaining unknown edges incident to $v$ must be normal edges; otherwise, the resulting graph would not belong to the admissible perturbation space.
This inference reduces the number of unknown edges, and hence shrinks the search space for Algorithm~\ref{algo:dfs}.

For undirected input graphs, this reasoning further tightens the over-approximated bounds of neighboring vertices,
since an inferred edge is shared by both endpoints.

In practice, this inference procedure can be applied at the beginning of each recursive call, thereby propagating the consequences of local budget constraints throughout the graph.


\section{Experiments} \label{sec:exper}

\subsubsection{Implementation of $\toolname$ and Setup}
Our method is implemented as $\toolname$,
which supports GNN robustness problems with a variety of aggregation functions
($\summ$, $\maxx$, and $\mean$)
for both node classification and graph classification;
for both directed and undirected graphs;
and for both deletion-only as well as deletion and insertion perturbations.
$\toolname$ is implemented in C,
and all experiments were conducted using the version compiled with GCC 11.4.
Experiments were performed on a cluster with Intel Xeon Platinum 8268 CPU @ 2.90GHz
with AVX2 support enabled running CentOS 8.
Each instance was solved using a single thread with 8GB of RAM.
The time limit was set to 300s for node classification instances and 600s for graph classification instances.

\subsubsection{Datasets and Models}
We evaluate the performance of $\toolname$ on GNN models with various numbers of layers and aggregation functions,
built and trained using PyG (PyTorch Geometric) 2.6~\cite{pyg},
on the benchmarks: Cora, CiteSeer~\cite{CoraOld, CoraNew}, Cornell, Texas, and Wisconsin~\cite{hetro} for node classification,
and MUTAG and ENZYMES~\cite{MUTAG} for graph classification.
Note that Cornell, Texas, and Wisconsin are different from the citation-based benchmarks in that
they are \emph{heterophilic}: the existence of an edge between two nodes is not tightly connected to the node labels.
We summarize the information on benchmarks in Table~\ref{tab:benchmarks}.

\begin{table}[t]
\centering
\caption{Information for benchmarks,
where Degree denotes the average incoming degree of vertices.
For graph datasets,
\#Vertices and \#Edges represent the average number of vertices and edges per graph, respectively.}
\label{tab:benchmarks}
\resizebox{\textwidth}{!}{
\begin{tblr}{
        rowsep  = 0pt,
        colspec = {l|rrrrr|l|rrrrrr},
    }
\toprule
Dataset & \#Vertices & \#Edges & Degree & \#Features & \#Classes &
Dataset & \#Graphs & \#Vertices & \#Edges & Degree & \#Features & \#Classes \\
\midrule
Cora      & 2,708 & 5,429 & 2.00 & 1,433 & 7 & MUTAG     &   188 &  17.9 &  39.6 & 2.21 & 7 & 2 \\
CiteSeer  & 3,312 & 4,715 & 1.42 & 3,703 & 6 & ENZYMES   &   600 &  32.6 & 124.3 & 3.81 & 3 & 6 \\
Cornell   &   183 &   298 & 1.63 & 1,703 & 5 & \\
Texas     &   183 &   325 & 1.78 & 1,703 & 5 & \\
Wisconsin &   251 &   515 & 2.05 & 1,703 & 5 & \\
\bottomrule
\end{tblr}
}
\end{table}

All models are trained for $1000$ epochs with a learning rate of $0.001$ and a weight decay of $5 \times 10^{-5}$.
For node datasets, we randomly select $30\%$ of the nodes as the training set, $20\%$ as the validation set, and the remaining $50\%$ as the test set.
For graph datasets, we use $80\%$, $10\%$, and $10\%$ of the graphs for training, validation, and testing, respectively.
The dimensions of hidden layers are set to $32$ for node classification and $16$ for graph classification.
We conduct experiments on directed graphs with deletion-only perturbations for node classification;
that is, the set of fragile edges $F$ corresponds to the set of edges of the input graph.
For graph classification,
we perform both deletion and insertion perturbations on undirected graphs;
that is, the set of fragile edges $F$ includes all edges excluding self-loops.

Recall that the condition of the robustness of a vertex $v$ with respect to the class $c$ is that,
for \emph{every} $1 \le c' \neq c \le \gnndim{L}$,
$\feat{L}_{\agraph'}(v)[c]\ \ge\ \feat{L}_{\agraph'}(v)[c']$.
We also consider \emph{weak robustness},
in which we only consider perturbations to one fixed target class $c'$.
In our experiments, we set $c$ to be the predicted class by the GNN and $c' = \left(c \mod \gnndim{L}\right) + 1$.

\subsubsection{Baselines}
We compare our $\toolname$ with the most recent available exact tools for GNN robustness checking,
$\scip$~\cite{HojnyZCM24} and $\gnnev$~\cite{martaminghaoch},
both of which are based on translating the robustness problem into mixed-integer programming (MIP).
$\scip$ implements a solver for the weaker version of the GNN robustness problem,
supporting node classification for directed graphs with deletion-only perturbations,
and graph classification for undirected graphs with both deletion and insertion perturbations;
both use $\summ$ aggregation only.
It relies on the open-source MIP solver SCIP~\cite{scip}.
In our experiments, we run $\scip$ using the SCIPsbt setting for node classification
and the SCIPabt setting for graph classification.
Note that $\scip$ also provides options to solve the MIP instance with the commercial MIP solver Gurobi~\cite{gurobi}.
However, this implementation is buggy, as mentioned in Appendix B.2 of~\cite{HojnyZCM24},
and can incorrectly report a feasible instance as infeasible.
For consistency, we do not run $\scip$ with Gurobi.
$\gnnev$ implements a solver for the GNN robustness problem
for node classification on directed graphs with both deletion and insertion perturbation,
supporting aggregation functions $\summ$, $\maxx$, and $\mean$,
relying on Gurobi~\cite{gurobi}.
We run $\gnnev$ with incremental solving enabled.

\subsubsection{End to End Performance on Node Classification}
We first looked to answer the question of how our lightweight solve-based methods compares to the
state of the art on standard node classification benchmarks, focusing
on the Cora, CiteSeer, Cornell, Texas, and Wisconsin datasets.
The number of GNN layers is set to $4$, noting that no prior tool has shown consistent performance passed $3$ layers.
We apply the analysis on each vertex with budgets $1$, $2$, $5$, and $10$.
We summarize the results on weak robustness for the $\summ$ aggregation
as well as the results on general robustness for the $\summ$, $\maxx$, and $\mean$ aggregations
in Table~\ref{tab:node_simple},
where we sum the number of instances with different budgets.
For the shifted geometric mean, we set the shift to $10$.
For the full results for each distinguished budget, see Appendix~\ref{app:exp1}.
Figure~\ref{fig:node} gives a different view,
showing how many instances for each budget can be completed as time increases.

\paragraph{Takeaways} The first conclusion is that $\toolname$ outperforms the baselines by more than an order of magnitude for every set up. In particular, it shows that it can handle $4$ layer GNNs, which were beyond the scope of the prior art.
In the case of larger budgets, the competitors cannot complete a significant portion of the instances --- note that we are showing the average time only for completed instances.

In our algorithm we are also doing constraint-solving.
Our advantage is that we are using the structure of the GNN --- in each call to our partial oracle and in the optimizations (e.g. caching) of the high-level algorithm. This structure is not transparent to a constraint solver like Gurobi.

For MIP-based solvers, robust instances are easier than non-robust ones, roughly corresponding to unsat vs. sat. For $\toolname$, non-robust instances are easier, since on non-robust instances our na\"ive counterexample finder turns out to be sufficient.

To fairly compare weak robustness and general robustness,
we summarize the runtime of instances that are both weakly and generally robust (or non-robust)
with budget $10$ in Table~\ref{tab:node_weak}.
For instances that are both weakly and generally robust, the runtime increases for both
$\toolname$ and $\gnnev$, as expected.
However, for non-robust instances, compared with
$\gnnev$, whose runtime remains roughly the same,
the runtime for general robustness decreases significantly for
$\toolname$. This is again because our lightweight non-robust tester can quickly find counterexamples.

In terms of the impact of the aggregation function, we see that max is the hardest for $\toolname$, and this is because we cannot apply the re-ordering optimization.

\begin{table}[t]
\centering
\caption{Comparison results of $\toolname$, $\gnnev$, and $\scip$
for weak and general robustness on the Cora, CiteSeer, Cornell, Texas, and Wisconsin datasets with various aggregation functions.
Note that $\scip$ only implements weak robustness for $\summ$ aggregation.
$t_{a}$ denotes the average runtime, 
and $t_{g}$ denotes the shifted geometric mean of the runtime.}
\label{tab:node_simple}
\resizebox{\textwidth}{!}{
\begin{tblr}{
        rows    = {abovesep=0.2pt, belowsep=0.2pt},
        row{1}  = {font=\large, abovesep=3pt, belowsep=0pt},
        colspec = {l|l|r|rrrrrr|rrrrrr|rrrrrr},
        cell{1}{4}  = {c=6}{c},
        cell{1}{10} = {c=6}{c},
        cell{1}{16} = {c=6}{c},
        cell{2}{4}  = {c=3}{c},
        cell{2}{7}  = {c=3}{c},
        cell{2}{10} = {c=3}{c},
        cell{2}{13} = {c=3}{c},
        cell{2}{16} = {c=3}{c},
        cell{2}{19} = {c=3}{c}
    }
\toprule
& & & $\toolname$ & & & & & & $\gnnev$ & & & & & & $\scip$ \\
& & &
    All instances & & & Robust instances & & &
    All instances & & & Robust instances & & &
    All instances & & & Robust instances \\
\cmidrule[lr]{4-6}   \cmidrule[lr]{7-9}
\cmidrule[lr]{10-12} \cmidrule[lr]{13-15}
\cmidrule[lr]{16-18} \cmidrule[lr]{19-21}
& & \#Instances &
\#Solved & $t_{a}(s)$ & $t_{g}(s)$ &
\#Solved & $t_{a}(s)$ & $t_{g}(s)$ &
\#Solved & $t_{a}(s)$ & $t_{g}(s)$ &
\#Solved & $t_{a}(s)$ & $t_{g}(s)$ &
\#Solved & $t_{a}(s)$ & $t_{g}(s)$ &
\#Solved & $t_{a}(s)$ & $t_{g}(s)$ \\
\midrule
$\summ$
&      Cora & 10,832 & 10,812 &   0.36 &   0.12 &  9,132 &   0.32 &   0.13 &  7,509 &  27.13 &  11.55 &  6,101 &  27.36 &  10.57 &  8,912 &  53.63 &  19.91 &  7,641 &  49.09 &  17.35 \\ 
(Weak)
&  CiteSeer & 13,248 & 13,222 &   0.13 &   0.04 & 11,930 &   0.14 &   0.05 & 12,193 &  14.70 &   7.58 & 10,936 &  13.77 &   6.73 & 12,935 &  13.06 &   4.17 & 11,720 &  11.32 &   3.52 \\ 
&   Cornell &    732 &    732 &   0.01 &   0.01 &    413 &   0.01 &   0.01 &    627 &  15.68 &   8.37 &    331 &  16.46 &   7.36 &    691 &  25.94 &   7.66 &    411 &  22.38 &   6.45 \\ 
&     Texas &    732 &    731 &   0.72 &   0.22 &    578 &   0.77 &   0.20 &    611 &  14.21 &   6.19 &    494 &   9.28 &   3.95 &    650 &  26.52 &   7.97 &    537 &  15.96 &   4.46 \\ 
& Wisconsin &  1,004 &    992 &   1.17 &   0.26 &    815 &   1.42 &   0.32 &    768 &  18.26 &   8.92 &    626 &  15.62 &   7.11 &    844 &  38.93 &  12.58 &    723 &  34.13 &  10.79 \\
\midrule
$\summ$
&      Cora & 10,832 & 10,826 &   0.22 &   0.07 &  6,824 &   0.27 &   0.08 &  8,122 &  23.27 &  11.11 &  4,741 &  17.98 &   7.17 \\ 
&  CiteSeer & 13,248 & 13,228 &   0.13 &   0.04 & 10,216 &   0.17 &   0.05 & 12,367 &  15.20 &   8.09 &  9,439 &  13.33 &   6.34 \\ 
&   Cornell &    732 &    732 &   0.01 &   0.01 &    327 &   0.01 &   0.01 &    675 &  14.39 &   8.54 &    292 &  15.18 &   6.76 \\ 
&     Texas &    732 &    732 &   0.01 &   0.01 &    305 &   0.02 &   0.02 &    655 &  10.42 &   6.09 &    273 &   5.96 &   3.17 \\ 
& Wisconsin &  1,004 &  1,001 &   0.30 &   0.13 &    645 &   0.42 &   0.18 &    853 &  22.11 &  10.58 &    547 &  14.81 &   6.57 \\
\midrule
$\maxx$
&      Cora & 10,832 & 10,813 &   0.36 &   0.14 &  7,096 &   0.51 &   0.19 &  7,092 &  93.64 &  29.53 &  4,394 &  55.68 &  13.75 \\ 
&  CiteSeer & 13,248 & 13,206 &   0.26 &   0.09 & 10,303 &   0.32 &   0.11 & 11,020 &  71.44 &  19.23 &  8,512 &  45.48 &  12.38 \\ 
&   Cornell &    732 &    732 &   0.16 &   0.05 &    390 &   0.30 &   0.09 &    559 &  41.00 &  18.09 &    267 &  27.93 &   9.98 \\ 
&     Texas &    732 &    730 &   0.76 &   0.25 &    548 &   0.72 &   0.21 &    551 &  48.44 &  11.32 &    438 &  27.05 &   5.36 \\ 
& Wisconsin &  1,004 &    993 &   1.69 &   0.44 &    727 &   2.05 &   0.52 &    621 &  77.87 &  17.24 &    471 &  52.91 &   9.97 \\
\midrule
$\mean$
&      Cora & 10,832 & 10,832 &   0.11 &   0.03 &  6,873 &   0.14 &   0.04 &  7,473 &  24.00 &  12.05 &  4,235 &  16.89 &   7.12 \\ 
&  CiteSeer & 13,248 & 13,234 &   0.09 &   0.04 & 10,215 &   0.11 &   0.04 & 11,708 &  13.90 &   7.62 &  8,887 &  11.97 &   5.87 \\ 
&   Cornell &    732 &    732 &   0.01 &   0.01 &    440 &   0.02 &   0.01 &    579 &  13.96 &   8.16 &    311 &  12.01 &   6.00 \\ 
&     Texas &    732 &    731 &   0.44 &   0.09 &    412 &   0.77 &   0.16 &    624 &  14.67 &   7.04 &    343 &  11.29 &   4.51 \\ 
& Wisconsin &  1,004 &  1,000 &   0.38 &   0.14 &    633 &   0.31 &   0.17 &    741 &  20.96 &   9.86 &    439 &  13.86 &   5.71 \\
\bottomrule
\end{tblr}
}
\end{table}

\begin{table}[t]
\centering
\caption{Detailed comparison results of $\toolname$ and $\gnnev$
for weak and general robustness on the Cora, CiteSeer, Cornell, Texas, and Wisconsin datasets with $\summ$ aggregation for global and local budget $10$.}
\label{tab:node_weak}
\resizebox{0.9\textwidth}{!}{
\begin{tblr}{
        rows    = {abovesep=0.2pt, belowsep=0.2pt},
        row{1}  = {font=\large, abovesep=3pt, belowsep=0pt},
        colspec = {l|l|rrrr|rrrr},
        cell{1}{3} = {c=4}{c},
        cell{1}{7} = {c=4}{c},
        cell{2}{3} = {c=2}{c},
        cell{2}{5} = {c=2}{c},
        cell{2}{7} = {c=2}{c},
        cell{2}{9} = {c=2}{c}
    }
\toprule
& & $\toolname$ & & & & $\gnnev$ \\
& & Robust instances & & Non-robust instances & & Robust instances & & Non-robust instances \\
\cmidrule[lr]{3-4} \cmidrule[lr]{5-6} \cmidrule[lr]{7-8} \cmidrule[lr]{9-10}
& &
$\quad\quad t_{a}(s)$ & $\quad\quad t_{g}(s)$ &
$\quad\quad t_{a}(s)$ & $\quad\quad t_{g}(s)$ &
$\quad\quad t_{a}(s)$ & $\quad\quad t_{g}(s)$ &
$\quad\quad t_{a}(s)$ & $\quad\quad t_{g}(s)$ \\
\midrule
$\summ$
&      Cora &   0.38 &   0.19 &   1.58 &   0.29 &   6.31 &   2.89 &  22.02 &  15.02 \\
(Weak)
&  CiteSeer &   0.19 &   0.07 &   0.02 &   0.01 &   7.63 &   4.19 &  22.84 &  17.64 \\
&   Cornell &   0.01 &   0.01 &   0.01 &   0.01 &  13.34 &   5.42 &  12.73 &   9.13 \\
&     Texas &   0.02 &   0.02 &   1.61 &   0.92 &   0.93 &   0.67 &  13.28 &  12.08 \\
& Wisconsin &   0.08 &   0.06 &   0.02 &   0.02 &   6.08 &   3.55 &  30.86 &  20.24 \\
\midrule
$\summ$
&      Cora &   1.02 &   0.34 &   0.06 &   0.03 &  10.44 &   3.49 &  25.01 &  15.95 \\
&  CiteSeer &   0.38 &   0.10 &   0.01 &   0.01 &  12.90 &   5.35 &  26.55 &  18.45 \\
&   Cornell &   0.01 &   0.01 &   0.01 &   0.01 &   8.62 &   4.06 &  11.17 &   9.04 \\
&     Texas &   0.02 &   0.02 &   0.01 &   0.01 &   1.09 &   0.95 &  19.19 &  13.58 \\
& Wisconsin &   0.22 &   0.14 &   0.01 &   0.01 &   8.80 &   4.10 &  30.47 &  19.09 \\
\bottomrule
\end{tblr}
}
\end{table}

\begin{figure}[t]
\caption{The number of instances solved by each tool plotted against runtime under different aggregations and budgets.
The solid line denotes $\toolname$, the dashed line denotes $\gnnev$, and the dotted line denotes $\scip$.
The blue, orange, green, and red lines correspond to budgets of $1$, $2$, $5$, and $10$, respectively.
Note that the $x$ axis is in logarithmic scale.}
  \centering
  \resizebox{\textwidth}{!}{
  \begin{tabular}{c c c c c c c}
    & & Cora & CiteSeer & Cornell & Texas & Wisconsin \\
      \makecell{\rotatebox{90}{$\summ$}} & \makecell{\rotatebox{90}{(weak)}}
    & \adjustbox{valign=c}{\includegraphics[width=0.25\linewidth]{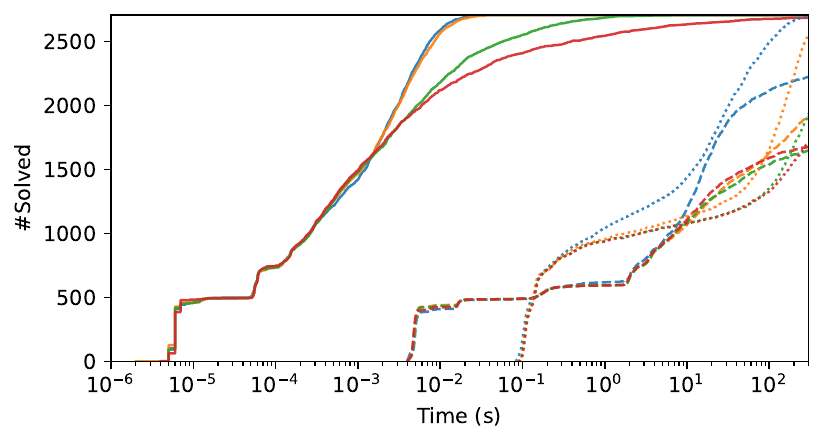}} 
    & \adjustbox{valign=c}{\includegraphics[width=0.25\linewidth]{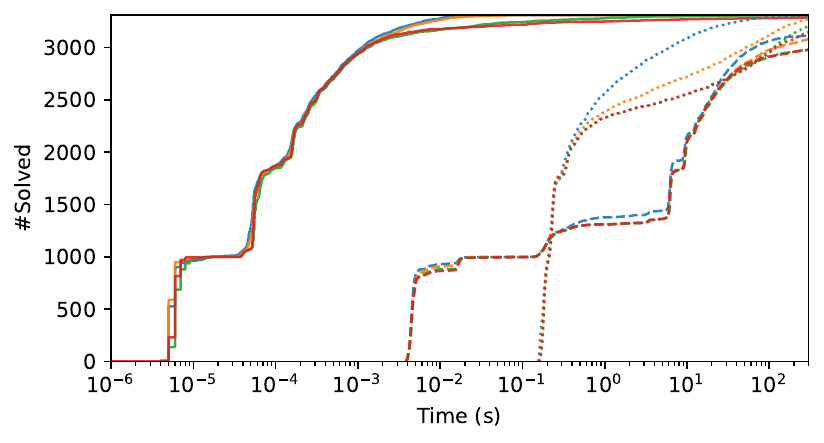}} 
    & \adjustbox{valign=c}{\includegraphics[width=0.25\linewidth]{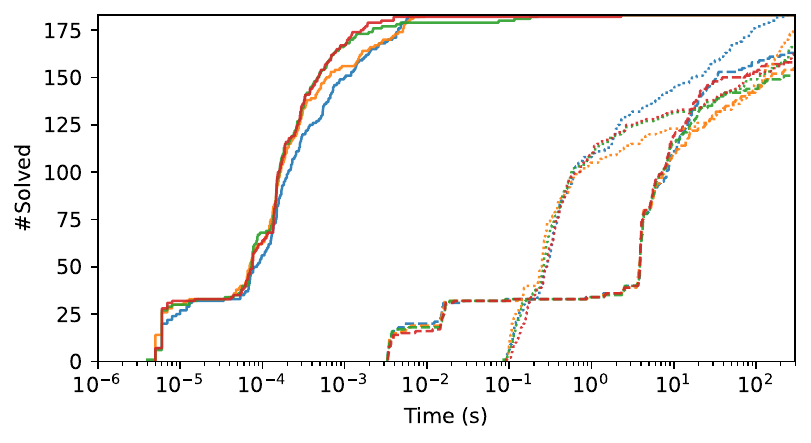}}
    & \adjustbox{valign=c}{\includegraphics[width=0.25\linewidth]{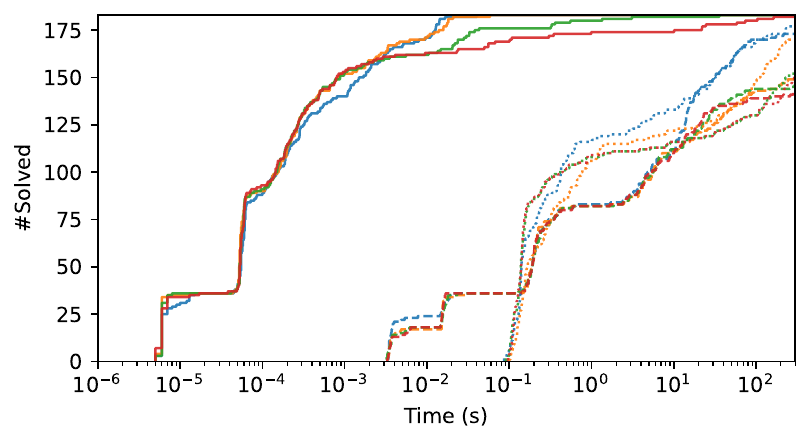}}
    & \adjustbox{valign=c}{\includegraphics[width=0.25\linewidth]{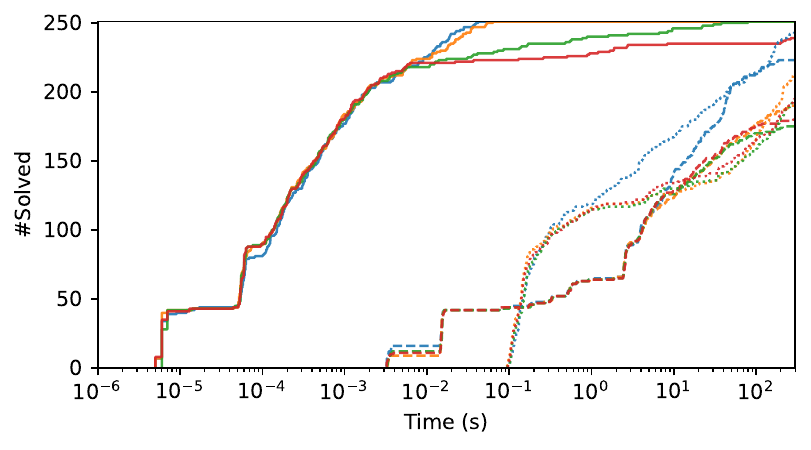}} \\
      \makecell{\rotatebox{90}{$\summ$}} &
    & \adjustbox{valign=c}{\includegraphics[width=0.25\linewidth]{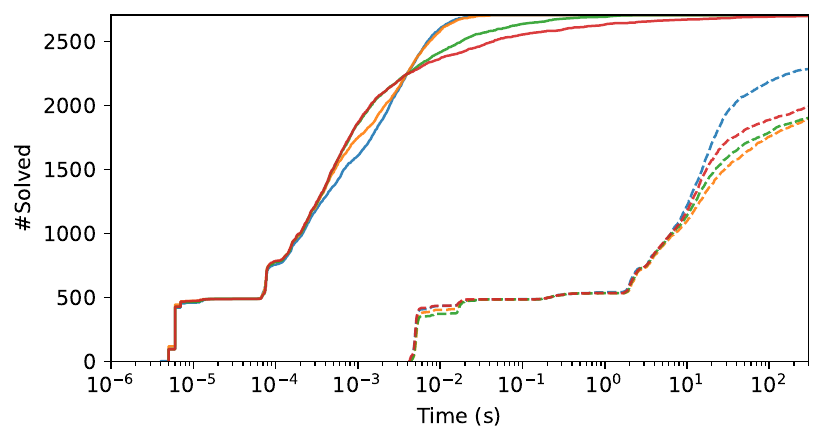}} 
    & \adjustbox{valign=c}{\includegraphics[width=0.25\linewidth]{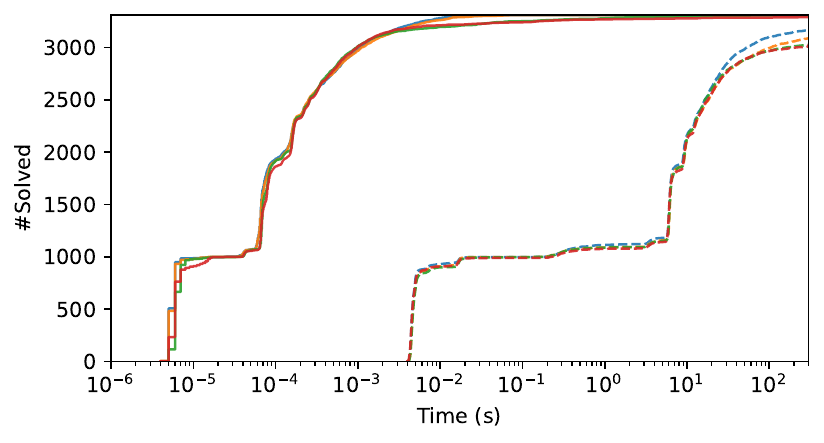}} 
    & \adjustbox{valign=c}{\includegraphics[width=0.25\linewidth]{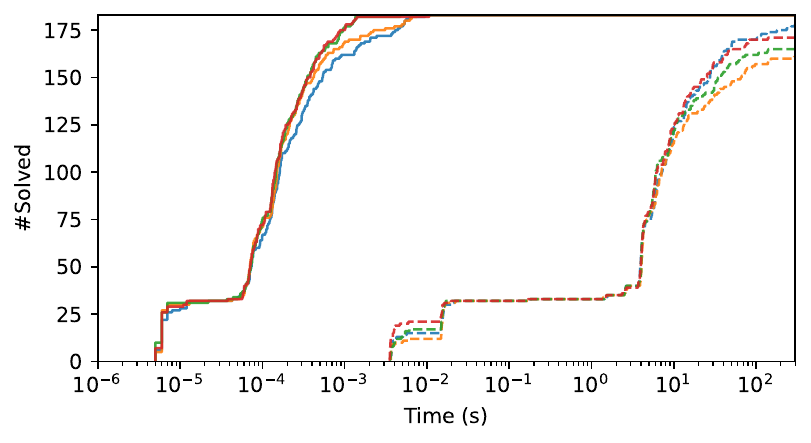}}
    & \adjustbox{valign=c}{\includegraphics[width=0.25\linewidth]{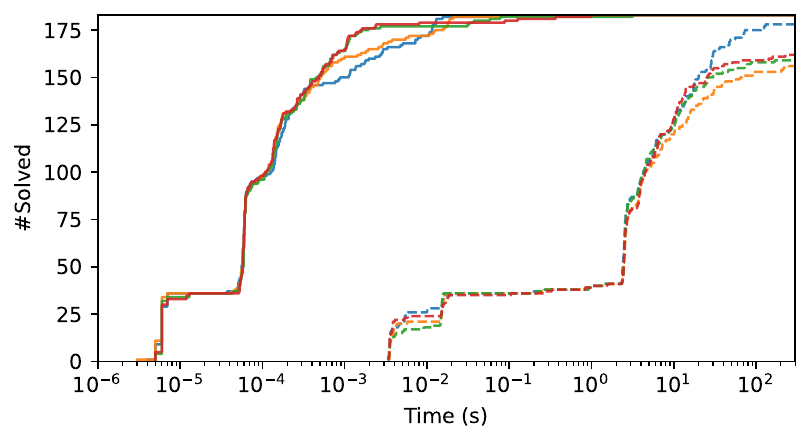}}
    & \adjustbox{valign=c}{\includegraphics[width=0.25\linewidth]{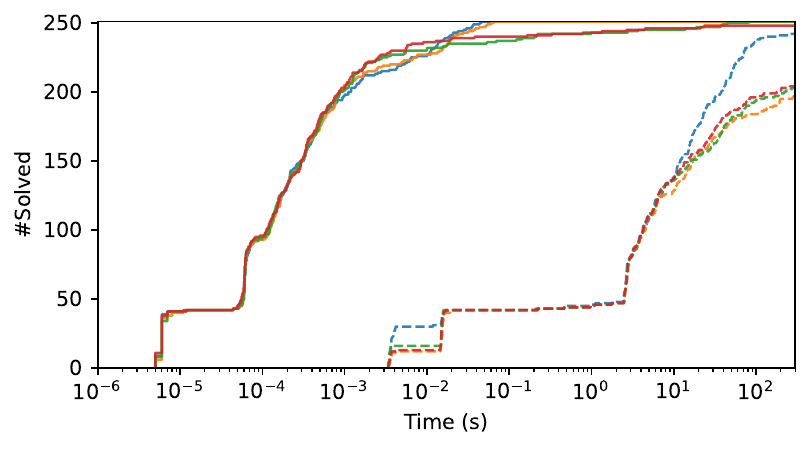}} \\
      \makecell{\rotatebox{90}{$\maxx$}} &
    & \adjustbox{valign=c}{\includegraphics[width=0.25\linewidth]{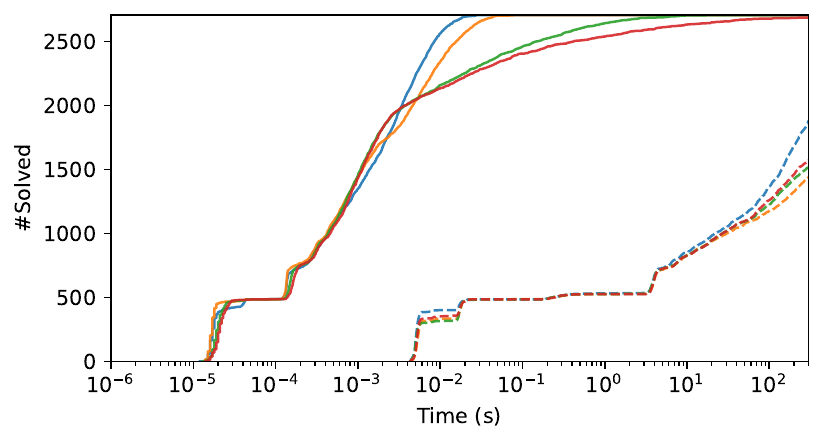}} 
    & \adjustbox{valign=c}{\includegraphics[width=0.25\linewidth]{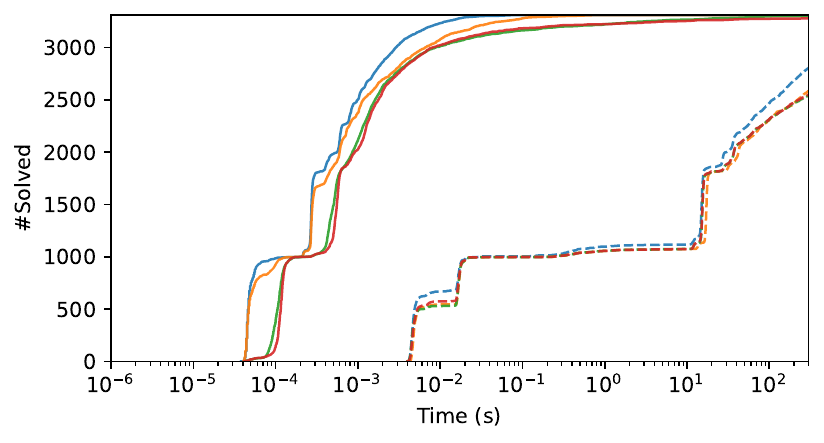}} 
    & \adjustbox{valign=c}{\includegraphics[width=0.25\linewidth]{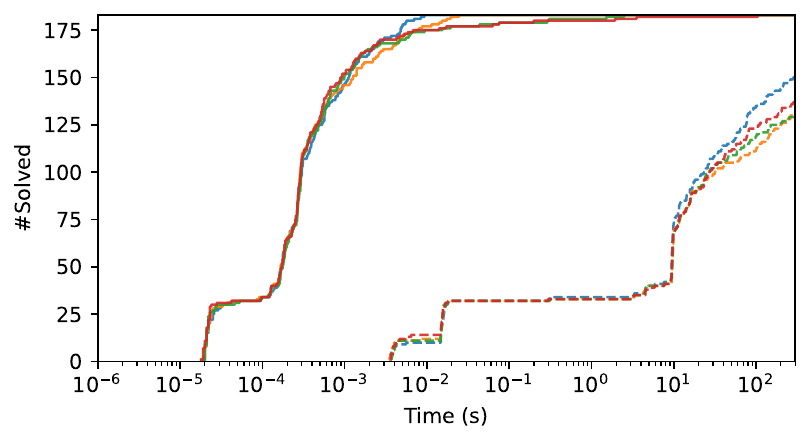}}
    & \adjustbox{valign=c}{\includegraphics[width=0.25\linewidth]{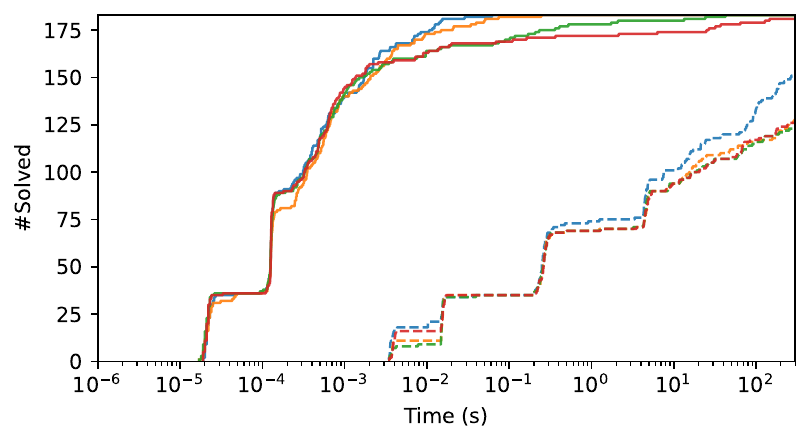}}
    & \adjustbox{valign=c}{\includegraphics[width=0.25\linewidth]{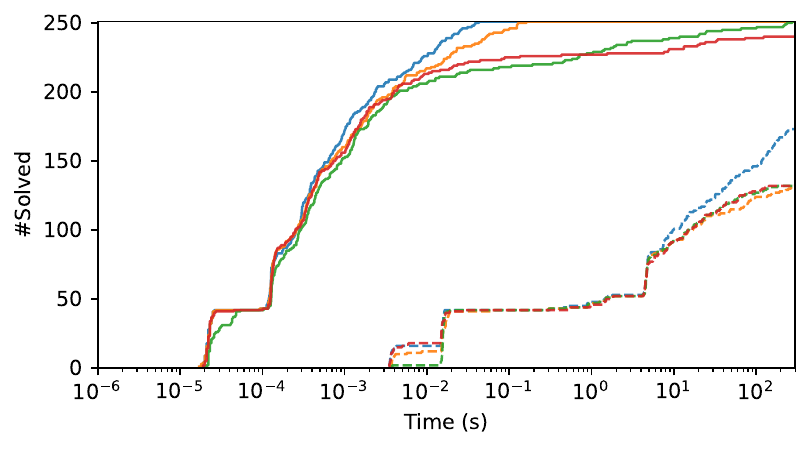}} \\
      \makecell{\rotatebox{90}{$\mean$}} &
    & \adjustbox{valign=c}{\includegraphics[width=0.25\linewidth]{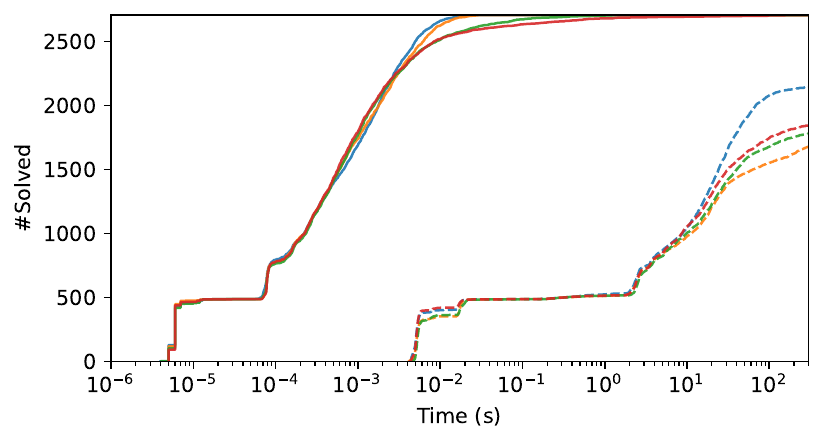}} 
    & \adjustbox{valign=c}{\includegraphics[width=0.25\linewidth]{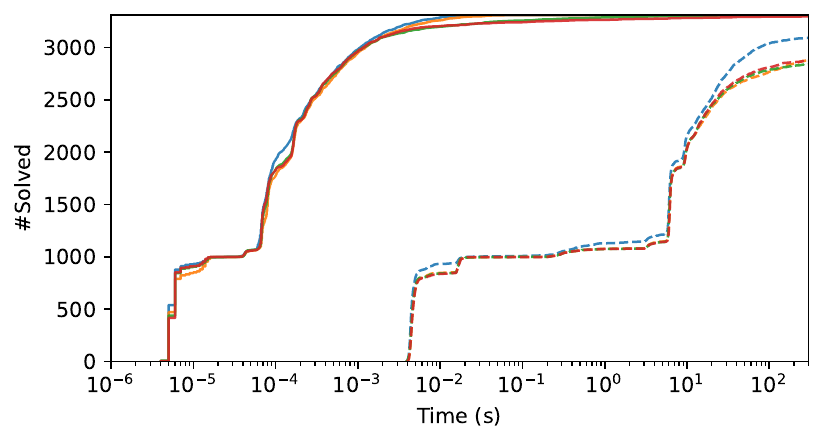}} 
    & \adjustbox{valign=c}{\includegraphics[width=0.25\linewidth]{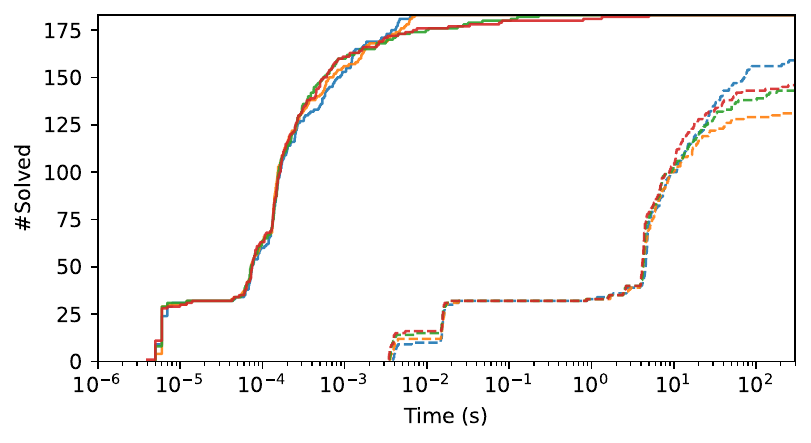}}
    & \adjustbox{valign=c}{\includegraphics[width=0.25\linewidth]{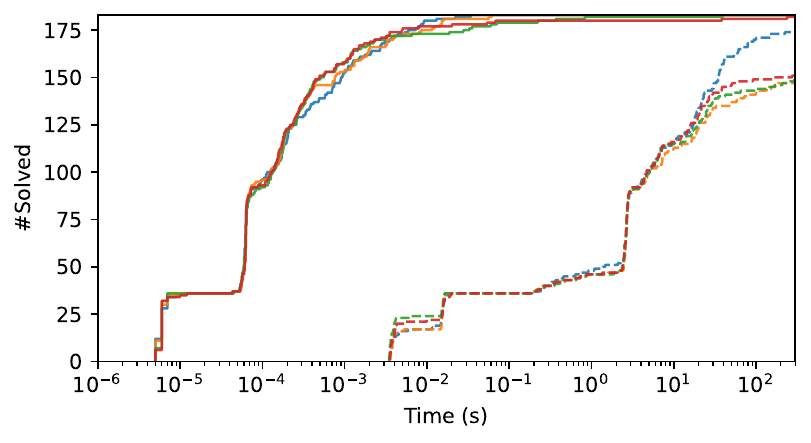}}
    & \adjustbox{valign=c}{\includegraphics[width=0.25\linewidth]{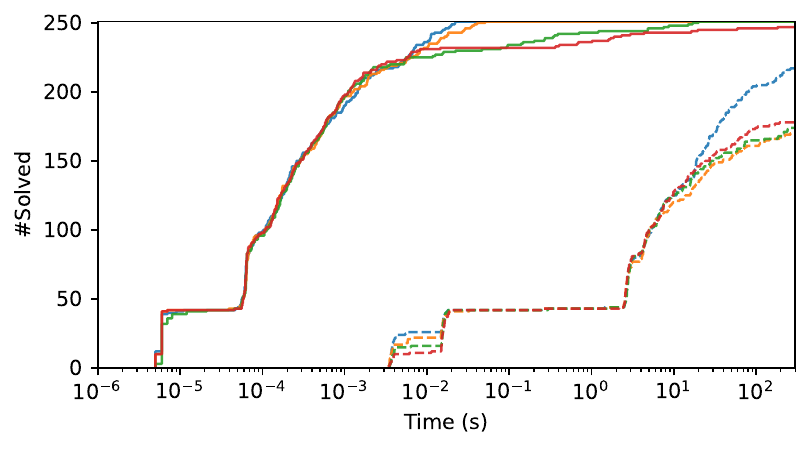}} \\
  \end{tabular}
  }\label{fig:node}
\end{figure}

\subsubsection{Evaluation of Optimization Strategies}
Our second experiment evaluates the optimization strategies described in Section~\ref{sec:opt}.
Recall that the runtime of Algorithm~\ref{algo:dfs} depends on
both the number of recursive calls and the runtime of each call.
We introduce the notion of the \emph{exploration ratio}
to quantify how these optimization strategies affect the number of recursive calls.
Let $F$ denote the set of fragile edges.
Since the number of recursive calls grows exponentially with $\abs{F}$,
we define the exploration ratio $\alpha$ to be such that $2^{\alpha \cdot (\abs{F}+1)}$ is the number of recursive calls.
That is, the exploration ratio is defined
as the logarithm of the number of recursive calls divided by $(\abs{F} + 1)$.
In the worst case, Algorithm~\ref{algo:dfs} explores all possible $2^{\abs{F}}$ graphs,
resulting in $2^{(\abs{F}+1)} - 1$ recursive calls,
which implies an exploration ratio close to $1$.

We conducted experiments on several variants of $\toolname$ that
exclude the optimizations mentioned in Section~\ref{sec:opt}.
The experiments were performed on the Cora, CiteSeer, Cornell, Texas, and Wisconsin datasets
using 4-layer GNNs with different aggregation functions, with both global and local budgets set to $10$.
The results are summarized in Table~\ref{tab:variants},
where N/A indicates that the strategy is not applicable in this case.
For a fair comparison, we only report instances that solved by all variants.
Therefore the average runtime for $\toolname$ is different from previous experiments.
Figure~\ref{fig:prof} provides a different view,
showing how many recursive calls $\toolname$ and $\toolname$ without optimization strategies need
to solve the instances.

\paragraph{Takeaways} The results show that the overall improvement exceeds an order of magnitude.
For incremental computation, the exploration ratio remains the same while the average runtime per call decreases,
which meets our expectation, since incremental computation only accelerates the oracle’s computation without altering its outcome,
thereby producing the same computational path for $\toolname$.
For operator reordering, the average runtime per call decrease as expected.
The average exploration ratio also decreases in this case since we incorporating operator reordering and incremental computation.
For bound tightening with budgets, the overall runtime remains roughly the same or slightly increases;
this may be because the benefit from shrinking is limited,
as indicated by the exploration ratio remaining nearly unchanged.
Finally, for heuristic edge picking, the exploration ratio decreases as expected,
but the average runtime per call increases because $\toolname$ must maintain and compute information about the $k$-hop neighbors in the current graph,
which introduces additional computational overhead.

For the non-robust instances,
most points lie near the bottom of the figure in Fig~\ref{fig:prof},
regardless of the optimizations applied to $\toolname$.
This is because most non-robust instances can be quickly identified by our lightweight non-robust tester.
For the robust instances, they form a clear line in the diagram,
which indicates that the exploration ratio is quite stable across different quantities of fragile edges in the same instance.
Since the $y$ axis is plotted on a logarithmic scale,
the slope of the line corresponds to the exploration ratio~$\alpha$.
From Fig~\ref{fig:prof}, we can also observe that the slope for $\toolname$ is smaller than that for $\toolname$ without optimization strategies.

\begin{table}[t]
\centering
\caption{The detailed comparison results of variants of $\toolname$ on the Cora, CiteSeer, Cornell, Texas, and Wisconsin datasets using 4-layer GNNs of varying aggregation function,
with both global and local budgets set to $10$.
$t_a$ denotes the average overall runtime, $t_c$ denotes the average runtime per call, and ER denoted the average exploration ratio.
N/A indicates that the strategy is not applicable in this case.}
\label{tab:variants}
\resizebox{\textwidth}{!}{
\begin{tblr}{
        rows    = {abovesep=0.2pt, belowsep=0.2pt},
        row{1}  = {font=\large, abovesep=3pt, belowsep=0pt},
        colspec = {l|l|rrr|rrr|rrr|rrr|rrr|rrr},
        cell{1}{3}  = {c=3}{c},
        cell{1}{6}  = {c=3}{c},
        cell{1}{9}  = {c=3}{c},
        cell{1}{12} = {c=3}{c},
        cell{1}{15} = {c=3}{c},
        cell{1}{18} = {c=3}{c},
        cell{2}{3}  = {c=3}{c},
        cell{2}{6}  = {c=3}{c},
        cell{2}{9}  = {c=3}{c},
        cell{2}{12} = {c=3}{c},
        cell{2}{15} = {c=3}{c},
        cell{2}{18} = {c=3}{c}
    }
\toprule
& & $\toolname$ & & & $\toolname$    & & & $\toolname$          & & & $\toolname$          & & & $\toolname$           & & & $\toolname$ \\
& &             & & & w/o inc. comp. & & & w/o operator reorder & & & w/o bound tightening & & & w/o heuristic picking & & & w/o all optimizations \\
\cmidrule[lr]{3-5} \cmidrule[lr]{6-8} \cmidrule[lr]{9-11} \cmidrule[lr]{12-14} \cmidrule[lr]{15-17} \cmidrule[lr]{18-20}
&
& $\quad t_a$(s) & $\quad t_c$(ms) & ER
& $\quad t_a$(s) & $\quad t_c$(ms) & ER
& $\quad t_a$(s) & $\quad t_c$(ms) & ER
& $\quad t_a$(s) & $\quad t_c$(ms) & ER
& $\quad t_a$(s) & $\quad t_c$(ms) & ER
& $\quad t_a$(s) & $\quad t_c$(ms) & ER \\
\midrule
$\summ$
&      Cora &   0.002 &   0.005 & 0.29 &   0.007 &   0.018 & 0.29 &   0.021 &   0.031 & 0.30 &   0.002 &   0.005 & 0.29 &   0.048 &   0.002 & 0.34 &   5.143 &   0.209 & 0.34 \\
&  CiteSeer &   0.001 &   0.005 & 0.42 &   0.002 &   0.013 & 0.42 &   0.037 &   0.159 & 0.43 &   0.001 &   0.005 & 0.42 &   0.002 &   0.003 & 0.44 &   1.111 &   1.052 & 0.44 \\
&   Cornell &   0.001 &   0.022 & 0.32 &   0.001 &   0.012 & 0.32 &   0.001 &   0.042 & 0.32 &   0.001 &   0.021 & 0.32 &   0.001 &   0.010 & 0.32 &   0.006 &   0.155 & 0.32 \\
&     Texas &   0.001 &   0.004 & 0.30 &   0.010 &   0.025 & 0.30 &   0.017 &   0.026 & 0.30 &   0.002 &   0.003 & 0.30 &   0.010 &   0.002 & 0.31 &   1.852 &   0.440 & 0.31 \\
& Wisconsin &   0.001 &   0.008 & 0.39 &   0.001 &   0.019 & 0.39 &   0.003 &   0.029 & 0.39 &   0.001 &   0.008 & 0.39 &   0.003 &   0.003 & 0.41 &   0.287 &   0.192 & 0.41 \\
\midrule
$\maxx$
&      Cora &   0.022 &   0.033 & 0.31 &   0.166 &   0.251 & 0.31 &     N/A &     N/A &  N/A &   0.022 &   0.033 & 0.31 &   0.622 &   0.022 & 0.36 &   6.771 &   0.239 & 0.36 \\
&  CiteSeer &   0.033 &   0.156 & 0.43 &   0.255 &   1.210 & 0.43 &     N/A &     N/A &  N/A &   0.033 &   0.157 & 0.43 &   0.163 &   0.119 & 0.45 &   1.574 &   1.147 & 0.45 \\
&   Cornell &   0.003 &   0.030 & 0.38 &   0.014 &   0.166 & 0.38 &     N/A &     N/A &  N/A &   0.003 &   0.030 & 0.38 &   0.006 &   0.013 & 0.39 &   0.066 &   0.143 & 0.39 \\
&     Texas &   0.040 &   0.035 & 0.49 &   0.351 &   0.306 & 0.49 &     N/A &     N/A &  N/A &   0.041 &   0.033 & 0.49 &   0.139 &   0.033 & 0.50 &   1.282 &   0.285 & 0.50 \\
& Wisconsin &   0.005 &   0.033 & 0.45 &   0.039 &   0.287 & 0.45 &     N/A &     N/A &  N/A &   0.005 &   0.033 & 0.45 &   0.128 &   0.012 & 0.48 &   2.645 &   0.254 & 0.48 \\
\midrule
$\mean$
&      Cora &   0.001 &   0.015 & 0.27 &   0.002 &   0.029 & 0.27 &   0.011 &   0.056 & 0.28 &   0.001 &   0.015 & 0.27 &   0.042 &   0.004 & 0.33 &   6.183 &   0.322 & 0.34 \\
&  CiteSeer &   0.001 &   0.008 & 0.42 &   0.002 &   0.021 & 0.42 &   0.044 &   0.241 & 0.43 &   0.001 &   0.008 & 0.42 &   0.003 &   0.004 & 0.44 &   1.720 &   1.330 & 0.45 \\
&   Cornell &   0.001 &   0.006 & 0.39 &   0.003 &   0.018 & 0.39 &   0.007 &   0.038 & 0.40 &   0.001 &   0.005 & 0.39 &   0.003 &   0.003 & 0.41 &   0.221 &   0.199 & 0.41 \\
&     Texas &   0.001 &   0.006 & 0.37 &   0.003 &   0.019 & 0.37 &   0.012 &   0.043 & 0.37 &   0.001 &   0.006 & 0.37 &   0.007 &   0.005 & 0.38 &   0.616 &   0.225 & 0.38 \\
& Wisconsin &   0.001 &   0.016 & 0.38 &   0.001 &   0.018 & 0.38 &   0.002 &   0.058 & 0.39 &   0.001 &   0.017 & 0.38 &   0.002 &   0.004 & 0.40 &   0.204 &   0.332 & 0.41 \\
\bottomrule
\end{tblr}
}
\end{table}

\begin{figure}[t]
\caption{The number of recursive calls plotted against the number of edges in the set of fragile edges.
The blue dots represent the robust instances solved by $\toolname$,
the orange dots represent the non-robust instances solved by $\toolname$,
the green dots represent the robust instances solved by $\toolname$ without any optimization strategies,
and the red dots represent the non-robust instances solved by $\toolname$ without any optimization strategies.
Note that the $y$ axis is in logarithmic scale.}
  \centering
  \resizebox{\textwidth}{!}{
  \begin{tabular}{c c c c c c}
    & Cora & CiteSeer & Cornell & Texas & Wisconsin \\
      \makecell{\rotatebox{90}{$\summ$}}
    & \adjustbox{valign=c}{\includegraphics[width=0.25\linewidth]{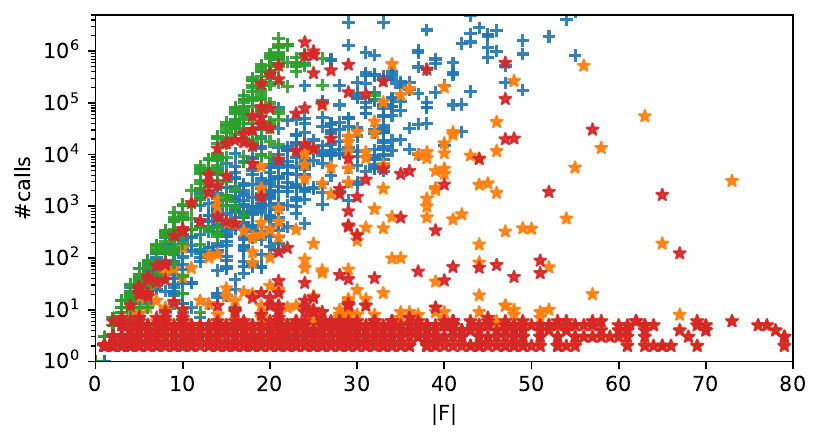}} 
    & \adjustbox{valign=c}{\includegraphics[width=0.25\linewidth]{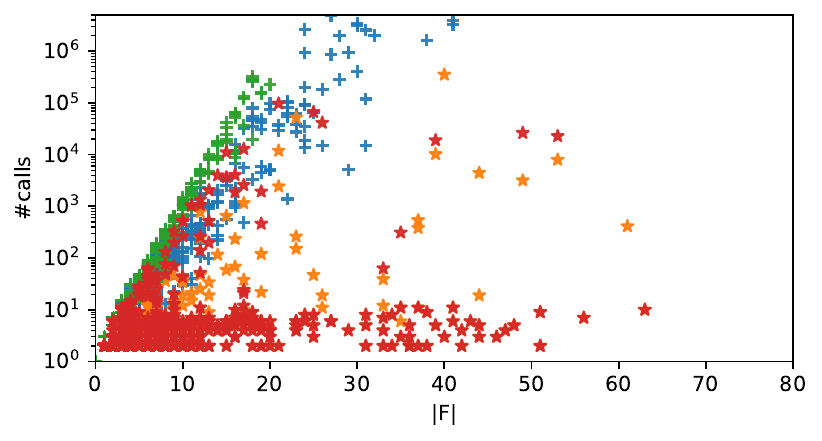}}
    & \adjustbox{valign=c}{\includegraphics[width=0.25\linewidth]{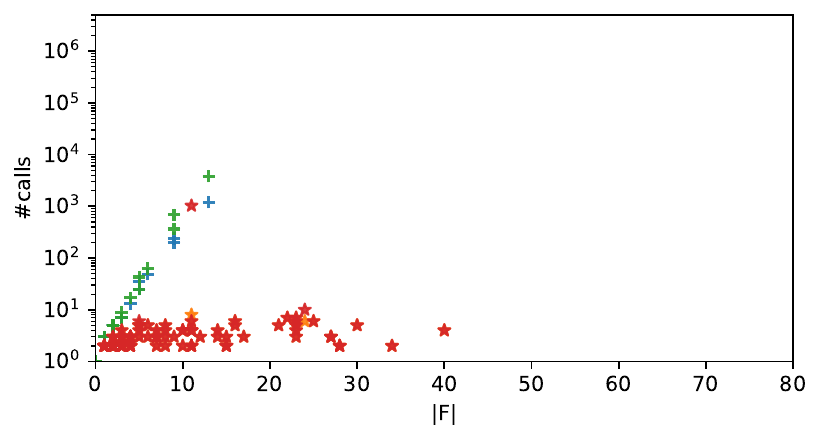}} 
    & \adjustbox{valign=c}{\includegraphics[width=0.25\linewidth]{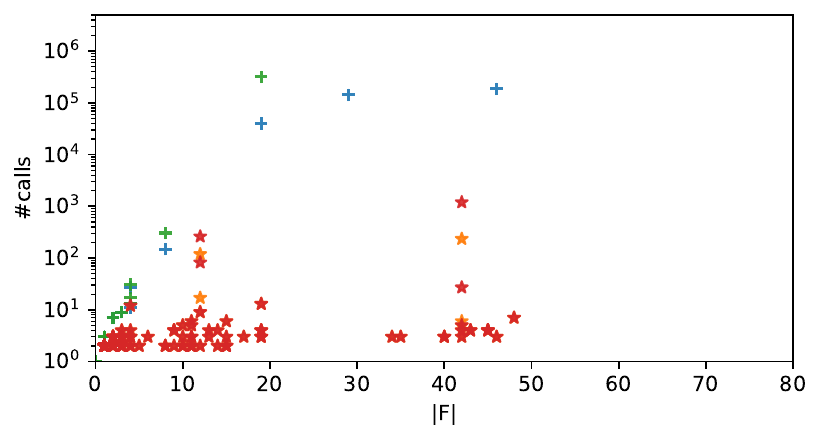}} 
    & \adjustbox{valign=c}{\includegraphics[width=0.25\linewidth]{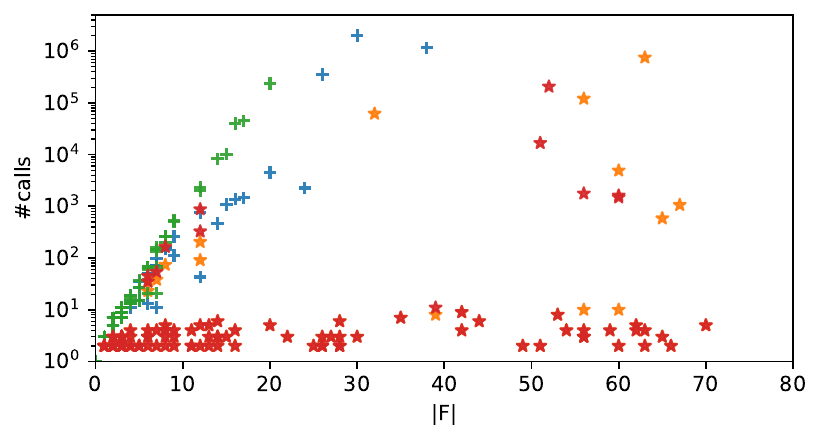}} \\
      \makecell{\rotatebox{90}{$\maxx$}}
    & \adjustbox{valign=c}{\includegraphics[width=0.25\linewidth]{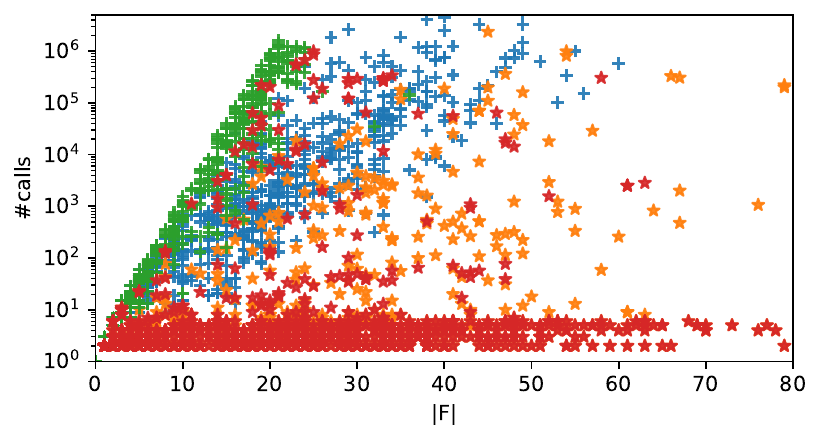}} 
    & \adjustbox{valign=c}{\includegraphics[width=0.25\linewidth]{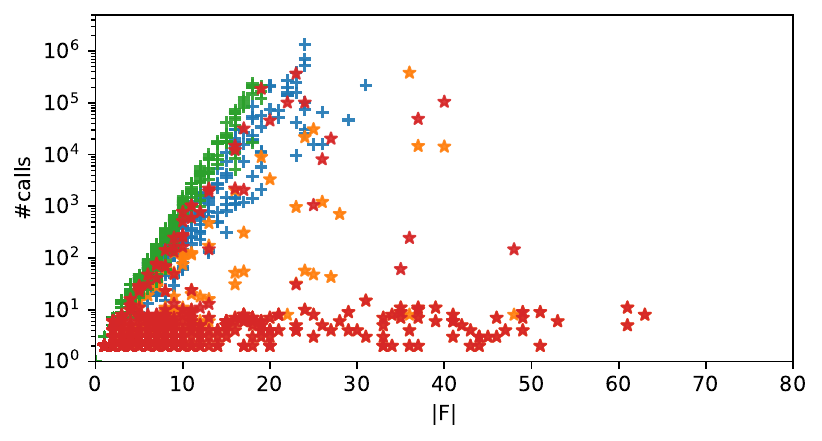}} 
    & \adjustbox{valign=c}{\includegraphics[width=0.25\linewidth]{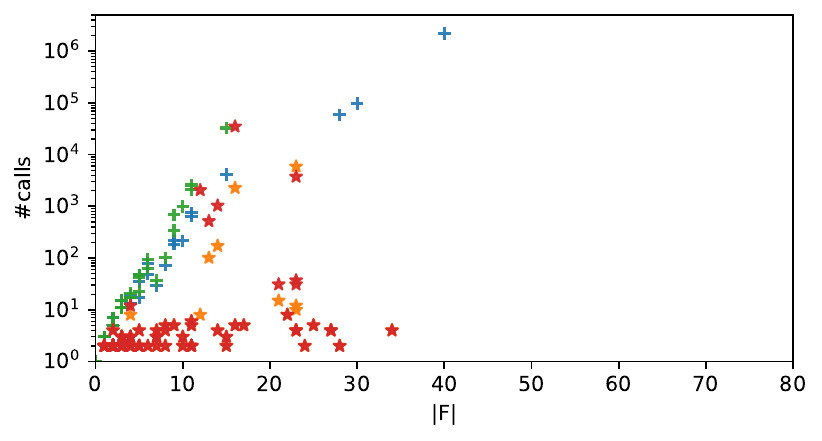}} 
    & \adjustbox{valign=c}{\includegraphics[width=0.25\linewidth]{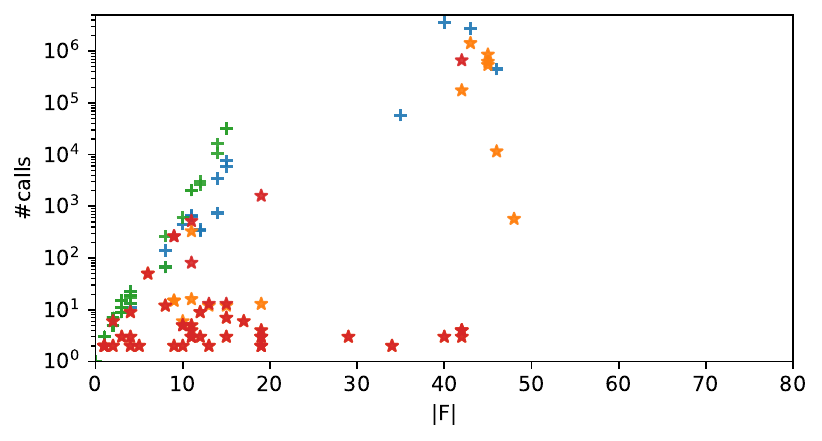}} 
    & \adjustbox{valign=c}{\includegraphics[width=0.25\linewidth]{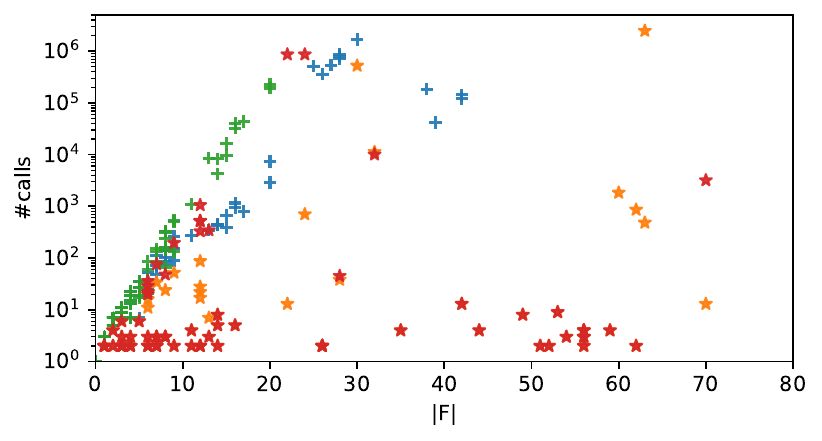}} \\
      \makecell{\rotatebox{90}{$\mean$}}
    & \adjustbox{valign=c}{\includegraphics[width=0.25\linewidth]{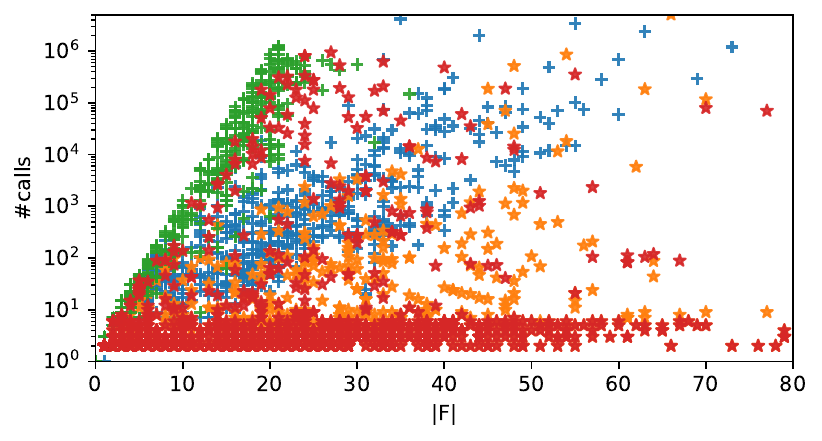}} 
    & \adjustbox{valign=c}{\includegraphics[width=0.25\linewidth]{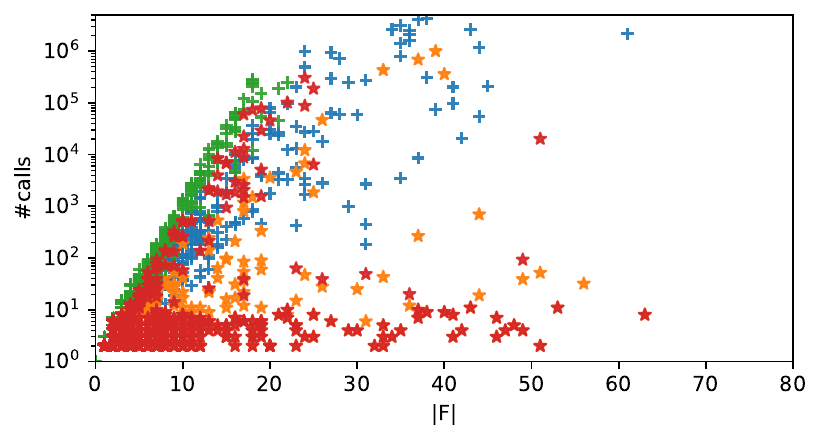}} 
    & \adjustbox{valign=c}{\includegraphics[width=0.25\linewidth]{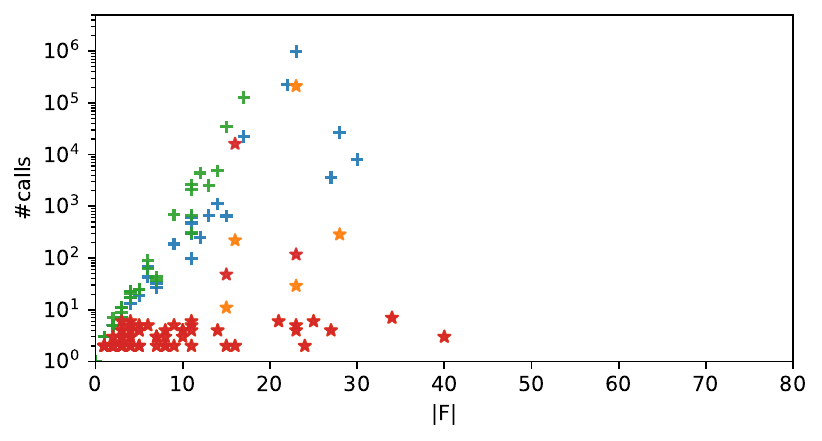}} 
    & \adjustbox{valign=c}{\includegraphics[width=0.25\linewidth]{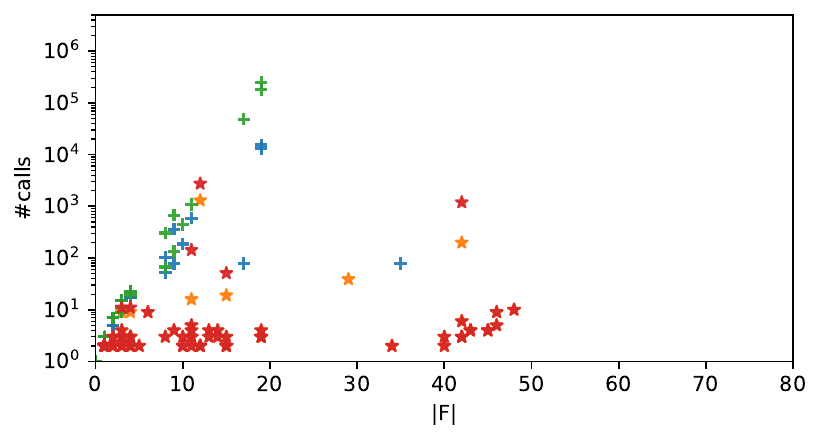}} 
    & \adjustbox{valign=c}{\includegraphics[width=0.25\linewidth]{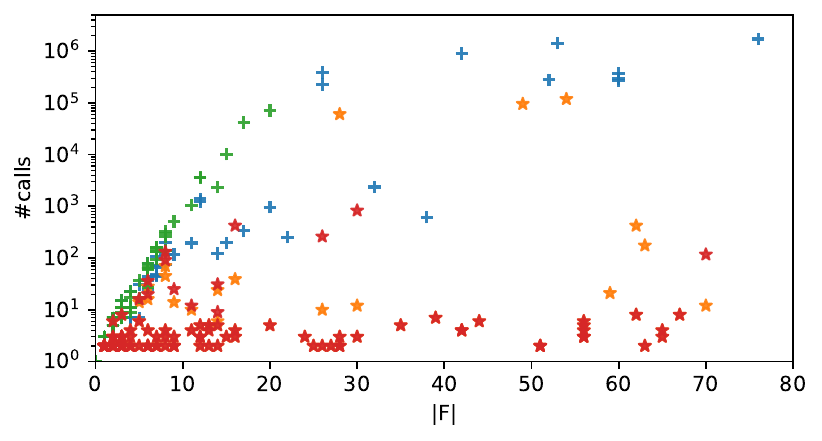}} \\
  \end{tabular}
  }
  \label{fig:prof}
\end{figure}

\subsubsection{Results on Robustness Radius}
The third experiment is about the robustness radius, which $\toolname$ can compute using a variation of the top-level algorithm
We conduct experiments on the Cora, CiteSeer, Cornell, Texas, and Wisconsin datasets with 4-layer GNNs with varying aggregation functions with timeout 600s.
We summarize the results in Table~\ref{tab:radius}.

Note that the true radius in each of these benchmarks is not large: at most~$11$.
Despite this, to our knowledge, \emph{no prior tool can compute the radius exactly}.

\begin{table}[t]
\centering
\caption{Robustness radius of the Cora, CiteSeer, Cornell, Texas, and Wisconsin datasets using 4-layer GNNs of varying aggregation function.
TO denotes time out, $\#$ is the number of instances, and $t_a$ is the average overall runtime.
Note that -- indicates that there is no solved instance for this radius.}
\label{tab:radius}
\resizebox{\textwidth}{!}{
\begin{tblr}{
        rows    = {abovesep=0.2pt, belowsep=0.2pt},
        row{1}  = {font=\large, abovesep=3pt, belowsep=0pt},
        colspec = {l|l|r|rr|rr|rr|rr|rr|rr|rr|rr|rr|rr|rr|rr|rr},
        cell{1}{4}  = {c=2}{c},
        cell{1}{6}  = {c=2}{c},
        cell{1}{8}  = {c=2}{c},
        cell{1}{10} = {c=2}{c},
        cell{1}{12} = {c=2}{c},
        cell{1}{14} = {c=2}{c},
        cell{1}{16} = {c=2}{c},
        cell{1}{18} = {c=2}{c},
        cell{1}{20} = {c=2}{c},
        cell{1}{22} = {c=2}{c},
        cell{1}{24} = {c=2}{c},
        cell{1}{26} = {c=2}{c},
        cell{1}{28} = {c=2}{c}
    }
\toprule
& & TO &
Robust   & &
$r = 0$  & &
$r = 1$  & &
$r = 2$  & &
$r = 3$  & &
$r = 4$  & &
$r = 5$  & &
$r = 6$  & &
$r = 7$  & &
$r = 8$  & &
$r = 9$  & &
$r = 10$ & &
$r = 11$  \\
\cmidrule[lr]{3-3}
\cmidrule[lr]{4-5}
\cmidrule[lr]{6-7}
\cmidrule[lr]{8-9}
\cmidrule[lr]{10-11}
\cmidrule[lr]{12-13}
\cmidrule[lr]{14-15}
\cmidrule[lr]{16-17}
\cmidrule[lr]{18-19}
\cmidrule[lr]{20-21}
\cmidrule[lr]{22-23}
\cmidrule[lr]{24-25}
\cmidrule[lr]{26-27}
\cmidrule[lr]{28-29}
& & \# &
\# & $t_{a}(s)$ &
\# & $t_{a}(s)$ &
\# & $t_{a}(s)$ &
\# & $t_{a}(s)$ &
\# & $t_{a}(s)$ &
\# & $t_{a}(s)$ &
\# & $t_{a}(s)$ &
\# & $t_{a}(s)$ &
\# & $t_{a}(s)$ &
\# & $t_{a}(s)$ &
\# & $t_{a}(s)$ &
\# & $t_{a}(s)$ &
\# & $t_{a}(s)$ \\
\midrule
$\summ$
&      Cora &  17 & 1,447 &   1.67 & 580 &   0.01 & 349 &   0.02 & 172 &   0.47 &  94 &   0.47 &  39 &   0.14 &   7 &   0.25 &   2 &   0.79 &   1 &   0.97 &  -- &     -- &  -- &     -- &  -- &     -- &  -- &     -- \\
&  CiteSeer &  26 & 2,388 &   0.38 & 521 &   0.01 & 206 &   0.01 &  87 &   0.01 &  40 &   0.01 &  12 &   0.10 &   9 &   0.21 &  11 &  14.77 &   5 &  54.93 &   2 &   3.90 &   2 & 265.39 &   3 & 150.66 &  -- &     -- \\
&   Cornell &   0 &    69 &   0.01 &  73 &   0.01 &  31 &   0.01 &   7 &   0.01 &   2 &   0.01 &   1 &   0.04 &  -- &     -- &  -- &     -- &  -- &     -- &  -- &     -- &  -- &     -- &  -- &     -- &  -- &     -- \\
&     Texas &   0 &    61 &   0.27 &  77 &   0.01 &  31 &   0.01 &   9 &   0.01 &   3 &   4.48 &  -- &     -- &   1 &   3.42 &   1 &   0.06 &  -- &     -- &  -- &     -- &  -- &     -- &  -- &     -- &  -- &     -- \\
& Wisconsin &   7 &   141 &   0.85 &  62 &   0.01 &  23 &   0.01 &  10 &   0.02 &   1 &   0.09 &   4 &   2.19 &   2 &  59.99 &  -- &     -- &   1 &  18.83 &  -- &     -- &  -- &     -- &  -- &     -- &  -- &     -- \\
\midrule
$\maxx$
&      Cora &  18 & 1,520 &   3.39 & 529 &   0.10 & 330 &   1.03 & 178 &   4.01 &  82 &   0.24 &  37 &   0.81 &   7 &   2.86 &   2 &   1.78 &   4 &  23.97 &   1 & 596.55 &  -- &     -- &  -- &     -- &  -- &     -- \\
&  CiteSeer &  38 & 2,438 &   1.01 & 537 &   0.01 & 159 &   0.08 &  85 &   0.12 &  27 &   0.86 &  14 &   4.45 &  10 & 105.20 &   2 &  10.67 &   2 & 132.81 &  -- &     -- &  -- &     -- &  -- &     -- &  -- &     -- \\
&   Cornell &   0 &    89 &   2.41 &  71 &   0.01 &  13 &   0.01 &   5 &   0.02 &   3 &   0.04 &   1 &   0.02 &  -- &     -- &  -- &     -- &  -- &     -- &  -- &     -- &   1 &  30.10 &  -- &     -- &  -- &     -- \\
&     Texas &   7 &   127 &   4.74 &  32 &   5.62 &  12 &  41.31 &   5 &   0.12 &  -- &     -- &  -- &     -- &  -- &     -- &  -- &     -- &  -- &     -- &  -- &     -- &  -- &     -- &  -- &     -- &  -- &     -- \\
& Wisconsin &  14 &   160 &   1.49 &  48 &   0.42 &  13 &   0.57 &   8 &   0.78 &   3 &   4.75 &   2 &  40.46 &   1 &   1.17 &   2 &  69.81 &  -- &     -- &  -- &     -- &  -- &     -- &  -- &     -- &  -- &     -- \\
\midrule
$\mean$
&      Cora &   2 & 1,422 &   1.08 & 535 &   0.01 & 356 &   0.01 & 212 &   1.48 &  97 &   0.15 &  47 &   0.07 &  23 &   0.41 &   7 &   2.53 &   6 &   1.77 &  -- &     -- &  -- &     -- &   1 & 166.18 &  -- &     -- \\
&  CiteSeer &  18 & 2,418 &   1.05 & 568 &   0.02 & 153 &   0.01 &  77 &   0.01 &  30 &   0.01 &  25 &   0.07 &   8 &   0.69 &   4 &   8.04 &   6 &  62.78 &   5 &  92.31 &  -- &     -- &  -- &     -- &  -- &     -- \\
&   Cornell &   0 &    97 &   0.08 &  53 &   0.01 &  16 &   0.01 &   7 &   0.01 &   5 &   0.01 &   3 &   0.68 &   1 &   0.11 &  -- &     -- &  -- &     -- &   1 &   0.08 &  -- &     -- &  -- &     -- &  -- &     -- \\
&     Texas &   2 &    87 &   6.42 &  56 &   0.01 &  23 &   0.01 &   8 &   0.01 &   1 &   0.01 &   2 &   0.84 &  -- &     -- &   2 &   0.04 &   1 &   2.55 &  -- &     -- &   1 &   0.13 &  -- &     -- &  -- &     -- \\
& Wisconsin &   7 &   133 &   1.14 &  65 &   0.01 &  19 &   0.03 &  14 &   0.08 &   3 &   0.05 &   4 &   0.86 &   4 &   6.54 &  -- &     -- &   1 &  19.35 &  -- &     -- &  -- &     -- &  -- &     -- &   1 & 111.57 \\
\bottomrule
\end{tblr}
}
\end{table}

\subsubsection{End to End Performance on Graph Classification}
Finally, we conduct experiments on the MUTAG and ENZYMES datasets,
which are undirected and involve graph-level classification instances.
In these cases, a final sum pooling operation is applied to obtain a single graph-level output.
Details of graph-level robustness and set up for undirected graphs can be found in Appendix~\ref{app:graphvariant} and Appendix~\ref{app:undirected}.
We consider both deletion and insertion perturbations, as well as a local budget constraint --- that is, a local budget smaller than the global one ---
to demonstrate that $\toolname$ can effectively handle robustness constraints under various settings.
We conduct the experiments with pairs of global and local budget of $(1, 1)$, $(2, 1)$, $(2, 2)$, $(5, 1)$, $(5, 2)$, and $(5, 5)$, respectively.
Table~\ref{tab:graph_simple} summarizes the results,
and
Figure~\ref{fig:graph} gives a different view, showing how many instances for each budget can be completed as time increases.
For the full results for each distinguished budget, see Appendix~\ref{app:exp4}.

Graph classification is more challenging.
Due to this final aggregation layer,
we need to keep features and bounds up to date in each layer for every vertices in the graph.
Again, we find that $\toolname$ can handle up to four layers (excluding the final aggregation layer),
surpassing the state of the art.
Our method can in fact go beyond four layers, but we did not include a comparison since other solvers timed out on all examples.

\begin{table}[t]
\centering
\caption{Comparison results of $\toolname$ and $\scip$
for weak and general robustness on the MUTAG and ENZYMES datasets with various aggregation functions.
Note that $\scip$ only implements weak robustness for $\summ$ aggregation, thus other entries in the table for $\scip$ are left blank.
$t_{a}$ denotes the average runtime, 
and $t_{g}$ denotes the shifted geometric mean of the runtime.}
\label{tab:graph_simple}
\resizebox{\textwidth}{!}{
\begin{tblr}{
        rows    = {abovesep=0.2pt, belowsep=0.2pt},
        row{1}  = {font=\large, abovesep=3pt, belowsep=0pt},
        colspec = {l|l|r|rrrrrr|rrrrrr},
        cell{1}{4}  = {c=6}{c},
        cell{1}{10} = {c=6}{c},
        cell{2}{4}  = {c=3}{c},
        cell{2}{7}  = {c=3}{c},
        cell{2}{10} = {c=3}{c},
        cell{2}{13} = {c=3}{c}
    }
\toprule
& & & $\toolname$ & & & & & & $\scip$  \\
& & &
    All instances & & & Robust instances & & &
    All instances & & & Robust instances \\
\cmidrule[lr]{4-6}   \cmidrule[lr]{7-9}
\cmidrule[lr]{10-12} \cmidrule[lr]{13-15}
& & \#Instances &
\#Solved & $t_{a}(s)$ & $t_{g}(s)$ &
\#Solved & $t_{a}(s)$ & $t_{g}(s)$ &
\#Solved & $t_{a}(s)$ & $t_{g}(s)$ &
\#Solved & $t_{a}(s)$ & $t_{g}(s)$ \\
\midrule
$\summ$
&     MUTAG &  1,128 &  1,128 &   0.14 &   0.11 &     58 &   0.36 &   0.31 &    478 & 128.72 &  49.44 &      5 & 357.92 & 341.21 \\
(Weak)
&   ENZYMES &  3,600 &  3,385 &  15.94 &   4.10 &    559 &  38.20 &   9.47 &    672 & 113.98 &  50.65 &     48 & 188.68 & 104.27 \\
\midrule
$\summ$
&     MUTAG &  1,128 &  1,128 &   0.14 &   0.11 &     58 &   0.36 &   0.31 \\
&   ENZYMES &  3,600 &  3,556 &   3.66 &   1.04 &    160 &  16.95 &   5.08 \\
\midrule
$\maxx$
&     MUTAG &  1,128 &  1,128 &   0.01 &   0.01 &      3 &   0.04 &   0.04 \\
&   ENZYMES &  3,600 &  3,586 &   0.52 &   0.20 &    115 &   2.18 &   1.19 \\
\midrule
$\mean$
&     MUTAG &  1,128 &  1,094 &   5.03 &   1.56 &    224 &   2.32 &   1.98 \\
&   ENZYMES &  3,600 &  3,487 &   8.60 &   2.16 &    280 &  23.34 &   5.40 \\
\bottomrule
\end{tblr}
}
\end{table}

\begin{figure}[t]
\caption{The number of instances solved by each tool plotted against runtime under different aggregations and budgets.
The solid line denotes $\toolname$ and the dotted line denotes $\scip$.
The blue, orange, green, red, purple, and brown lines correspond
to pairs of global and local budget of $(1, 1)$, $(2, 1)$, $(2, 2)$, $(5, 1)$, $(5, 2)$, and $(5, 5)$, respectively.
Note that the $x$ axis is in logarithmic scale.}
\centering
\resizebox{\textwidth}{!}{
\begin{tabular}{c c c c c c c}
    & & MUTAG & ENZYMES & & MUTAG & ENZYMES \\
    \makecell{\rotatebox{90}{$\summ$}} & \makecell{\rotatebox{90}{(Weak)}}
    & \adjustbox{valign=c}{\includegraphics[width=0.25\linewidth]{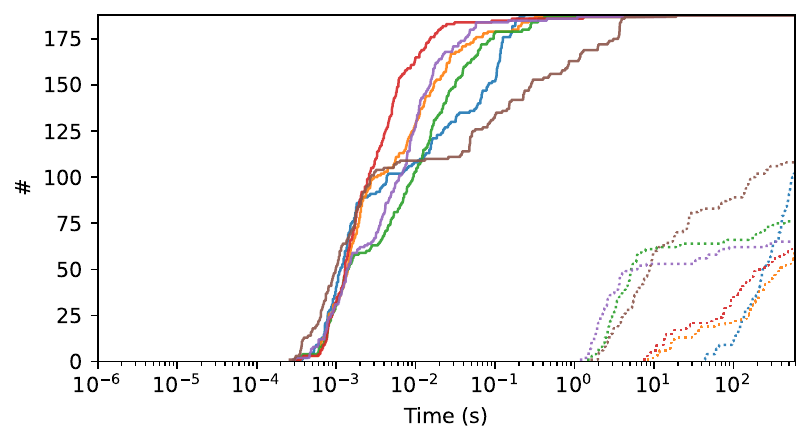}} 
    & \adjustbox{valign=c}{\includegraphics[width=0.25\linewidth]{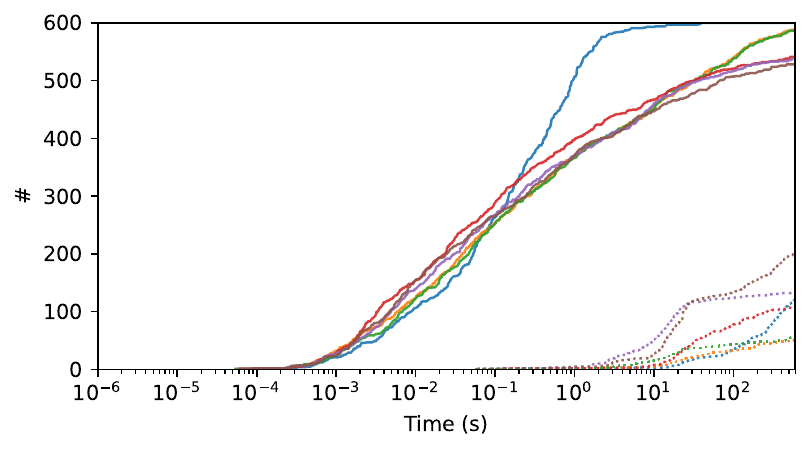}} &
    \makecell{\rotatebox{90}{$\maxx$}}
    & \adjustbox{valign=c}{\includegraphics[width=0.25\linewidth]{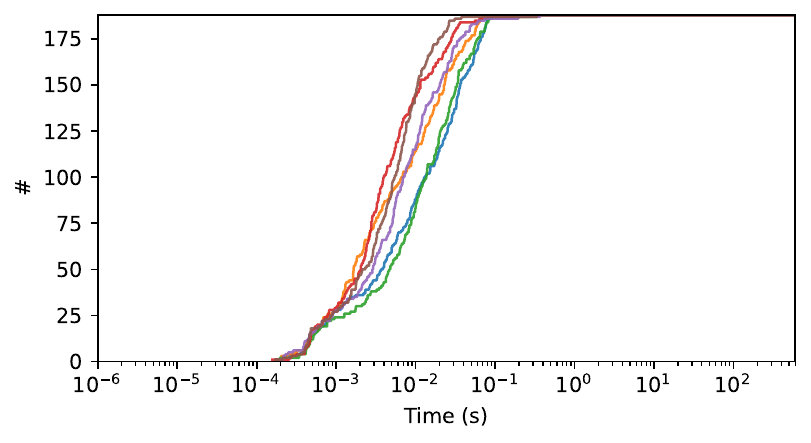}} 
    & \adjustbox{valign=c}{\includegraphics[width=0.25\linewidth]{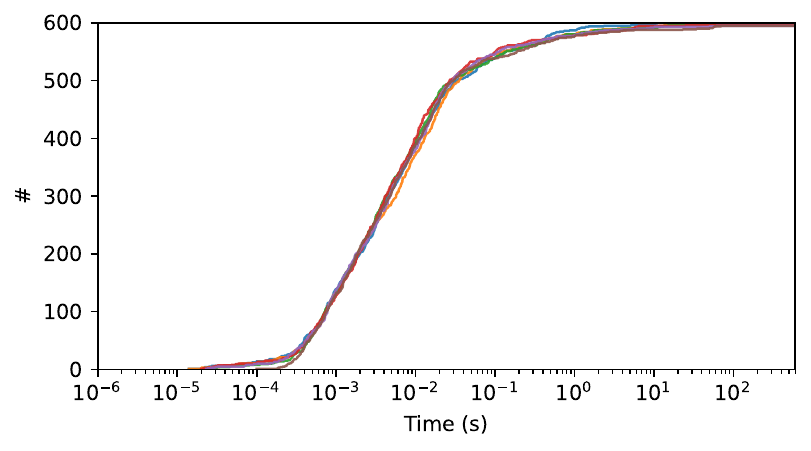}} \\
    \makecell{\rotatebox{90}{$\summ$}} &
    & \adjustbox{valign=c}{\includegraphics[width=0.25\linewidth]{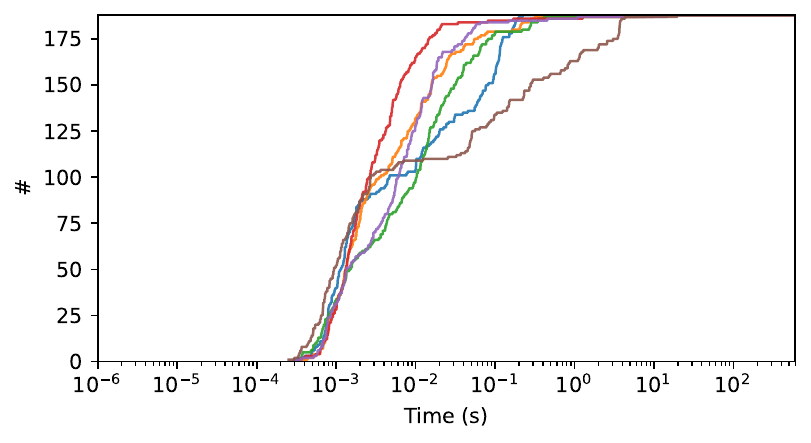}} 
    & \adjustbox{valign=c}{\includegraphics[width=0.25\linewidth]{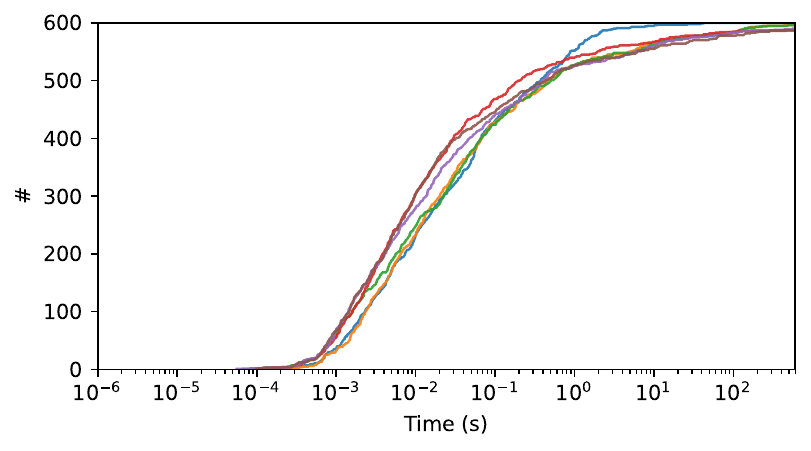}} &
    \makecell{\rotatebox{90}{$\mean$}}
    & \adjustbox{valign=c}{\includegraphics[width=0.25\linewidth]{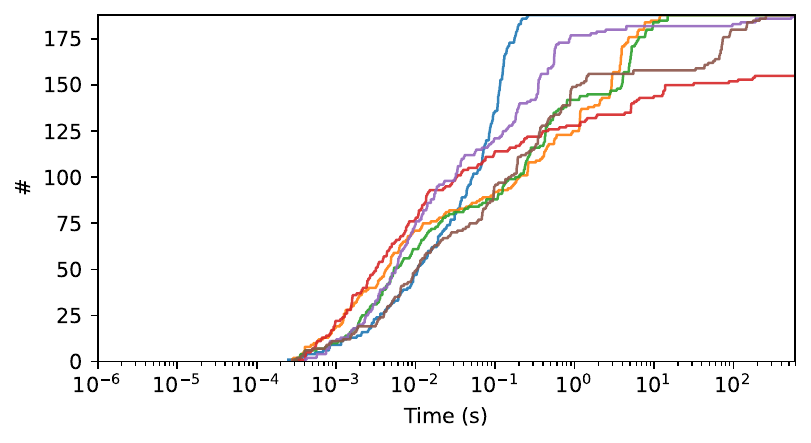}} 
    & \adjustbox{valign=c}{\includegraphics[width=0.25\linewidth]{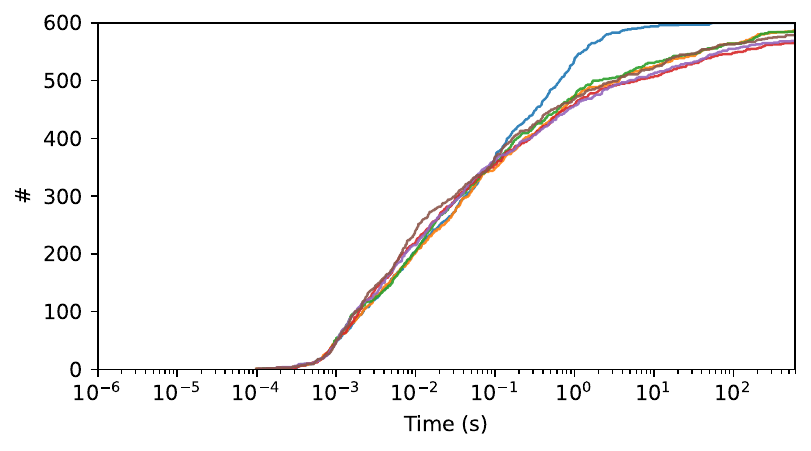}} \\
\end{tabular}
}
\label{fig:graph}
\end{figure}


\section{Additional related work} \label{sec:related}
A wide range of adversarial attacks can be considered in training machine learning systems, including neural ones: see e.g. \cite{robustbookchap}. Here we consider the case of attacks on test data at inference time. While our work considers exact methods for determining robustness, there are also incomplete methods \cite{ZugnerAG18,benf}, which will sometimes be able to find attacks or infer robustness but sometimes will be inconclusive.

The verification problems we study are straightforwardly decidable since they enumerate over a finite set of inputs.
In contrast \cite{SalzerL23,NunnSST24, BenediktLMT24} consider GNN static analysis problems, quantifying over all inputs.


\section{Conclusion}\label{sec:conc}

Adversarial robustness problems for discrete settings, such as graph learning, are examples of computationally hard problems, and it is thus natural that prior methods for attacking robustness work on top of solvers for classic hard problems, such as integer programming or SAT.
Surprisingly, we show that direct approaches that apply heuristic search on top of lightweight solvers can outperform the state of the art. 
This indicates that solvers still lack the ability to recognize and exploit structure inherent arising from problems in neural verification, and may prompt investigation in how to adapt general purpose solving tools with these applications in mind.

We note that our work does not consider perturbation to both the features and the edges. For feature perturbation, the graph neural network robustness problem is quite similar to the robustness problem for standard feedforward neural networks, and so it is natural to proceed via reduction to one of the tools available for analyzing feedforward networks, such as Marabou \cite{marabou}.
We also note that, while our work increases the range of robustness analysis for GNNs, the scalability of such systems is still extremely limited. We know of no tools that report exact analysis beyond GNNs with 4 layers and dimension of the hidden channel in each layer equal to 32.

\subsubsection{Acknowledgements}
We thank the anonymous reviewers for their insightful comments and suggestions.
We also thank Marta Kwiatkowska, Minghao Liu and Xiaowei Huang for insightful discussions.

\subsubsection{Data-Availability Statements}
The data and code used in this paper are available at \url{https://zenodo.org/records/17485304}.

\bibliographystyle{plain}
\bibliography{robust,minghaomarta}

\appendix


\section{Variant of the high-level algorithm for computing the radius of satisfaction}

To compute the radius of satisfaction, 
we can exhaustively searches for the radius
under the guidance of
the partial oracle.
There are four cases.
\begin{itemize}
\item 
If $\oracleres$ is $\unsat$,
then no normal graph satisfies the condition.
By definition, $d_m$ itself is the radius.
\item 
If $\oracleres$ is $\sat$ and $d_m = 0$,
then there exists a completion $\agraph'$ of $\incgraph$ with $\dist{\agraph', \agraph} = \dist{\incgraph, \agraph}$,
which implies that the radius is $-1$.
\item 
If $\oracleres = \sat$ and $d_m > 0$,
then the infimum 
must be strictly less than $d_m$. In this case the algorithm proceeds recursively with budget
$d_m - 1$.
\item 
If $\oracleres = \unknowntext$,
then 
the algorithm converts an unknown edge to either a non-edge or a normal edge and proceeds recursively.
\end{itemize}

We now formalize this by presenting a variant of the high-level algorithm 
for computing the radius  for satisfaction (Algorithm~\ref{algo:dfs_down}).
It exhaustively searches for the smallest $d\le d_m$ such that 
the partial oracle returns $\unsat$.
The manner in which the partial oracle is used is the same as in Algorithm~\ref{algo:dfs}.

\begin{algorithm}[ht!]
\caption{Algorithm for computing the radius of satisfaction.}
\label{algo:dfs_down}
\begin{algorithmic}[1]
        \Procedure{Search}{$\query$, $\cG$, $v$, $\cH$, $d_m$}
        \State $\oracleres \gets \oracle_{\query, \cG, v}(\cH, d_m)$.
        \If{$\oracleres$ is $\sat$}
        \If{$d_m$ is $0$}
        \State \Return $-1$
        \Else
        \State \Return \Call{Search}{$\query$, $\cG$, $v$, $\cH$, $d_m-1$}
        \EndIf
        \ElsIf{$\oracleres$ is $\unsat$}
        \State \Return $d_m$
        \Else
        \State Pick $e \in E_\unknown$.
        \If{$e$ is an edge in $\cG$}
        \State $\cH_1 \gets \cH$ by converting $e$ into a normal edge.
        \State $\cH_2 \gets \cH$ by converting $e$ into a non-edge.
        \Else
        \State $\cH_1 \gets \cH$ by converting $e$ into a non-edge.
        \State $\cH_2 \gets \cH$ by converting $e$ into a normal edge.
        \EndIf
        \State $d_1 \gets$ \Call{Search}{$\query$, $\cG$, $v$, $\cH_1$, $d_m$}
        \State $d_2 \gets$ \Call{Search}{$\query$, $\cG$, $v$, $\cH_2$, $d_m-1$}
        \State \Return $\min(d_1, d_2+1)$
        \EndIf
        \EndProcedure
\end{algorithmic}
\end{algorithm}

The correctness of Algorithm~\ref{algo:dfs_down} can be
established by induction on the sum $d_m+ |E^{\unknown}|$.

\begin{itemize}
\item 
If $\oracleres$ is $\unsat$,
then no normal graph satisfies the condition.
By definition, $d_m$ itself is the infimum.
\item 
If $\oracleres$ is $\sat$ and $d_m = 0$,
then there exists a completion $\agraph'$ of $\incgraph$ with $\dist{\agraph', \agraph} = \dist{\incgraph, \agraph}$,
which implies that there is no such infimum.
\end{itemize}
For the induction step:
\begin{itemize}
\item 
If $\oracleres = \sat$ and $d_m > 0$,
then the infimum 
must be strictly less than $d_m$.
By the induction hypothesis, it 
coincides with the result for $d_m - 1$.
\item 
If $\oracleres = \unknowntext$,
then the maximum budget is the minimum of the budgets obtained
from $\incgraph_1$ with parameter $d_m$ and from $\incgraph_2$ with parameter $d_m - 1$.
Our choice of $\incgraph_1$ and $\incgraph_2$ ensures that
for every completion $\agraph_1$ of $\incgraph_1$,
$\dist{\agraph_1, \incgraph_1} = \dist{\agraph', \incgraph}$,
and for every completion $\agraph_2$ of $\incgraph_2$,
$\dist{\agraph_2, \incgraph_2} = \dist{\agraph', \incgraph} + 1$.
Moreover, we have $\comp{\incgraph} = \comp{\incgraph_1} \cup \comp{\incgraph_2}$.
\end{itemize}

\section{Variants of algorithms for graph classification} \label{app:graphvariant}
Recall that in the paper body, we focused on GNNs as node classifiers.
However, we mentioned that GNNs can also be used as graph classifiers, by applying additional pooling and linear transformation layers
within a final layer.
We now explain this in detail, focusing on sum pooling, where the final outcome depends on the summation
of the last-layer features of each vertex.
Here, we provide the formal definitions and the adversarial robustness problem for this type of task.

\begin{definition}[Graph classifier induced by a GNN] 
    Let $\cA$ be an $L$-layer GNN and $\agraph\in\frG_{V,\ifeatmap}$.
    For the entire graph $\agraph$,
    the \emph{predicted class} of $\cA$, denoted by $\pred{\agraph}$, is
    \begin{equation*}
        \pred{\agraph}
        \ :=\ 
        \argmax_{1 \le i \le \gnndim{L+1}} \left(\coefC{L+1} \sum_{v \in V} \feat{L}_\agraph(v) + \coefb{L+1}\right)[i],
    \end{equation*}
    where $\cC{L+1} \in \bbR^{\gnndim{L+1} \times \gnndim{L}}$ and
    $\coefb{L+1} \in \bbR^{\gnndim{L+1}}$ are
    extra learnable coefficient matrix and bias vector.
\end{definition}

\begin{definition}[Adversarial robustness radius of graph classify GNNs]
    Let $\cA$ be an $L$-layer GNN,
    $\agraph$ a  graph,
    $F \subseteq V \times V$ a set of edges,
    $\Delta \in \bbN$ a global budget,
    and $\delta \in \bbN$ a local budget.
    For a class $1 \le c \le \gnndim{L}$,
    we define the \emph{adversarial robustness radius} of $\cA$ with respect to $c$ as follows:

    For the entire graph $\agraph$,
    we say that $\cA$ is \emph{adversarially robust} for $\agraph$ with class $c$
    if, for every perturbed normal graph $\agraph' \in \frQ(\agraph, F, \Delta, \delta)$,
    it holds that $\pred{\agraph'} = c$.
\end{definition}
Note that the condition in the above definition is equivalent to requiring that,
for every $1 \le c' \neq c \le \gnndim{L}$,
\begin{equation*}
    \left(\coefC{L+1}\sum_{v \in V}\feat{L}_{\agraph'}(v) + \coefb{L+1}\right)[c]
    \ \ge\ 
    \left(\coefC{L+1}\sum_{v \in V}\feat{L}_{\agraph'}(v) + \coefb{L+1}\right)[c'].
\end{equation*}
As with node classification,
we also consider a weaker notion of adversarial robustness,
in which the target class $c'$ is fixed in advance.

Regarding the partial oracle for graph classification,
the non-robustness tester is straightforward.
For the bound propagator, we can extend the bound propagation rules to apply to the pooling and linear layers.
Specifically, we define
\begin{equation*}
    \begin{aligned}
        \featu{L+1}_\incgraph\ =\ &\relaxu\left(\coefC{L+1}, \sum_{v \in V} \featu{L}_\incgraph(v), \sum_{v \in V} \featl{L}_\incgraph(v)\right) + \coefb{L+1} \\
        \featl{L+1}_\incgraph\ =\ &\relaxl\left(\coefC{L+1}, \sum_{v \in V} \featu{L}_\incgraph(v), \sum_{v \in V} \featl{L}_\incgraph(v)\right) + \coefb{L+1}.
    \end{aligned}
\end{equation*}
By applying Lemma~\ref{lemma:relax}, we can establish the desired correctness condition:
for every $\agraph' \subseteq \incgraph$,
\begin{equation*}
    \featl{L+1}_\incgraph
    \ \le\ 
    \coefC{L+1} \sum_{v \in V} \feat{L}_{\agraph'}(v) + \coefb{L+1}
    \ \le\ 
    \featu{L+1}_\incgraph.
\end{equation*}

\section{Variation of the algorithms for undirected graphs} \label{app:undirected}

Recall that in the body of the paper, we presented our results for a data model consisting of directed graphs, where the GNNs aggregated over incoming neighbors only. We mentioned that this model is commonly used in prior papers on GNN robustness, but that it is more common in the graph learning literature as a whole to work with undirected graphs, and with a GNN variant that aggregates over (undirected) neighbors.

We explain that very little needs to change for the undirected variant.
In general, the algorithm and oracle described in Section~\ref{sec:oracle} remain the same for both directed and undirected graphs.
The main adjustment lies in how features and bounds are propagated when the optimization strategy for incremental computation is applied.
For a directed graph, if we convert an edge from $u$ to $v$,
we only need to update the features and bounds of vertex $v$ at the first layer;
in the second layer, the update is propagated to the neighbors of $v$.
However, for an undirected GNN, we need to update both $u$ and $v$ at the first layer, and then propagate the updates to the neighbors of both $u$ and $v$ in subsequent layers.


\section{Missing proofs}
\label{app:missing-proof}

\subsection{Proof of coNP-completeness of robustness problem}

In this section, we will prove that checking the adversarial robustness of GNNs is coNP-complete.
For aggregation function $\summ$ and $\mean$ we show that
there is a GNN $\cA$ using $\summ$ ($\mean$, resp.) such that
the problem of checking whether $\cA$ is adversarially robust for a given node in a graph is coNP-complete.
For aggregation function $\maxx$, the GNN $\cA$ is part of the input.

Note that membership in coNP is straightforward for every GNN $\cA$,
so we will focus on the hardness.
All reductions are from the subset-sum problem:
Given a set of positive integers $S = \{s_1, s_2, \ldots, s_n\}$ and a positive integer $t$,
decide whether there exists a subset $S_0 \subseteq S$ such that the sum
of the integers in $S_0$ is exactly $t$.
This problem is known to be NP-complete.

\subsubsection{For aggregation function $\summ$}

Given an instance of subset-sum problem $S=\{s_1,\ldots,s_n\}$ and $t$ as above,
consider the following graph $\agraph$:
\begin{center}
\begin{tikzpicture}[every edge/.style = {->, >=stealth,draw,auto,shorten <=2pt,shorten >= 2pt},
 vrtx/.style args = {#1/#2}{circle, draw, fill=black, inner sep=1.5pt,label=#1:#2}]

\node(v) [vrtx=left/$(0)$] at (-0.5,0) {};
\node(vp) [vrtx=right/$v$] at (-0.5,0) {};

\node(u0) [vrtx=left/$(-t)$]at (-3,-2) {};
\node(u0p) [vrtx=right/$u_0$] at (-3,-2) {};

\node(u1) [vrtx=left/$(s_1)$]at (-1,-2) {};
\node(u1p) [vrtx=right/$u_1$] at (-1,-2) {};

\node (u2) at (1,-2) {$\cdots\cdots$};

\node(u3) [vrtx=left/$(s_n)$]at (3,-2) {};
\node(u3p) [vrtx=right/$u_n$] at (3,-2) {};

\path [->] (u0) edge (v);
\path [->] (u1) edge (v);
\path [->] (u3) edge (v);

\end{tikzpicture}
\end{center}
On each node, the left side is the feature value and the right side is the node name.
The initial feature of each vertex is $1$-dimensional,
where $X(u_i) = s_i$ for $1 \le i \le n$,
$X(u_{0}) = -t$ and $X(v) = 0$.
All edges are fragile edges and 
the global budget $\Delta=n$.

The GNN $\cA$ has two layers,
where $d^{(0)}=1$, $d^{(1)}=2$ and $d^{(2)}=2$.
On perturbed graph $\agraph' \in \frQ(\agraph, F, \Delta)$,
we describe the computation of $\cA$ on $\langle\agraph',v\rangle$ layer by layer.
In the first layer, the GNN $\cA$ computes $\feat{1}_{\agraph'}(v)$:
\begin{equation*}
    \feat{1}_{\agraph'}(v)
    \ =\ 
\relu\left(
\begin{array}{c}
\sum_{u \in \nbr_{\agraph'}(v)} X(u)
\\
\sum_{u \in \nbr_{\agraph'}(v)} -X(u)
\end{array}
\right)
\end{equation*}
In the second layer, the GNN $\cA$ computes $\feat{2}_{\agraph'}(v)$:
\begin{equation*}
    \feat{2}_{\agraph'}(v)
    \ =\ 
\relu\left(
    \begin{array}{c}
    1/2
    \\
    \feat{1}_{\agraph'}(v)[1]+\feat{1}_{\agraph'}(v)[2]
    \end{array}\right)
\end{equation*}
Note that $\feat{2}_{\agraph'}(v)[1]$ is always $1/2$.
Since all $s_1,\ldots,s_n$ and $t$ are positive integers,
we have $\feat{2}_{\agraph}(v)[2] > 1/2$.
Thus, $\pred{\agraph, v} = 2$.

We claim that $\cA$ is not adversarially robust for $v$ with class $1$
iff the subset-sum instance $S$ and $t$ has a solution.
To see this, suppose that the subset-sum instance $S$ and $t$ has a solution.
Suppose that $S_0 \subseteq S$ is such a solution, i.e.
$\sum_{s\in S_0} s = t$.
Consider $\agraph' \in \frQ(\agraph, F, \Delta)$ where the set of edges
$E'= \{(u_0,v)\}\cup \{(u_i,v) \mid s_i \in S_0\}$.
Then, we have $\feat{2}_{\agraph'}(v) = (1/2,0)$.
Thus, $\pred{\agraph', v} = 1$.

For the converse, we will use the identity that 
$\relu(a) + \relu(-a) = |a|$, for every real number $a\in \bbR$.
Suppose that $\cA$ is not adversarially robust for $v$ with class $1$.
Then, there is a perturbed graph $\agraph' \in \frQ(\agraph, F, \Delta)$,
such that $\pred{\agraph', v} = 1$.
By definition, $\feat{2}_{\agraph'}(v)[1] > \feat{2}_{\agraph'}(v)[2]$.
Since all $s_1,\ldots,s_n$ and $t$ are positive integers
and $\feat{2}_{\agraph'}(v)[1] = 1/2$,
we have $\feat{2}_{\agraph'}(v)[2] = 0$.
Since by the definition,
$$
\feat{2}_{\agraph'}(v)[2] \ = \ 
\relu\left(\sum_{u \in \nbr_{\agraph'}(v)} X(u)\right) + 
\relu\left(\sum_{u \in \nbr_{\agraph'}(v)} -X(u)\right)=
\left|\sum_{u \in \nbr_{\agraph'}(v)} X(u)\right|
$$
we have:
$$
\sum_{u \in \nbr_{\agraph'}(v)} X(u) \ = \ 0
$$
This implies that there is $S_0\subseteq S$
such that $t=\sum_{s\in S_0} s $.

\subsubsection{For aggregation function $\mean$}
The reduction is exactly the same as the case for $\summ$.
Given a set of positive integers 
$S=\{s_1,\ldots,s_n\}$ and a positive integer $t$,
we construct the graph $\agraph$, the set of fragile edges $F$ and the global budget $\Delta$ in a similar manner as in the previous case.
The difference is in the initial feature of 
each vertex,
where $X(u_i) = (n+1) s_i$ for $1 \le i \le n$,
$X(u_{0}) = -(n+1)t$ and $X(v) = 0$.

The GNN $\cA$ uses aggregation function $\mean$.
On perturbed graph $\agraph' \in \frQ(\agraph, F, \Delta)$,
it computes $\feat{1}_{\agraph'}(v)$ in the first layer:
\begin{equation*}
    \feat{1}_{\agraph'}(v)
    \ =\ 
\relu\left(
\begin{array}{c}
\mean\left(\msetc{X(u)}{u \in \nbr_{\agraph'}(v)}\right)
\\
\mean\left(\msetc{-X(u)}{u \in \nbr_{\agraph'}(v)}\right)
\end{array}
\right)
\end{equation*}
The second layer is exactly the same as in the case of $\summ$.
We observe that
for every subset $S_0\subseteq S$:
\begin{itemize}
\item 
If $t=\sum_{s\in S_0} s $,
then 
$$
\frac{(n+1)t}{|S_0|}-
\frac{\sum_{s\in S_0} (n+1)s}{|S_0|+1}
 \ =\ 
\frac{n+1}{|S_0|+1}\left(t- \sum_{s\in S_0} s\right) \ =\ 
0
$$
\item 
If $t\neq \sum_{s\in S_0} s$,
then 
$$
\left|\frac{(n+1)t}{|S_0|+1}-
\frac{\sum_{s\in S_0} (n+1)s}{|S_0|+1}\right|
 \ =\ 
\frac{n+1}{|S_0|+1}\left|t- \sum_{s\in S_0} s\right|
 \ \geq\ 1.
$$
\end{itemize}
To establish  correctness of the reduction,
we note that the term $\frac{n+1}{|S_0|+1}\left|t- \sum_{s\in S_0} s\right|$
is precisely $\feat{2}_{\agraph'}(v)[2]$,
where $\agraph' \in \frQ(\agraph, F, \Delta)$ is the perturbed graph.

\subsubsection{For aggregation function $\maxx$}

The reduction is also similar to the case for $\summ$, except that the GNN $\cA$ is no longer fixed.
Given an instance of subset-sum problem $S=\{s_1,\ldots,s_n\}$ and $t$ as above,
we construct the following graph $\agraph$:
\begin{center}
\begin{tikzpicture}[every edge/.style = {->, >=stealth,draw,auto,shorten <=2pt,shorten >= 2pt},
 vrtx/.style args = {#1/#2}{circle, draw, fill=black, inner sep=1.5pt,label=#1:#2}]

\node(v) [vrtx=left/$(0)$] at (-0.5,0) {};
\node(vp) [vrtx=right/$v$] at (-0.5,0) {};

\node(u0) [vrtx=left/$t\cdot e_{n+1}$]at (-5,-2) {};
\node(u0p) [vrtx=right/$u_{n+1}$] at (-5,-2) {};

\node(u1) [vrtx=left/$s_1\cdot e_1$]at (-1,-2) {};
\node(u1p) [vrtx=right/$u_1$] at (-1,-2) {};

\node (u2) at (1,-2) {$\cdots\cdots$};

\node(u3) [vrtx=left/$s_n\cdot e_n$]at (3,-2) {};
\node(u3p) [vrtx=right/$u_n$] at (3,-2) {};

\path [->] (u0) edge (v);
\path [->] (u1) edge (v);
\path [->] (u3) edge (v);

\end{tikzpicture}
\end{center}
where $e_i \in \bbR^{n+1}$ is the unit vector in $\bbR^{n+1}$ with $1$ in the $i$-th coordinate and $0$ elsewhere.
The set of fragile edges is $F=\{(v,u_1), (v,u_2), \ldots, (v,u_{n})\}$.
The global budget $\Delta=n$.

We construct the GNN $\cA$ with three layers,
where $d^{(0)}=n+1$, $d^{(1)}=n+1$, $d^{(2)}=2$ and $d^{(3)}=2$.
Note that the dimension $d^{(0)}=d^{(1)}=n+1$,
which depends on the subset-sum instance.
On perturbed graph $\agraph' \in \frQ(\agraph, F, \Delta)$,
we describe the computation of $\cA$ on $\langle\agraph',v\rangle$ layer by layer.
In the first layer, the GNN $\cA$ computes $\feat{1}_{\agraph'}(v)$:
\begin{equation*}
    \feat{1}_{\agraph'}(v)
    \ =\ 
\relu\
\left(
    \maxx \msetc{\feat{0}_{\agraph'}(u)}{u \in \nbr_{\agraph'}(v)}
\right)
\end{equation*}
In the second layer, the GNN $\cA$ computes $\feat{2}_{\agraph'}(v)$:
\begin{equation*}
    \feat{2}_{\agraph'}(v)
    \ =\ 
\relu\left(
    \begin{array}{c}
    \feat{1}_{\agraph'}(v)[n+1] \ -\ \sum_{i=1}^{n} \feat{1}_{\agraph'}(v)[i]
    \\
    \sum_{i=1}^{n} \feat{1}_{\agraph'}(v)[i] \ - \  \feat{1}_{\agraph'}(v)[n+1]
    \end{array}\right)
\end{equation*}
In the third layer, the GNN $\cA$ computes $\feat{3}_{\agraph'}(v)$:
\begin{equation*}
    \feat{3}_{\agraph'}(v)
    \ =\ 
\relu\left(
    \begin{array}{c}
    1/2
    \\
    \feat{2}_{\agraph'}(v)[1]+\feat{2}_{\agraph'}(v)[2]
    \end{array}\right)
\end{equation*}
Note that in the second and third layer in this GNN are similar to the first and second layer in the GNN for $\summ$.
The first layer in this case the GNN simply computes the vector $(b_1,\ldots,b_{n+1})$
that corresponds to a subset $S_0 \subseteq S$ where 
$b_i = s_i$ if $s_i \in S_0$ and $b_i = 0$ otherwise,
and $b_{n+1} = t$.

In a manner similar to the $\summ$ case, 
we can show that that t $\cA$ is not adversarially robust for $v$ with class $1$
iff the subset-sum instance $S$ and $t$ has a solution.

\subsection{Proof of Lemma~\ref{lemma:ground}}


\noindent
Recall Lemma~\ref{lemma:ground}:

\medskip

\noindent
\emph{For every incomplete graph $\incgraph \in \frH_{V, \ifeatmap}$
and graph $\agraph \in \frH_{V, \ifeatmap}$,
letting $\agraph'$ be the grounding of $\incgraph$ to $\agraph$,
 $\dist{\agraph', \agraph} = \dist{\incgraph, \agraph}$.}

\medskip

\begin{proof}
    For every $(v, u) \in V \times V$,
    if $(v, u)$ is inconsistent between $\incgraph$ and $\agraph$,
    then either $(v, u)$ is an edge in $\incgraph$ and a non-edge in $\agraph$, or
    $(v, u)$ is a non-edge in $\incgraph$ and an edge in $\agraph$.
    Note that by the definition of the grounding,
    if $(v, u)$ is an edge in $\incgraph$,
    then is is also an edge in $\agraph'$;
    if $(v, u)$ is a non-edge in $\incgraph$,
    then is is also a non-edge in $\agraph'$.
    Thus, for both cases, $(v, u)$ is inconsistent between $\agraph'$ and $\agraph$.

    On the other hand,
    if $(v, u)$ is consistent between $\incgraph$ and $\agraph$,
    then either $(v, u)$ is a unknown edge in $\incgraph$,
    or $(v, u)$ is the same type of edge in both $\incgraph$ and $\agraph$.
    For both cases, by the definition of grounding,
    $(v, u)$ are consistent between $\agraph'$ and $\agraph$.
    Thus, $\dist{\agraph', \agraph} = \dist{\incgraph, \agraph}$.
\end{proof}

\subsection{Proof of Lemma~\ref{lemma:relax}}

Recall Lemma~\ref{lemma:relax}:

\medskip

\noindent
For every graph $\agraph \in \frH_{V, \ifeatmap}$ and
    every set $E_p$ of pairs of nodes, $\agraph$ is a completion of its relaxation with respect to $E_p$, and hence
    $\dist{\incgraph_{\agraph, E_p}, \agraph} = 0$.

\medskip

\begin{proof}
    For every $(v, u) \in V \times V$,
    by the definition of relaxation,
    one of the following three cases hold.
    \begin{itemize}
        \item $(v, u)$ is an edge in $\cG$ and an edge $\incgraph_{\agraph, E_p}$.
        \item $(v, u)$ is a non-edge in $\cG$ and a non-edge $\incgraph_{\agraph, E_p}$.
        \item $(v, u) \in E_p$ and it is a unknown edge $\incgraph_{\agraph, E_p}$.
    \end{itemize}
    For all cases, $(v, u)$ is consistent between
    $\incgraph_{\agraph, E_p}$ and $\agraph$.
    Therefore, $\agraph$ is a completion, and the distance between them is zero.
\end{proof}

\subsection{Proof of the correctness of Algorithm~\ref{algo:dfs}}
Recall the high-level algorithm, Algorithm ~\ref{algo:dfs}, for solving the $d$ radius satisfaction problem using a partial oracle.
We will show that the algorithm is correct by induction on $|E^\unknown|$.

The base case is $|E^{\unknown}|=0$.
In this case, by definition, the oracle always outputs $\sat$ or $\unsat$ and the correctness follows immediately.

For the induction step, we note that the number of unknown edges in $\incgraph_1$ and $\incgraph_2$ decreases by $1$.
Thus, the correctness follows immediately from the induction hypothesis.

\subsection{Proof of Lemma~\ref{lem:matrix}}

Recall Lemma \ref{lem:matrix}:

\medskip

\noindent
\emph{For every $\bfA \in \bbR^{m \times n}$ and $\bfvu, \bfv, \bfvl \in \bbR^n$
    with $\bfvl \le \bfv \le \bfvu$,
    it holds that
    \begin{equation*}
        \relaxl\left(\bfA, \bfvu, \bfvl\right)
        \ \le\ 
        \bfA \cdot \bfv
        \ \le\ 
        \relaxu\left(\bfA, \bfvu, \bfvl\right).
    \end{equation*}
}

\medskip

\begin{proof}
    For $1 \le i \le m$, note that
    \begin{equation*}
        \begin{aligned}
        \left(\bfA \cdot \bfv\right)_i
        \ =\ &
        \sum_{1 \le j \le n} \left(\max\left(\bfA_{i, j}, 0\right) + \min\left(\bfA_{i, j}, 0\right)\right) \bfv_j \\
        \ \le\ &
        \sum_{1 \le j \le n} \left(\max\left(\bfA_{i, j}, 0\right)\bfvu_j\right) +
        \left(\min\left(\bfA_{i, j}, 0\right)\bfvl_j\right) \\
        \ =\ &
        \left(\bfA^+\cdot\bfvu\right)_i + \left(\bfA^-\cdot\bfvl\right)_i \\
        \end{aligned}
    \end{equation*}
    The lower bound can be proved analogously.
\end{proof}

\subsection{Proof of Lemma \ref{lem:boundcorrect}}

Recall Lemma \ref{lem:boundcorrect}:

\medskip

\noindent
\emph{For every completion $\agraph'$ of $\incgraph$,
    $0 \le \ell \le L$, 
    and vertex $v \in V$,
    it holds that
    \begin{equation} \label{eq:bound}.
        \featl{\ell}_\incgraph(v)
        \ \le\ 
        \feat{\ell}_{\agraph'}(v)
        \ \le\ 
        \featu{\ell}_{\incgraph}(v)
    \end{equation}
}

\medskip

\begin{proof}
    The proof is by induction on layers.
    For the base case $\ell = 0$,
    notice that for every completion $\agraph'$ of $\incgraph$,
    $\feat{0}_{\agraph'}(v) = X(v)$.
    Thus, it is clear that \eqref{eq:bound} holds.

    For the induction step $1 \le \ell \le L$,
    it suffices to prove the relaxation for the aggregation functions:
    \begin{equation}
        \label{eq:aggr}
        \aggrl\left( \setsl^\norm, \setsl^\unknown \right)
        \ \le\ 
        \aggr\left(\msetc{\feat{\ell-1}_\agraph(u)}{u \in \nbr_\agraph(v)}\right)
        \ \le\ 
        \aggru\left( \setsu^\norm, \setsu^\unknown \right).
    \end{equation}
    The remainder then follows from Lemma~\ref{lem:matrix}.

    Since $\agraph'$ is a completion of $\incgraph$, we have
    \begin{equation*}
        \nbr^\norm_{\incgraph}(v) \subseteq \nbr_{\agraph'}(v) \subseteq \nbr^\norm_{\incgraph}(v) \cup \nbr^\unknown_{\incgraph}(v).
    \end{equation*}
    The proof for the lower bound is symmetric; here we focus only on the upper bound.
    \begin{itemize}
        \item Suppose that $\aggr$ is $\summ$. Then
            \begin{equation*}
                \begin{aligned}
                    &\summ\left(\msetc{\feat{\ell-1}_\agraph(u)}{u \in \nbr_\agraph(v)}\right) \\
                    \ =\ &                    
                    \sum_{u \in \nbr_{\agraph'}(v)} \feat{\ell-1}_{\agraph'}(u) \\
                    \ \le\ &
                    \sum_{u \in \nbr^\norm_{\incgraph}(v)} \featu{\ell-1}_{\incgraph}(u) +
                    \sum_{u \in \nbr_{\agraph'}(v) \cap \nbr^\unknown_{\incgraph}(v)} \featu{\ell-1}_{\incgraph}(u) \\
                    \ \le\ &
                    \sum_{u \in \nbr^\norm_{\incgraph}(v)} \featu{\ell-1}_{\incgraph}(u) +
                    \sum_{u \in \nbr^\unknown_{\incgraph}(v)} \max\left(\featu{\ell-1}_{\incgraph}(u), 0\right) \\
                    \ =\ &
                    \sumu\left( \setsu^\norm, \setsu^\unknown \right).
                \end{aligned}
            \end{equation*}
            Note that by the induction hypothesis,
            $\feat{\ell-1}_{\agraph'}(u) \le \featu{\ell-1}_{\incgraph}(u)$.

        \item Suppose that $\aggr$ is $\maxx$.
            Inequality~\eqref{eq:aggr} 
            can be established by a straightforward case analysis.

        \item Suppose that $\aggr$ is $\mean$.
            Let $k$ be the cardinality of the set $\nbr_{\agraph'}(v) \cap \nbr^\unknown_{\incgraph}(v)$,
            and let $s_1, \ldots, s_k$ be the $k$-largest element in $\bfsu^\unknown$.
            \begin{equation*}
                \begin{aligned}
                    \sum_{u \in \nbr_{\agraph'}(v)} \feat{\ell-1}_{\agraph'}(u)
                    \ \le\ &
                    \sum_{u \in \nbr^\norm_{\incgraph}(v)} \featu{\ell-1}_{\incgraph}(u) +
                    \sum_{u \in \nbr_{\agraph'}(v) \cap \nbr^\unknown_{\incgraph}(v)} \featu{\ell-1}_{\incgraph}(u) \\
                    \ \le\ &
                    \sum_{u \in \nbr^\norm_{\incgraph}(v)} \featu{\ell-1}_{\incgraph}(u) +
                    \sum_{1 \le j \le k} s_j
                \end{aligned}
            \end{equation*}
            Consequently,
            \begin{equation*}
                \begin{aligned}
                    &\mean\left(\msetc{\feat{\ell-1}_\agraph(u)}{u \in \nbr_\agraph(v)}\right) \\
                    \ \le\ &                    
                    \left(
                        \sum_{u \in \nbr^\norm_{\incgraph}(v)} \featu{\ell-1}_{\incgraph}(u) +
                        \sum_{1 \le j \le k} s_j
                    \right) / \left(\abs{\nbr^\norm_{\incgraph}(v)} + k\right) \\
                    \ \le\ &
                    \meanu\left( \setsu^\norm, \setsu^\unknown \right).
                \end{aligned}
            \end{equation*}
    \end{itemize}
\end{proof}

\section{Budge-specific experimental results }

Recall that tables \ref{tab:node_simple} and \ref{tab:graph_simple} in the main text reports results summed over different budgets.
Here, we provide additional details on the experimental results for each individual budget.

\subsection{End to end performance on node classification}
\label{app:exp1}

Table~\ref{tab:node_all_1} to Table~\ref{tab:node_all_10} present detailed comparison results of $\toolname$, $\gnnev$, and $\scip$
for both weak and general robustness on the Cora, CiteSeer, Cornell, Texas, and Wisconsin datasets,
using various aggregation functions and perturbation budgets.
All experiments are conducted on node classification tasks, 
where the fragile set of edges is restricted to deletions only,
and the local budget is set equal to the global budget.
Note that $\scip$ only supports weak robustness for the $\summ$ aggregation.
$t_{a}$ denotes the average runtime, 
and $t_{g}$ denotes the shifted geometric mean of the runtime.

\begin{table}[t]
\centering
\caption{Detailed comparison results of $\toolname$, $\gnnev$, and $\scip$
for weak and general robustness on the Cora, CiteSeer, Cornell, Texas, and Wisconsin datasets with various aggregation functions with global and local budgets $\Delta = 1$.
Note that $\scip$ only implements weak robustness for $\summ$ aggregation.
$t_{a}$ denotes the average runtime, 
and $t_{g}$ denotes the shifted geometric mean of the runtime.}
\label{tab:node_all_1}
\resizebox{\textwidth}{!}{
\begin{tblr}{
        rows    = {abovesep=0.2pt, belowsep=0.2pt},
        row{1}  = {font=\large, abovesep=3pt, belowsep=0pt},
        colspec = {l|l|r|rrrrrr|rrrrrr|rrrrrr},
        cell{1}{4}  = {c=6}{c},
        cell{1}{10} = {c=6}{c},
        cell{1}{16} = {c=6}{c},
        cell{2}{4}  = {c=3}{c},
        cell{2}{7}  = {c=3}{c},
        cell{2}{10} = {c=3}{c},
        cell{2}{13} = {c=3}{c},
        cell{2}{16} = {c=3}{c},
        cell{2}{19} = {c=3}{c}
    }
\toprule
& & & $\toolname$ & & & & & & $\gnnev$ & & & & & & $\scip$ \\
& & &
    All instances & & & Robust instances & & &
    All instances & & & Robust instances & & &
    All instances & & & Robust instances \\
\cmidrule[lr]{4-6}   \cmidrule[lr]{7-9}
\cmidrule[lr]{10-12} \cmidrule[lr]{13-15}
\cmidrule[lr]{16-18} \cmidrule[lr]{19-21}
& & \#Instances &
\#Solved & $t_{a}(s)$ & $t_{g}(s)$ &
\#Solved & $t_{a}(s)$ & $t_{g}(s)$ &
\#Solved & $t_{a}(s)$ & $t_{g}(s)$ &
\#Solved & $t_{a}(s)$ & $t_{g}(s)$ &
\#Solved & $t_{a}(s)$ & $t_{g}(s)$ &
\#Solved & $t_{a}(s)$ & $t_{g}(s)$ \\
\midrule
$\summ$
&      Cora &  2,708 &  2,708 &   0.01 &   0.01 &  2,514 &   0.01 &   0.01 &  2,228 &  21.39 &  11.19 &  2,034 &  22.36 &  11.33 &  2,699 &  27.19 &  13.12 &  2,541 &  25.98 &  12.58 \\
(Weak)
&  CiteSeer &  3,312 &  3,312 &   0.01 &   0.01 &  3,107 &   0.01 &   0.01 &  3,121 &  13.32 &   7.40 &  2,916 &  13.25 &   7.10 &  3,305 &   3.16 &   1.67 &  3,126 &   3.09 &   1.62 \\
&   Cornell &    183 &    183 &   0.01 &   0.01 &    137 &   0.01 &   0.01 &    163 &  14.72 &   8.48 &    117 &  17.83 &   9.51 &    183 &  14.14 &   5.66 &    143 &  14.70 &   5.80 \\
&     Texas &    183 &    183 &   0.01 &   0.01 &    163 &   0.01 &   0.01 &    173 &  13.89 &   7.23 &    153 &  13.74 &   6.67 &    178 &  18.12 &   6.41 &    162 &  17.35 &   5.60 \\
& Wisconsin &    251 &    251 &   0.01 &   0.01 &    228 &   0.01 &   0.01 &    223 &  18.87 &  10.33 &    201 &  19.10 &   9.97 &    243 &  28.36 &   9.84 &    228 &  28.58 &   9.76 \\
\midrule
$\summ$
&      Cora &  2,708 &  2,708 &   0.01 &   0.01 &  2,126 &   0.01 &   0.01 &  2,289 &  20.23 &  11.23 &  1,714 &  21.34 &  10.79 \\
&  CiteSeer &  3,312 &  3,312 &   0.01 &   0.01 &  2,791 &   0.01 &   0.01 &  3,173 &  14.45 &   8.17 &  2,652 &  14.71 &   7.62 \\
&   Cornell &    183 &    183 &   0.01 &   0.01 &    110 &   0.01 &   0.01 &    178 &  17.81 &   9.60 &    105 &  22.62 &  10.40 \\
&     Texas &    183 &    183 &   0.01 &   0.01 &    106 &   0.01 &   0.01 &    178 &  11.36 &   7.04 &    101 &  13.45 &   7.34 \\
& Wisconsin &    251 &    251 &   0.01 &   0.01 &    189 &   0.01 &   0.01 &    242 &  18.82 &  11.17 &    181 &  18.39 &   9.88 \\
\midrule
$\maxx$
&      Cora &  2,708 &  2,708 &   0.01 &   0.01 &  2,179 &   0.01 &   0.01 &  2,177 & 123.36 &  42.39 &  1,685 & 117.38 &  36.42 \\
&  CiteSeer &  3,312 &  3,312 &   0.01 &   0.01 &  2,774 &   0.01 &   0.01 &  3,042 & 100.75 &  23.76 &  2,518 &  92.96 &  19.80 \\
&   Cornell &    183 &    183 &   0.01 &   0.01 &    112 &   0.01 &   0.01 &    153 &  42.42 &  19.49 &     85 &  39.47 &  14.83 \\
&     Texas &    183 &    183 &   0.01 &   0.01 &    151 &   0.01 &   0.01 &    166 &  87.22 &  18.58 &    138 &  74.72 &  13.44 \\
& Wisconsin &    251 &    251 &   0.01 &   0.01 &    202 &   0.01 &   0.01 &    200 & 135.63 &  29.91 &    156 & 132.31 &  23.21 \\
\midrule
$\mean$
&      Cora &  2,708 &  2,708 &   0.01 &   0.01 &  2,173 &   0.01 &   0.01 &  2,147 &  22.13 &  12.97 &  1,622 &  23.21 &  12.66 \\
&  CiteSeer &  3,312 &  3,312 &   0.01 &   0.01 &  2,744 &   0.01 &   0.01 &  3,095 &  13.50 &   7.88 &  2,528 &  13.51 &   7.18 \\
&   Cornell &    183 &    183 &   0.01 &   0.01 &    130 &   0.01 &   0.01 &    159 &  17.99 &  10.30 &    106 &  20.72 &  10.66 \\
&     Texas &    183 &    183 &   0.01 &   0.01 &    127 &   0.01 &   0.01 &    174 &  15.29 &   8.46 &    118 &  15.41 &   7.95 \\
& Wisconsin &    251 &    251 &   0.01 &   0.01 &    186 &   0.01 &   0.01 &    217 &  25.26 &  12.62 &    153 &  23.35 &  11.19 \\
\bottomrule
\end{tblr}
}
\end{table}

\begin{table}[t]
\centering
\caption{Detailed comparison results of $\toolname$, $\gnnev$, and $\scip$
for weak and general robustness on the Cora, CiteSeer, Cornell, Texas, and Wisconsin datasets with various aggregation functions with global and local budgets $\Delta = 2$.
Note that $\scip$ only implements weak robustness for $\summ$ aggregation.
$t_{a}$ denotes the average runtime, 
and $t_{g}$ denotes the shifted geometric mean of the runtime.}
\label{tab:node_all_2}
\resizebox{\textwidth}{!}{
\begin{tblr}{
        rows    = {abovesep=0.2pt, belowsep=0.2pt},
        row{1}  = {font=\large, abovesep=3pt, belowsep=0pt},
        colspec = {l|l|r|rrrrrr|rrrrrr|rrrrrr},
        cell{1}{4}  = {c=6}{c},
        cell{1}{10} = {c=6}{c},
        cell{1}{16} = {c=6}{c},
        cell{2}{4}  = {c=3}{c},
        cell{2}{7}  = {c=3}{c},
        cell{2}{10} = {c=3}{c},
        cell{2}{13} = {c=3}{c},
        cell{2}{16} = {c=3}{c},
        cell{2}{19} = {c=3}{c}
    }
\toprule
& & & $\toolname$ & & & & & & $\gnnev$ & & & & & & $\scip$ \\
& & &
    All instances & & & Robust instances & & &
    All instances & & & Robust instances & & &
    All instances & & & Robust instances \\
\cmidrule[lr]{4-6}   \cmidrule[lr]{7-9}
\cmidrule[lr]{10-12} \cmidrule[lr]{13-15}
\cmidrule[lr]{16-18} \cmidrule[lr]{19-21}
& & \#Instances &
\#Solved & $t_{a}(s)$ & $t_{g}(s)$ &
\#Solved & $t_{a}(s)$ & $t_{g}(s)$ &
\#Solved & $t_{a}(s)$ & $t_{g}(s)$ &
\#Solved & $t_{a}(s)$ & $t_{g}(s)$ &
\#Solved & $t_{a}(s)$ & $t_{g}(s)$ &
\#Solved & $t_{a}(s)$ & $t_{g}(s)$ \\
\midrule
$\summ$
&      Cora &  2,708 &  2,708 &   0.01 &   0.01 &  2,334 &   0.01 &   0.01 &  1,946 &  43.89 &  16.61 &  1,602 &  46.57 &  16.11 &  2,560 &  75.19 &  31.23 &  2,197 &  75.28 &  29.51 \\
(Weak)
&  CiteSeer &  3,312 &  3,312 &   0.01 &   0.01 &  3,014 &   0.01 &   0.01 &  3,100 &  18.01 &   8.47 &  2,803 &  17.80 &   7.81 &  3,273 &  13.31 &   4.41 &  2,984 &  12.56 &   4.00 \\
&   Cornell &    183 &    183 &   0.01 &   0.01 &    107 &   0.01 &   0.01 &    155 &  21.54 &  10.04 &     80 &  22.39 &   8.30 &    176 &  38.92 &  11.74 &    103 &  38.22 &  10.12 \\
&     Texas &    183 &    183 &   0.01 &   0.01 &    146 &   0.01 &   0.01 &    150 &  19.75 &   7.25 &    117 &   8.41 &   3.22 &    170 &  30.10 &   9.78 &    138 &  22.83 &   6.24 \\
& Wisconsin &    251 &    251 &   0.01 &   0.01 &    211 &   0.01 &   0.01 &    190 &  23.38 &  10.25 &    154 &  20.37 &   8.10 &    213 &  50.25 &  17.05 &    181 &  45.03 &  14.61 \\
\midrule
$\summ$
&      Cora &  2,708 &  2,708 &   0.01 &   0.01 &  1,777 &   0.01 &   0.01 &  1,920 &  29.23 &  12.45 &  1,101 &  22.79 &   7.57 \\
&  CiteSeer &  3,312 &  3,312 &   0.01 &   0.01 &  2,585 &   0.01 &   0.01 &  3,097 &  15.88 &   8.40 &  2,372 &  15.06 &   6.87 \\
&   Cornell &    183 &    183 &   0.01 &   0.01 &     79 &   0.01 &   0.01 &    161 &  16.05 &   8.89 &     65 &  14.82 &   5.97 \\
&     Texas &    183 &    183 &   0.01 &   0.01 &     75 &   0.01 &   0.01 &    156 &  12.33 &   6.36 &     58 &   2.16 &   1.42 \\
& Wisconsin &    251 &    251 &   0.01 &   0.01 &    166 &   0.01 &   0.01 &    200 &  25.36 &  10.57 &    128 &  17.53 &   6.30 \\
\midrule
$\maxx$
&      Cora &  2,708 &  2,708 &   0.01 &   0.01 &  1,849 &   0.01 &   0.01 &  1,564 &  76.71 &  23.31 &    955 &  27.85 &   7.83 \\
&  CiteSeer &  3,312 &  3,312 &   0.01 &   0.01 &  2,615 &   0.01 &   0.01 &  2,705 &  61.50 &  19.31 &  2,078 &  34.02 &  12.06 \\
&   Cornell &    183 &    183 &   0.01 &   0.01 &     99 &   0.01 &   0.01 &    133 &  43.23 &  18.05 &     64 &  30.42 &   9.95 \\
&     Texas &    183 &    183 &   0.01 &   0.01 &    138 &   0.01 &   0.01 &    134 &  46.85 &  10.92 &    102 &   3.99 &   2.66 \\
& Wisconsin &    251 &    251 &   0.01 &   0.01 &    189 &   0.01 &   0.01 &    139 &  45.33 &  12.88 &    106 &  20.25 &   6.52 \\
\midrule
$\mean$
&      Cora &  2,708 &  2,708 &   0.01 &   0.01 &  1,816 &   0.01 &   0.01 &  1,689 &  27.39 &  11.70 &    934 &  22.46 &   6.66 \\
&  CiteSeer &  3,312 &  3,312 &   0.01 &   0.01 &  2,591 &   0.01 &   0.01 &  2,888 &  16.39 &   8.05 &  2,195 &  15.13 &   6.37 \\
&   Cornell &    183 &    183 &   0.01 &   0.01 &    114 &   0.01 &   0.01 &    131 &  10.90 &   6.76 &     71 &   9.62 &   4.55 \\
&     Texas &    183 &    183 &   0.01 &   0.01 &    104 &   0.01 &   0.01 &    149 &  17.00 &   7.21 &     78 &  11.72 &   3.91 \\
& Wisconsin &    251 &    251 &   0.01 &   0.01 &    167 &   0.01 &   0.01 &    172 &  22.54 &   9.47 &    100 &  18.32 &   4.98 \\
\bottomrule
\end{tblr}
}
\end{table}

\begin{table}[t]
\centering
\caption{Detailed comparison results of $\toolname$, $\gnnev$, and $\scip$
for weak and general robustness on the Cora, CiteSeer, Cornell, Texas, and Wisconsin datasets with various aggregation functions with global and local budgets $\Delta = 5$.
Note that $\scip$ only implements weak robustness for $\summ$ aggregation.
$t_{a}$ denotes the average runtime, 
and $t_{g}$ denotes the shifted geometric mean of the runtime.}
\label{tab:node_all_5}
\resizebox{\textwidth}{!}{
\begin{tblr}{
        rows    = {abovesep=0.2pt, belowsep=0.2pt},
        row{1}  = {font=\large, abovesep=3pt, belowsep=0pt},
        colspec = {l|l|r|rrrrrr|rrrrrr|rrrrrr},
        cell{1}{4}  = {c=6}{c},
        cell{1}{10} = {c=6}{c},
        cell{1}{16} = {c=6}{c},
        cell{2}{4}  = {c=3}{c},
        cell{2}{7}  = {c=3}{c},
        cell{2}{10} = {c=3}{c},
        cell{2}{13} = {c=3}{c},
        cell{2}{16} = {c=3}{c},
        cell{2}{19} = {c=3}{c}
    }
\toprule
& & & $\toolname$ & & & & & & $\gnnev$ & & & & & & $\scip$ \\
& & &
    All instances & & & Robust instances & & &
    All instances & & & Robust instances & & &
    All instances & & & Robust instances \\
\cmidrule[lr]{4-6}   \cmidrule[lr]{7-9}
\cmidrule[lr]{10-12} \cmidrule[lr]{13-15}
\cmidrule[lr]{16-18} \cmidrule[lr]{19-21}
& & \#Instances &
\#Solved & $t_{a}(s)$ & $t_{g}(s)$ &
\#Solved & $t_{a}(s)$ & $t_{g}(s)$ &
\#Solved & $t_{a}(s)$ & $t_{g}(s)$ &
\#Solved & $t_{a}(s)$ & $t_{g}(s)$ &
\#Solved & $t_{a}(s)$ & $t_{g}(s)$ &
\#Solved & $t_{a}(s)$ & $t_{g}(s)$ \\
\midrule
$\summ$
&      Cora &  2,708 &  2,708 &   0.04 &   0.04 &  2,160 &   0.05 &   0.04 &  1,656 &  23.19 &   9.71 &  1,251 &  21.62 &   7.76 &  1,937 &  65.49 &  22.14 &  1,539 &  58.14 &  17.60 \\
(Weak)
&  CiteSeer &  3,312 &  3,312 &   0.20 &   0.07 &  2,929 &   0.22 &   0.08 &  2,990 &  14.22 &   7.36 &  2,620 &  12.62 &   6.17 &  3,208 &  19.14 &   5.85 &  2,832 &  16.65 &   4.78 \\
&   Cornell &    183 &    183 &   0.01 &   0.01 &     86 &   0.01 &   0.01 &    151 &  14.22 &   7.68 &     67 &  11.78 &   5.25 &    167 &  26.83 &   7.28 &     83 &  19.86 &   5.22 \\
&     Texas &    183 &    183 &   0.23 &   0.12 &    136 &   0.31 &   0.16 &    146 &  11.81 &   5.37 &    112 &   6.33 &   2.64 &    152 &  28.42 &   8.06 &    119 &  11.81 &   3.01 \\
& Wisconsin &    251 &    251 &   0.94 &   0.44 &    194 &   1.21 &   0.57 &    175 &  15.06 &   7.34 &    136 &  10.56 &   4.92 &    193 &  40.52 &  12.64 &    158 &  32.80 &   9.92 \\
\midrule
$\summ$
&      Cora &  2,708 &  2,708 &   0.02 &   0.02 &  1,470 &   0.03 &   0.03 &  1,913 &  22.52 &  10.56 &    962 &  12.26 &   4.33 \\
&  CiteSeer &  3,312 &  3,312 &   0.24 &   0.08 &  2,446 &   0.32 &   0.11 &  3,034 &  13.36 &   7.55 &  2,201 &   9.43 &   5.02 \\
&   Cornell &    183 &    183 &   0.01 &   0.01 &     69 &   0.01 &   0.01 &    165 &  12.31 &   7.93 &     61 &   8.99 &   4.55 \\
&     Texas &    183 &    183 &   0.02 &   0.02 &     63 &   0.05 &   0.05 &    159 &   8.39 &   5.34 &     57 &   1.26 &   1.09 \\
& Wisconsin &    251 &    251 &   0.97 &   0.40 &    149 &   1.62 &   0.65 &    205 &  24.15 &  10.71 &    119 &  12.11 &   4.81 \\
\midrule
$\maxx$
&      Cora &  2,708 &  2,708 &   0.12 &   0.10 &  1,549 &   0.21 &   0.17 &  1,645 &  79.28 &  24.68 &    878 &  12.34 &   4.68 \\
&  CiteSeer &  3,312 &  3,303 &   0.63 &   0.22 &  2,478 &   0.82 &   0.29 &  2,621 &  54.70 &  16.38 &  1,958 &  21.08 &   8.77 \\
&   Cornell &    183 &    183 &   0.03 &   0.02 &     90 &   0.05 &   0.05 &    133 &  38.32 &  16.57 &     59 &  18.16 &   7.04 \\
&     Texas &    183 &    183 &   0.47 &   0.24 &    131 &   0.65 &   0.33 &    124 &  22.40 &   7.32 &     99 &   5.73 &   2.61 \\
& Wisconsin &    251 &    251 &   5.17 &   1.16 &    176 &   7.13 &   1.57 &    140 &  47.65 &  12.06 &    104 &   8.66 &   4.77 \\
\midrule
$\mean$
&      Cora &  2,708 &  2,708 &   0.01 &   0.01 &  1,460 &   0.01 &   0.01 &  1,789 &  24.25 &  11.89 &    842 &   7.95 &   3.37 \\
&  CiteSeer &  3,312 &  3,312 &   0.11 &   0.05 &  2,459 &   0.15 &   0.07 &  2,845 &  12.58 &   7.14 &  2,082 &   9.35 &   4.85 \\
&   Cornell &    183 &    183 &   0.01 &   0.01 &     99 &   0.01 &   0.01 &    143 &  13.47 &   7.95 &     67 &   6.47 &   3.78 \\
&     Texas &    183 &    183 &   0.22 &   0.10 &     93 &   0.44 &   0.19 &    149 &  14.14 &   6.35 &     74 &   9.14 &   2.77 \\
& Wisconsin &    251 &    251 &   0.33 &   0.22 &    145 &   0.56 &   0.38 &    174 &  20.21 &   8.83 &     93 &   3.69 &   2.63 \\
\bottomrule
\end{tblr}
}
\end{table}

\begin{table}[t]
\centering
\caption{Detailed comparison results of $\toolname$, $\gnnev$, and $\scip$
for weak and general robustness on the Cora, CiteSeer, Cornell, Texas, and Wisconsin datasets with various aggregation functions with global and local budgets $\Delta = 10$.
Note that $\scip$ only implements weak robustness for $\summ$ aggregation.
$t_{a}$ denotes the average runtime, 
and $t_{g}$ denotes the shifted geometric mean of the runtime.}
\label{tab:node_all_10}
\resizebox{\textwidth}{!}{
\begin{tblr}{
        rows    = {abovesep=0.2pt, belowsep=0.2pt},
        row{1}  = {font=\large, abovesep=3pt, belowsep=0pt},
        colspec = {l|l|r|rrrrrr|rrrrrr|rrrrrr},
        cell{1}{4}  = {c=6}{c},
        cell{1}{10} = {c=6}{c},
        cell{1}{16} = {c=6}{c},
        cell{2}{4}  = {c=3}{c},
        cell{2}{7}  = {c=3}{c},
        cell{2}{10} = {c=3}{c},
        cell{2}{13} = {c=3}{c},
        cell{2}{16} = {c=3}{c},
        cell{2}{19} = {c=3}{c}
    }
\toprule
& & & $\toolname$ & & & & & & $\gnnev$ & & & & & & $\scip$ \\
& & &
    All instances & & & Robust instances & & &
    All instances & & & Robust instances & & &
    All instances & & & Robust instances \\
\cmidrule[lr]{4-6}   \cmidrule[lr]{7-9}
\cmidrule[lr]{10-12} \cmidrule[lr]{13-15}
\cmidrule[lr]{16-18} \cmidrule[lr]{19-21}
& & \#Instances &
\#Solved & $t_{a}(s)$ & $t_{g}(s)$ &
\#Solved & $t_{a}(s)$ & $t_{g}(s)$ &
\#Solved & $t_{a}(s)$ & $t_{g}(s)$ &
\#Solved & $t_{a}(s)$ & $t_{g}(s)$ &
\#Solved & $t_{a}(s)$ & $t_{g}(s)$ &
\#Solved & $t_{a}(s)$ & $t_{g}(s)$ \\
\midrule
$\summ$
&      Cora &  2,708 &  2,688 &   1.39 &   0.46 &  2,124 &   1.34 &   0.50 &  1,679 &  19.21 &   8.85 &  1,214 &  16.28 &   6.46 &  1,716 &  49.69 &  15.60 &  1,364 &  39.75 &  11.39 \\
(Weak)
&  CiteSeer &  3,312 &  3,286 &   0.32 &   0.10 &  2,880 &   0.36 &   0.11 &  2,982 &  13.18 &   7.08 &  2,597 &  11.16 &   5.78 &  3,149 &  16.99 &   5.25 &  2,778 &  13.81 &   4.12 \\
&   Cornell &    183 &    183 &   0.01 &   0.01 &     83 &   0.03 &   0.03 &    158 &  12.34 &   7.39 &     67 &  11.68 &   5.14 &    165 &  24.28 &   6.51 &     82 &  18.43 &   4.81 \\
&     Texas &    183 &    182 &   2.64 &   0.79 &    133 &   3.02 &   0.74 &    142 &  11.19 &   4.80 &    112 &   7.03 &   2.77 &    150 &  30.52 &   7.86 &    118 &  10.20 &   2.67 \\
& Wisconsin &    251 &    239 &   3.87 &   0.63 &    182 &   5.08 &   0.83 &    180 &  15.20 &   7.53 &    135 &  10.10 &   4.63 &    195 &  38.17 &  11.70 &    156 &  30.96 &   9.24 \\
\midrule
$\summ$
&      Cora &  2,708 &  2,702 &   0.85 &   0.25 &  1,451 &   1.23 &   0.36 &  2,000 &  21.72 &  10.27 &    964 &  12.22 &   4.27 \\
&  CiteSeer &  3,312 &  3,292 &   0.28 &   0.08 &  2,394 &   0.38 &   0.10 &  3,063 &  17.11 &   8.25 &  2,214 &  13.68 &   5.71 \\
&   Cornell &    183 &    183 &   0.01 &   0.01 &     69 &   0.01 &   0.01 &    171 &  11.29 &   7.74 &     61 &   8.94 &   4.50 \\
&     Texas &    183 &    183 &   0.01 &   0.01 &     61 &   0.02 &   0.02 &    162 &   9.53 &   5.57 &     57 &   1.26 &   1.09 \\
& Wisconsin &    251 &    248 &   0.22 &   0.13 &    141 &   0.22 &   0.14 &    206 &  20.80 &   9.80 &    119 &   9.12 &   4.29 \\
\midrule
$\maxx$
&      Cora &  2,708 &  2,689 &   1.33 &   0.47 &  1,519 &   2.14 &   0.73 &  1,706 &  85.07 &  26.61 &    876 &  10.77 &   4.49 \\
&  CiteSeer &  3,312 &  3,279 &   0.39 &   0.14 &  2,436 &   0.49 &   0.18 &  2,652 &  64.49 &  17.35 &  1,958 &  20.98 &   8.76 \\
&   Cornell &    183 &    183 &   0.61 &   0.16 &     89 &   1.24 &   0.34 &    140 &  39.90 &  18.13 &     59 &  18.39 &   7.15 \\
&     Texas &    183 &    181 &   2.59 &   0.79 &    128 &   2.41 &   0.57 &    127 &  24.87 &   8.18 &     99 &   5.70 &   2.66 \\
& Wisconsin &    251 &    240 &   1.59 &   0.64 &    160 &   1.48 &   0.71 &    142 &  58.16 &  13.21 &    105 &  11.76 &   5.31 \\
\midrule
$\mean$
&      Cora &  2,708 &  2,708 &   0.45 &   0.12 &  1,424 &   0.67 &   0.17 &  1,848 &  22.83 &  11.48 &    837 &   7.41 &   3.14 \\
&  CiteSeer &  3,312 &  3,298 &   0.24 &   0.09 &  2,421 &   0.30 &   0.11 &  2,880 &  13.14 &   7.40 &  2,082 &   9.40 &   4.89 \\
&   Cornell &    183 &    183 &   0.04 &   0.03 &     97 &   0.06 &   0.05 &    146 &  12.78 &   7.48 &     67 &   6.29 &   3.69 \\
&     Texas &    183 &    182 &   1.52 &   0.27 &     88 &   3.14 &   0.56 &    152 &  12.19 &   6.03 &     73 &   6.33 &   2.26 \\
& Wisconsin &    251 &    247 &   1.20 &   0.35 &    135 &   0.82 &   0.41 &    178 &  14.92 &   8.20 &     93 &   3.63 &   2.59 \\
\bottomrule
\end{tblr}
}
\end{table}

\subsection{End to end performance on graph classification}
\label{app:exp4}

Table~\ref{tab:graph_all_1} to Table~\ref{tab:graph_all_6} present detailed comparison results of $\toolname$ and $\scip$
for both weak and general robustness on the MUTAG and ENZYME datasets,
using various aggregation functions and perturbation budgets.
All experiments are conducted on graph classification tasks,
where the fragile set includes all pairs of vertices excluding self-loops.
Note that $\scip$ only supports weak robustness for the $\summ$ aggregation.
$t_{a}$ denotes the average runtime, 
and $t_{g}$ denotes the shifted geometric mean of the runtime.

\begin{table}[t]
\centering
\caption{Detailed comparison results of $\toolname$ and $\scip$
for weak and general robustness on the MUTAG and ENZYMES datasets with various aggregation functions with global budget $\Delta = 1$ and local budgets $\delta = 1$.
Note that $\scip$ only implements weak robustness for $\summ$ aggregation.
$t_{a}$ denotes the average runtime, 
and $t_{g}$ denotes the shifted geometric mean of the runtime.}
\label{tab:graph_all_1}
\resizebox{\textwidth}{!}{
\begin{tblr}{
        rows    = {abovesep=0.2pt, belowsep=0.2pt},
        row{1}  = {font=\large, abovesep=3pt, belowsep=0pt},
        colspec = {l|l|r|rrrrrr|rrrrrr},
        cell{1}{4}  = {c=6}{c},
        cell{1}{10} = {c=6}{c},
        cell{2}{4}  = {c=3}{c},
        cell{2}{7}  = {c=3}{c},
        cell{2}{10} = {c=3}{c},
        cell{2}{13} = {c=3}{c}
    }
\toprule
& & & $\toolname$ & & & & & & $\scip$  \\
& & &
    All instances & & & Robust instances & & &
    All instances & & & Robust instances \\
\cmidrule[lr]{4-6}   \cmidrule[lr]{7-9}
\cmidrule[lr]{10-12} \cmidrule[lr]{13-15}
& & \#Instances &
\#Solved & $t_{a}(s)$ & $t_{g}(s)$ &
\#Solved & $t_{a}(s)$ & $t_{g}(s)$ &
\#Solved & $t_{a}(s)$ & $t_{g}(s)$ &
\#Solved & $t_{a}(s)$ & $t_{g}(s)$ \\
\midrule
$\summ$
&     MUTAG &    188 &    188 &   0.04 &   0.04 &     53 &   0.11 &   0.11 &    104 & 285.14 & 236.39 &      5 & 357.92 & 341.21 \\
(Weak)
&   ENZYMES &    600 &    600 &   0.75 &   0.58 &    354 &   1.19 &   0.93 &    121 & 233.41 & 159.82 &     34 & 234.70 & 157.37 \\
\midrule
$\summ$
&     MUTAG &    188 &    188 &   0.04 &   0.04 &     53 &   0.11 &   0.11 \\
&   ENZYMES &    600 &    600 &   0.46 &   0.33 &    109 &   1.74 &   1.23 \\
\midrule
$\maxx$
&     MUTAG &    188 &    188 &   0.02 &   0.02 &      3 &   0.04 &   0.04 \\
&   ENZYMES &    600 &    600 &   0.12 &   0.10 &     49 &   1.21 &   0.98 \\
\midrule
$\mean$
&     MUTAG &    188 &    188 &   0.06 &   0.06 &    111 &   0.10 &   0.10 \\
&   ENZYMES &    600 &    600 &   0.68 &   0.47 &    183 &   1.92 &   1.27 \\
\bottomrule
\end{tblr}
}
\end{table}

\begin{table}[t]
\centering
\caption{Detailed comparison results of $\toolname$ and $\scip$
for weak and general robustness on the MUTAG and ENZYMES datasets with various aggregation functions with global budget $\Delta = 2$ and local budgets $\delta = 1$.
Note that $\scip$ only implements weak robustness for $\summ$ aggregation.
$t_{a}$ denotes the average runtime, 
and $t_{g}$ denotes the shifted geometric mean of the runtime.}
\label{tab:graph_all_2}
\resizebox{\textwidth}{!}{
\begin{tblr}{
        rows    = {abovesep=0.2pt, belowsep=0.2pt},
        row{1}  = {font=\large, abovesep=3pt, belowsep=0pt},
        colspec = {l|l|r|rrrrrr|rrrrrr},
        cell{1}{4}  = {c=6}{c},
        cell{1}{10} = {c=6}{c},
        cell{2}{4}  = {c=3}{c},
        cell{2}{7}  = {c=3}{c},
        cell{2}{10} = {c=3}{c},
        cell{2}{13} = {c=3}{c}
    }
\toprule
& & & $\toolname$ & & & & & & $\scip$  \\
& & &
    All instances & & & Robust instances & & &
    All instances & & & Robust instances \\
\cmidrule[lr]{4-6}   \cmidrule[lr]{7-9}
\cmidrule[lr]{10-12} \cmidrule[lr]{13-15}
& & \#Instances &
\#Solved & $t_{a}(s)$ & $t_{g}(s)$ &
\#Solved & $t_{a}(s)$ & $t_{g}(s)$ &
\#Solved & $t_{a}(s)$ & $t_{g}(s)$ &
\#Solved & $t_{a}(s)$ & $t_{g}(s)$ \\
\midrule
$\summ$
&     MUTAG &    188 &    188 &   0.05 &   0.04 &      2 &   2.98 &   2.64 &     61 & 210.07 & 123.20 &      0 &    N/A &    N/A \\
(Weak)
&   ENZYMES &    600 &    588 &  24.63 &   6.71 &    110 &  96.06 &  42.28 &     52 & 120.58 &  60.49 &      4 & 133.26 &  58.31 \\
\midrule
$\summ$
&     MUTAG &    188 &    188 &   0.05 &   0.04 &      2 &   3.06 &   2.71 \\
&   ENZYMES &    600 &    596 &   5.14 &   1.40 &     24 &  55.18 &  22.22 \\
\midrule
$\maxx$
&     MUTAG &    188 &    188 &   0.01 &   0.01 &      0 &    N/A &    N/A \\
&   ENZYMES &    600 &    599 &   0.78 &   0.23 &     16 &   0.63 &   0.45 \\
\midrule
$\mean$
&     MUTAG &    188 &    188 &   1.36 &   1.17 &     71 &   3.37 &   3.12 \\
&   ENZYMES &    600 &    587 &  11.72 &   2.83 &     46 &  64.00 &  17.58 \\
\bottomrule
\end{tblr}
}
\end{table}

\begin{table}[t]
\centering
\caption{Detailed comparison results of $\toolname$ and $\scip$
for weak and general robustness on the MUTAG and ENZYMES datasets with various aggregation functions with global budget $\Delta = 2$ and local budgets $\delta = 2$.
Note that $\scip$ only implements weak robustness for $\summ$ aggregation.
$t_{a}$ denotes the average runtime, 
and $t_{g}$ denotes the shifted geometric mean of the runtime.}
\label{tab:graph_all_3}
\resizebox{\textwidth}{!}{
\begin{tblr}{
        rows    = {abovesep=0.2pt, belowsep=0.2pt},
        row{1}  = {font=\large, abovesep=3pt, belowsep=0pt},
        colspec = {l|l|r|rrrrrr|rrrrrr},
        cell{1}{4}  = {c=6}{c},
        cell{1}{10} = {c=6}{c},
        cell{2}{4}  = {c=3}{c},
        cell{2}{7}  = {c=3}{c},
        cell{2}{10} = {c=3}{c},
        cell{2}{13} = {c=3}{c}
    }
\toprule
& & & $\toolname$ & & & & & & $\scip$  \\
& & &
    All instances & & & Robust instances & & &
    All instances & & & Robust instances \\
\cmidrule[lr]{4-6}   \cmidrule[lr]{7-9}
\cmidrule[lr]{10-12} \cmidrule[lr]{13-15}
& & \#Instances &
\#Solved & $t_{a}(s)$ & $t_{g}(s)$ &
\#Solved & $t_{a}(s)$ & $t_{g}(s)$ &
\#Solved & $t_{a}(s)$ & $t_{g}(s)$ &
\#Solved & $t_{a}(s)$ & $t_{g}(s)$ \\
\midrule
$\summ$
&     MUTAG &    188 &    188 &   0.07 &   0.06 &      2 &   3.87 &   3.35 &     76 &  39.47 &  11.67 &      0 &    N/A &    N/A \\
(Weak)
&   ENZYMES &    600 &    588 &  26.06 &   6.78 &     85 & 121.91 &  54.86 &     55 &  84.76 &  32.26 &      3 &  75.90 &  38.95 \\
\midrule
$\summ$
&     MUTAG &    188 &    188 &   0.07 &   0.06 &      2 &   3.92 &   3.39 \\
&   ENZYMES &    600 &    596 &   5.15 &   1.40 &     22 &  54.40 &  20.91 \\
\midrule
$\maxx$
&     MUTAG &    188 &    188 &   0.02 &   0.02 &      0 &    N/A &    N/A \\
&   ENZYMES &    600 &    599 &   0.81 &   0.24 &     16 &   0.74 &   0.52 \\
\midrule
$\mean$
&     MUTAG &    188 &    188 &   1.58 &   1.29 &     42 &   6.40 &   6.15 \\
&   ENZYMES &    600 &    586 &  10.80 &   2.59 &     35 &  58.87 &  17.14 \\
\bottomrule
\end{tblr}
}
\end{table}

\begin{table}[t]
\centering
\caption{Detailed comparison results of $\toolname$ and $\scip$
for weak and general robustness on the MUTAG and ENZYMES datasets with various aggregation functions with global budget $\Delta = 5$ and local budgets $\delta = 1$.
Note that $\scip$ only implements weak robustness for $\summ$ aggregation.
$t_{a}$ denotes the average runtime, 
and $t_{g}$ denotes the shifted geometric mean of the runtime.}
\label{tab:graph_all_4}
\resizebox{\textwidth}{!}{
\begin{tblr}{
        rows    = {abovesep=0.2pt, belowsep=0.2pt},
        row{1}  = {font=\large, abovesep=3pt, belowsep=0pt},
        colspec = {l|l|r|rrrrrr|rrrrrr},
        cell{1}{4}  = {c=6}{c},
        cell{1}{10} = {c=6}{c},
        cell{2}{4}  = {c=3}{c},
        cell{2}{7}  = {c=3}{c},
        cell{2}{10} = {c=3}{c},
        cell{2}{13} = {c=3}{c}
    }
\toprule
& & & $\toolname$ & & & & & & $\scip$  \\
& & &
    All instances & & & Robust instances & & &
    All instances & & & Robust instances \\
\cmidrule[lr]{4-6}   \cmidrule[lr]{7-9}
\cmidrule[lr]{10-12} \cmidrule[lr]{13-15}
& & \#Instances &
\#Solved & $t_{a}(s)$ & $t_{g}(s)$ &
\#Solved & $t_{a}(s)$ & $t_{g}(s)$ &
\#Solved & $t_{a}(s)$ & $t_{g}(s)$ &
\#Solved & $t_{a}(s)$ & $t_{g}(s)$ \\
\midrule
$\summ$
&     MUTAG &    188 &    188 &   0.07 &   0.05 &      1 &   1.29 &   1.29 &     63 & 139.06 &  78.47 &      0 &    N/A &    N/A \\
(Weak)
&   ENZYMES &    600 &    541 &  13.57 &   3.42 &      6 &   0.30 &   0.28 &    108 &  89.03 &  52.18 &      3 &  46.86 &  29.38 \\
\midrule
$\summ$
&     MUTAG &    188 &    188 &   0.07 &   0.05 &      1 &   1.26 &   1.26 \\
&   ENZYMES &    600 &    587 &   2.15 &   0.70 &      3 &   0.61 &   0.58 \\
\midrule
$\maxx$
&     MUTAG &    188 &    188 &   0.01 &   0.01 &      0 &    N/A &    N/A \\
&   ENZYMES &    600 &    598 &   0.46 &   0.20 &     15 &   0.01 &   0.01 \\
\midrule
$\mean$
&     MUTAG &    188 &    155 &   4.83 &   1.61 &      0 &    N/A &    N/A \\
&   ENZYMES &    600 &    566 &  10.36 &   2.55 &      9 &  82.40 &  24.80 \\
\bottomrule
\end{tblr}
}
\end{table}

\begin{table}[t]
\centering
\caption{Detailed comparison results of $\toolname$ and $\scip$
for weak and general robustness on the MUTAG and ENZYMES datasets with various aggregation functions with global budget $\Delta = 5$ and local budgets $\delta = 2$.
Note that $\scip$ only implements weak robustness for $\summ$ aggregation.
$t_{a}$ denotes the average runtime, 
and $t_{g}$ denotes the shifted geometric mean of the runtime.}
\label{tab:graph_all_5}
\resizebox{\textwidth}{!}{
\begin{tblr}{
        rows    = {abovesep=0.2pt, belowsep=0.2pt},
        row{1}  = {font=\large, abovesep=3pt, belowsep=0pt},
        colspec = {l|l|r|rrrrrr|rrrrrr},
        cell{1}{4}  = {c=6}{c},
        cell{1}{10} = {c=6}{c},
        cell{2}{4}  = {c=3}{c},
        cell{2}{7}  = {c=3}{c},
        cell{2}{10} = {c=3}{c},
        cell{2}{13} = {c=3}{c}
    }
\toprule
& & & $\toolname$ & & & & & & $\scip$  \\
& & &
    All instances & & & Robust instances & & &
    All instances & & & Robust instances \\
\cmidrule[lr]{4-6}   \cmidrule[lr]{7-9}
\cmidrule[lr]{10-12} \cmidrule[lr]{13-15}
& & \#Instances &
\#Solved & $t_{a}(s)$ & $t_{g}(s)$ &
\#Solved & $t_{a}(s)$ & $t_{g}(s)$ &
\#Solved & $t_{a}(s)$ & $t_{g}(s)$ &
\#Solved & $t_{a}(s)$ & $t_{g}(s)$ \\
\midrule
$\summ$
&     MUTAG &    188 &    188 &   0.09 &   0.06 &      0 &    N/A &    N/A &     65 &  22.90 &   8.22 &      0 &    N/A &    N/A \\
(Weak)
&   ENZYMES &    600 &    539 &  14.82 &   3.97 &      2 &   0.01 &   0.01 &    133 &  31.23 &  15.76 &      2 &  40.27 &  20.17 \\
\midrule
$\summ$
&     MUTAG &    188 &    188 &   0.10 &   0.07 &      0 &    N/A &    N/A \\
&   ENZYMES &    600 &    590 &   4.75 &   1.22 &      1 &   0.01 &   0.01 \\
\midrule
$\maxx$
&     MUTAG &    188 &    188 &   0.01 &   0.01 &      0 &    N/A &    N/A \\
&   ENZYMES &    600 &    595 &   0.30 &   0.16 &     10 &   0.15 &   0.14 \\
\midrule
$\mean$
&     MUTAG &    188 &    187 &   6.28 &   1.04 &      0 &    N/A &    N/A \\
&   ENZYMES &    600 &    569 &   8.85 &   2.44 &      4 &  41.62 &  10.57 \\
\bottomrule
\end{tblr}
}
\end{table}

\begin{table}[t]
\centering
\caption{Detailed comparison results of $\toolname$ and $\scip$
for weak and general robustness on the MUTAG and ENZYMES datasets with various aggregation functions with global budget $\Delta = 5$ and local budgets $\delta = 5$.
Note that $\scip$ only implements weak robustness for $\summ$ aggregation.
$t_{a}$ denotes the average runtime, 
and $t_{g}$ denotes the shifted geometric mean of the runtime.}
\label{tab:graph_all_6}
\resizebox{\textwidth}{!}{
\begin{tblr}{
        rows    = {abovesep=0.2pt, belowsep=0.2pt},
        row{1}  = {font=\large, abovesep=3pt, belowsep=0pt},
        colspec = {l|l|r|rrrrrr|rrrrrr},
        cell{1}{4}  = {c=6}{c},
        cell{1}{10} = {c=6}{c},
        cell{2}{4}  = {c=3}{c},
        cell{2}{7}  = {c=3}{c},
        cell{2}{10} = {c=3}{c},
        cell{2}{13} = {c=3}{c}
    }
\toprule
& & & $\toolname$ & & & & & & $\scip$  \\
& & &
    All instances & & & Robust instances & & &
    All instances & & & Robust instances \\
\cmidrule[lr]{4-6}   \cmidrule[lr]{7-9}
\cmidrule[lr]{10-12} \cmidrule[lr]{13-15}
& & \#Instances &
\#Solved & $t_{a}(s)$ & $t_{g}(s)$ &
\#Solved & $t_{a}(s)$ & $t_{g}(s)$ &
\#Solved & $t_{a}(s)$ & $t_{g}(s)$ &
\#Solved & $t_{a}(s)$ & $t_{g}(s)$ \\
\midrule
$\summ$
&     MUTAG &    188 &    188 &   0.52 &   0.44 &      0 &    N/A &    N/A &    109 &  53.29 &  21.68 &      0 &    N/A &    N/A \\
(Weak)
&   ENZYMES &    600 &    529 &  15.80 &   4.14 &      2 &   0.01 &   0.01 &    203 & 116.51 &  50.24 &      2 &  47.47 &  22.48 \\
\midrule
$\summ$
&     MUTAG &    188 &    188 &   0.52 &   0.44 &      0 &    N/A &    N/A \\
&   ENZYMES &    600 &    587 &   4.35 &   1.25 &      1 &   0.01 &   0.01 \\
\midrule
$\maxx$
&     MUTAG &    188 &    188 &   0.01 &   0.01 &      0 &    N/A &    N/A \\
&   ENZYMES &    600 &    595 &   0.66 &   0.28 &      9 &  18.71 &  10.99 \\
\midrule
$\mean$
&     MUTAG &    188 &    188 &  16.07 &   4.70 &      0 &    N/A &    N/A \\
&   ENZYMES &    600 &    579 &   9.48 &   2.34 &      3 &  89.99 &  20.38 \\
\bottomrule
\end{tblr}
}
\end{table}

\end{document}